%% file: tro.tex
\documentclass[journal]{IEEEtran}
\usepackage{decar-common}    % Commonly used commands and packages

\usepackage{amsmath,amsfonts}
\usepackage{algorithmic}
\usepackage{algorithm}
\usepackage{array}
\usepackage[caption=false]{subfig}
\usepackage{textcomp}
\usepackage{stfloats}
\usepackage{url}
\usepackage{verbatim}
\usepackage{graphicx}
\usepackage{cite}
\usepackage{adjustbox}
\usepackage{lipsum}  
\newcommand{\jln}[1]{\textcolor{black}{#1}}
\newcommand{\ms}[1]{\textcolor{black}{#1}}
\newcommand{\tro}[1]{\textcolor{black}{#1}}

\begin{document}

\title{Multi-Robot Relative Pose Estimation and IMU Preintegration Using Passive UWB Transceivers}

\author{Mohammed Ayman Shalaby, Charles Champagne Cossette, Jerome Le Ny, James Richard Forbes 
        % <-this % stops a space
\thanks{This work was supported by the NSERC Alliance Grant program, the NSERC Discovery Grant program, the CFI JELF program, and FRQNT Award 2018-PR-253646.}% <-this % stops a space
\thanks{M. A. Shalaby, C. C. Cossette, and J. R. Forbes are with the department of Mechanical Engineering, McGill University, Montreal, QC H3A 0C3, Canada. {\{\tt\small mohammed.shalaby@mail.mcgill.ca, charles.cossette@mail.mcgill.ca, james.richard.forbes@mcgill.ca\}.} J. Le Ny is with the department of Electrical Engineering, Polytechnique Montreal, Montreal, QC H3T 1J4, Canada. \{\tt\small jerome.le-ny@polymtl.ca\}.}}

% The paper headers
% \markboth{Journal of \LaTeX\ Class Files,~Vol.~14, No.~8, August~2021}%
% {Shell \MakeLowercase{\textit{et al.}}: A Sample Article Using IEEEtran.cls for IEEE Journals}

% \IEEEpubid{0000--0000/00\$00.00~\copyright~2021 IEEE}
% Remember, if you use this you must call \IEEEpubidadjcol in the second
% column for its text to clear the IEEEpubid mark.

\maketitle

\begin{abstract}
Ultra-wideband (UWB) systems are becoming increasingly popular as a means of inter-robot ranging and communication. A major constraint associated with UWB is that only one pair of UWB transceivers can range at a time to avoid interference, hence hindering the scalability of UWB-based localization. In this paper, a ranging protocol is proposed that allows all robots to passively listen on neighbouring communicating robots without any hierarchical restrictions on the role of the robots. This is utilized to allow each robot to obtain more \ms{range} measurements and to broadcast preintegrated inertial measurement unit (IMU) measurements for relative extended pose state \ms{estimation} directly on $SE_2(3)$. Consequently, a simultaneous clock-synchronization and relative-pose estimator (CSRPE) is formulated using an on-manifold extended Kalman filter (EKF) and is evaluated in simulation using Monte-Carlo runs for up to 7 robots. The ranging protocol is implemented in C on custom-made UWB boards fitted to 3 quadcopters, and the proposed filter is evaluated over multiple \jln{experimental} trials, yielding up to \tro{48\%} improvement in localization accuracy. 
\end{abstract}


\section{Introduction}
\input{sections/intro.tex}

\section{Related Work} \label{sec:related_work}
\input{sections/related_work.tex}

\section{Preliminaries} \label{sec:prelims}
\input{sections/prelims.tex}

\section{Problem Formulation} \label{sec:problem_formulation}
\input{sections/problem_formulation.tex}

\section{Ranging Protocol} \label{sec:ranging_protocol}
\input{sections/ranging_protocol.tex}

\section{The Process Model} \label{sec:process_model}
\input{sections/process_model.tex}

\section{Relative Pose State Preintegration} \label{sec:preint}
\input{sections/preintegration.tex}

% \section{Initialization}
% \input{sections/initialization.tex}

\section{Simulation Results} \label{sec:sim}
\input{sections/simulation.tex}

\section{Experimental Results} \label{sec:exp}
\input{sections/experiment.tex}

\section{\tro{Further Practical Considerations}} \label{sec:practical}
\input{sections/practical.tex}

\section{Conclusion} \label{sec:conclusion}
\input{sections/conclusion.tex}

\appendices

\section{Fold Increase in Measurements} \label{appx:fold}
\input{sections/appx_fold.tex}

% The following appendix is now obsolete
%\section{Approximating the Time Delays} \label{appx:approx_delta_t}
%\input{sections/appx_approx_delta_t.tex}

\section{Linearizing the Range Measurement Model} \label{appx:linearize_range}
\input{sections/appx_linearize_range.tex}

\section{Discretizing the Input Matrix} \label{appx:U}
\input{sections/appx_U.tex}

\addcontentsline{toc}{section}{References}
\bibliographystyle{IEEEtran}
\bibliography{tro}

% \newpage
\section{Biography Section}
\begin{IEEEbiography}[{\includegraphics[width=1in,height=1.25in,clip,keepaspectratio,trim={1cm 5cm 1cm 0cm}]{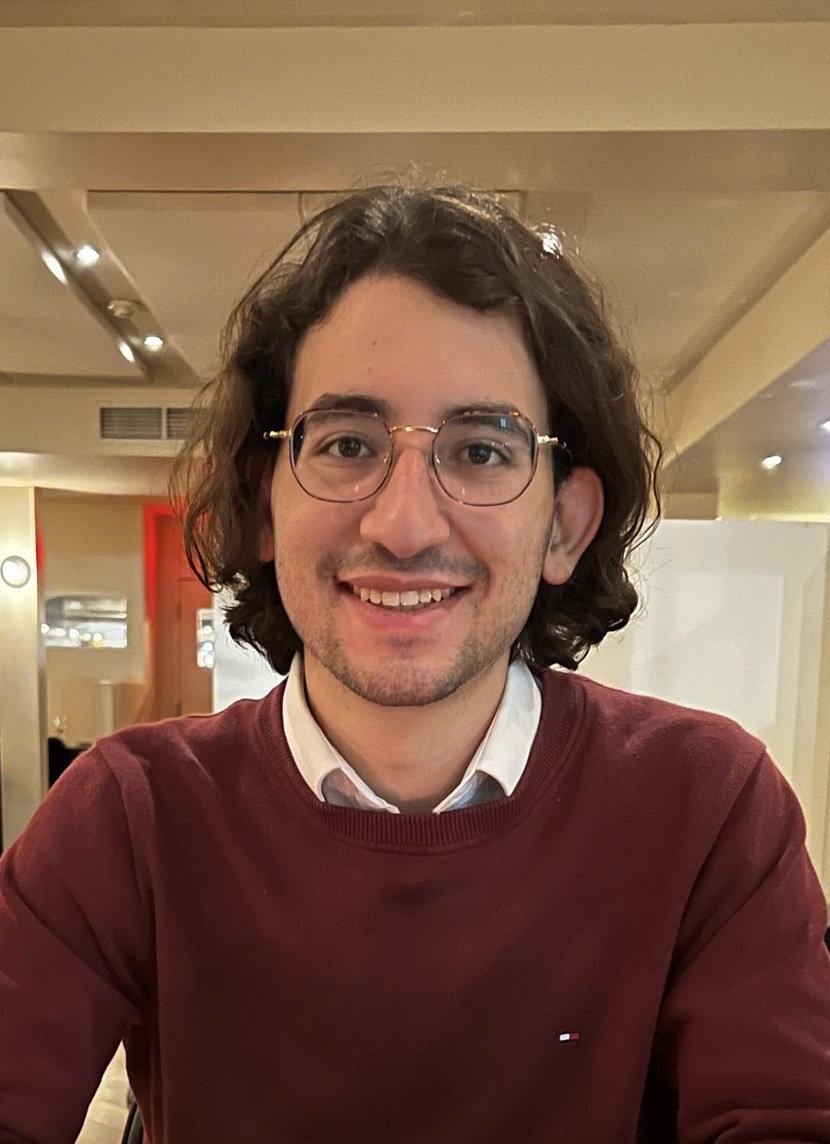}}]{Mohammed Ayman Shalaby}
        (Student Member, IEEE) received the B.Eng. degree in Mechanical Engineering from McGill University, Montreal, QC, Canada in 2019. He is currently a Ph.D. Candidate at McGill University. His research interests include state estimation, multi-robot systems, and ultra-wideband communication, with applications to autonomous navigation of robotic systems.
\end{IEEEbiography}

\begin{IEEEbiography}[{\includegraphics[width=1in,height=1.25in,clip,keepaspectratio,trim={1cm 0 1cm 0}]{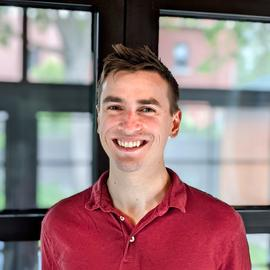}}]{Charles Champagne Cossette} 
        (Student Member, IEEE) earned his Ph.D from McGill University in 2023. Charles is working on state estimation and planning for multi-robot teams using ultra-wideband radio.
\end{IEEEbiography}

\begin{IEEEbiography}[{\includegraphics[width=1in,height=1.25in,clip,keepaspectratio,trim={1cm 0 1cm 0}]{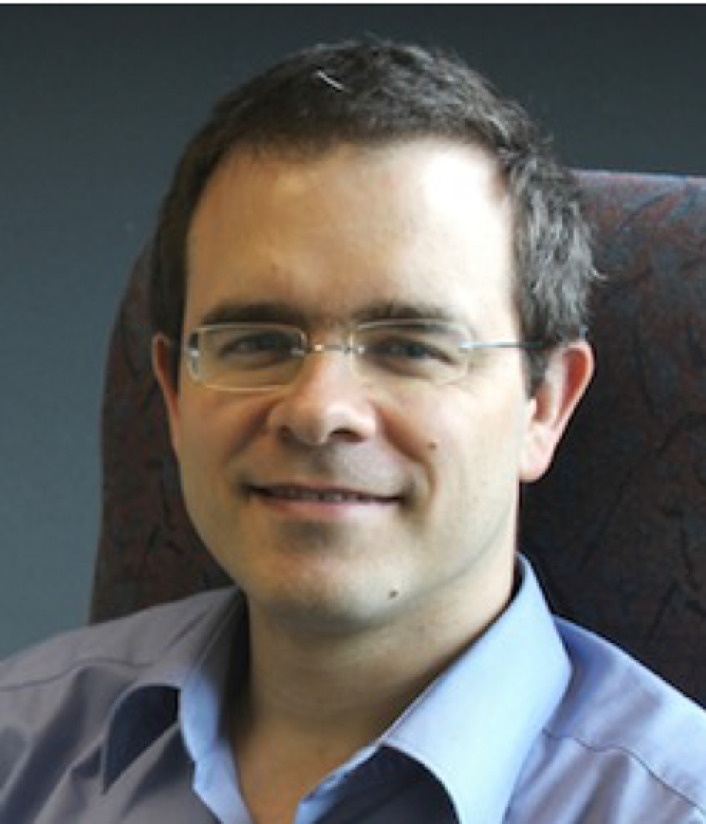}}]{Jerome Le Ny}
        (Senior Member, IEEE) received the Ph.D. degree in Aeronautics and Astronautics from the Massachusetts Institute of Technology, Cambridge, in 2008. He is currently an Associate Professor with the Department of Electrical Engineering, Polytechnique Montreal, Canada, and a member of GERAD, a multi-university research center on decision analysis. From 2008 to 2012 he was a Postdoctoral Researcher with the GRASP Laboratory at the University of Pennsylvania. In 2018-2019, he was an Alexander von Humboldt Fellow at the Technical University of Munich. His research interests include robust and stochastic control, mean-field control, networked control systems, dynamic resource allocation problems, privacy and security in sensor and actuator networks, with applications to autonomous multi-robot systems and intelligent infrastructure systems. He currently serves as an Associate Editor for the IEEE Transactions on Robotics.
\end{IEEEbiography}

\begin{IEEEbiography}[{\includegraphics[width=1in,height=1.25in,clip,keepaspectratio]{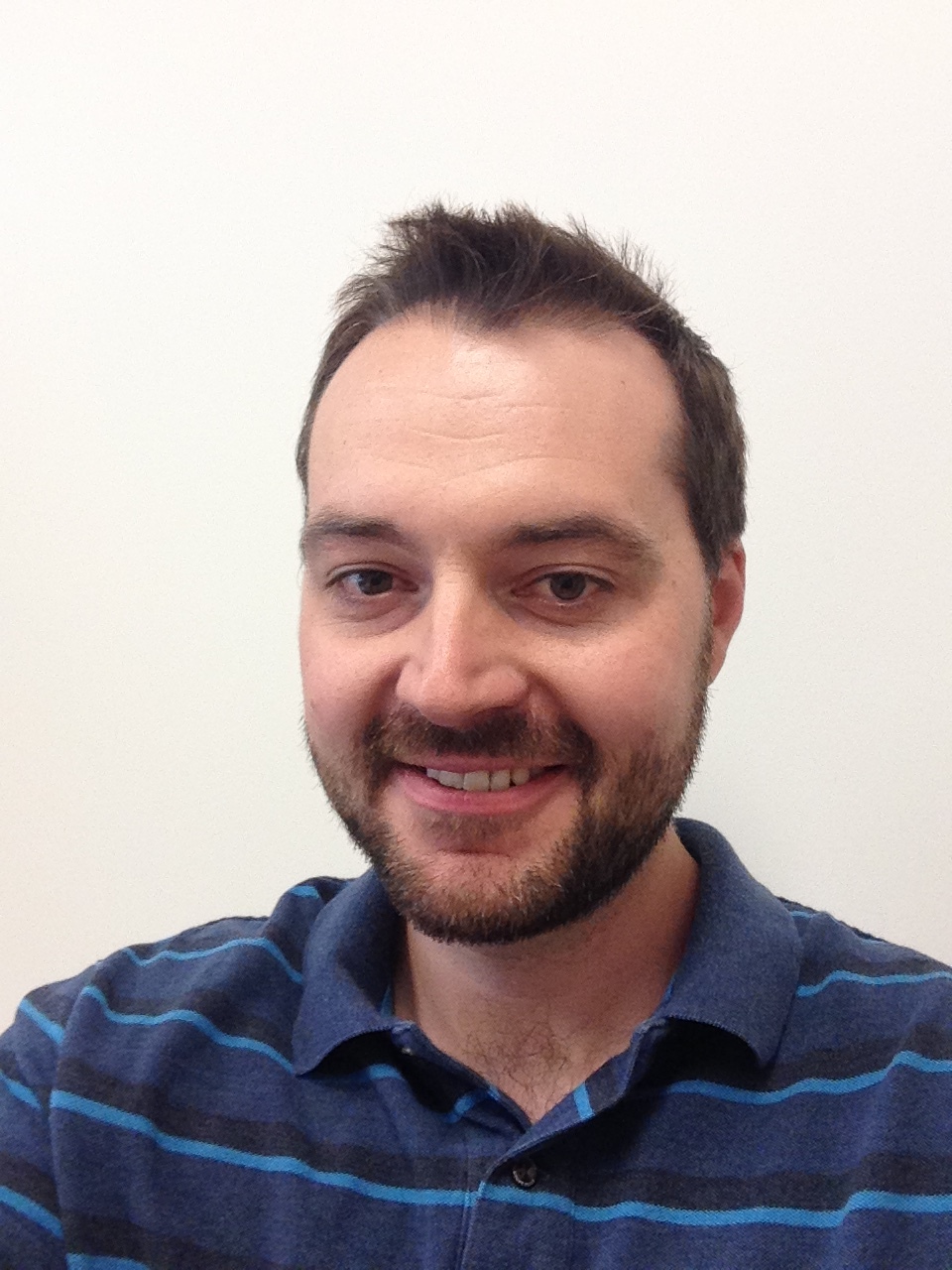}}]{James Richard Forbes}
        (Member, IEEE) received the B.A.Sc. degree in Mechanical Engineering (Honours, Co-op) from the University of Waterloo, Waterloo, ON, Canada in 2006, and the M.A.Sc. and Ph.D. degrees in Aerospace Science and Engineering from the University of Toronto Institute for Aerospace Studies (UTIAS), Toronto, ON, Canada in 2008 and 2011, respectively. James is currently an Associate Professor and William Dawson Scholar in the Department of Mechanical Engineering at McGill University, Montreal, QC, Canada. James is a Member of the Centre for Intelligent Machines (CIM), a Member of the Group for Research in Decision Analysis (GERAD), and a Member of the Trottier Institute for Sustainability in Engineering and Design (TISED). James was awarded the McGill Association of Mechanical Engineers (MAME) Professor of the Year Award in 2016, the Engineering Class of 1944 Outstanding Teaching Award in 2018, and the Carrie M. Derick Award for Graduate Supervision and Teaching in 2020. The focus of James' research is navigation, guidance, and control of robotic systems. James is currently an Associate Editor of the International Journal of Robotics Research (IJRR). 
\end{IEEEbiography}

\vfill

\end{document}

%% file: sections/intro.tex
% #TODO: things when I get the first revision:
% 1) Show algorithm for the scheduling protocol with failure handling.
% 2) Could explain why assuming tag 0 is real-time is fine, and how much error is introduced in the speed of light formulation.
% 3) WHAT ABOUT SHOWING CRLB REDUCTION DUE TO PASSIVE LISTENING MEASUREMENTS?
% 4) What happens when only a subset of the pairs communicate and all the other ones passively listen at all times?

\begin{figure}
    \centering
    \begin{minipage}{0.48\columnwidth}%
        \centering
        \subfloat[Subfigure 1 list of figures text][Custom-made board fitted with \\ a DWM1000 UWB transceiver.]{
        \includegraphics[trim={2.2cm 0cm 0cm 0cm},clip,width=\textwidth]{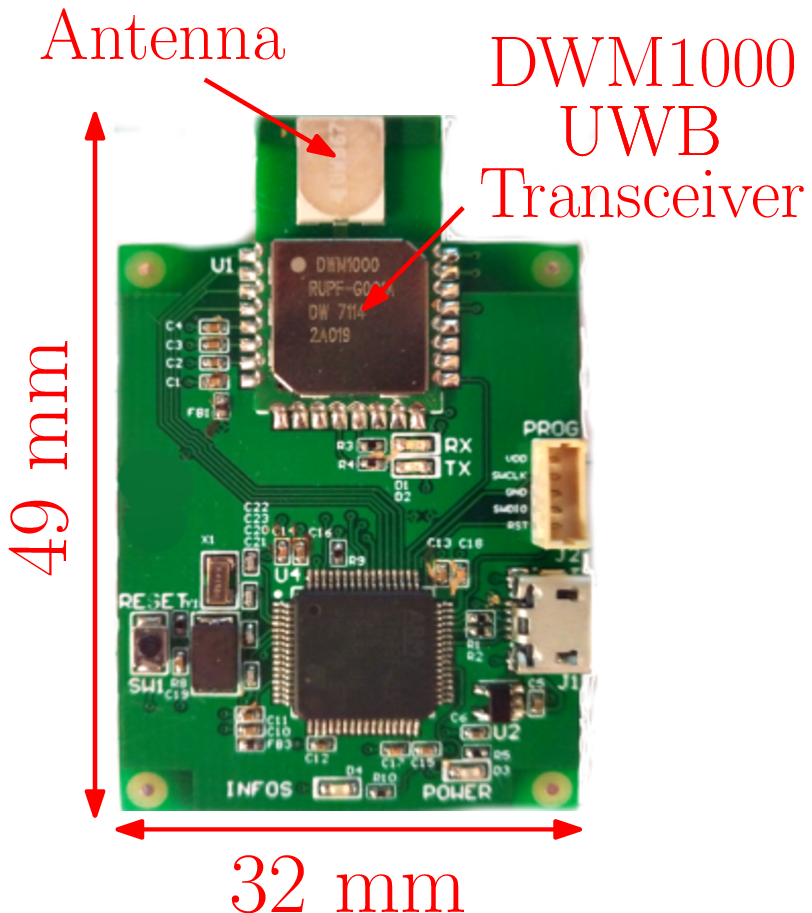}
        \label{fig:ourboard_compressed}}
    \end{minipage}%
    \begin{minipage}{0.49\columnwidth}%
        \centering
        \subfloat[Subfigure 1 list of figures text][A Uvify IFO-S quadcopter equipped with two UWB transceivers 45 cm apart.]{
        \includegraphics[trim={2cm 0cm 2cm 0cm},clip,width=\textwidth]{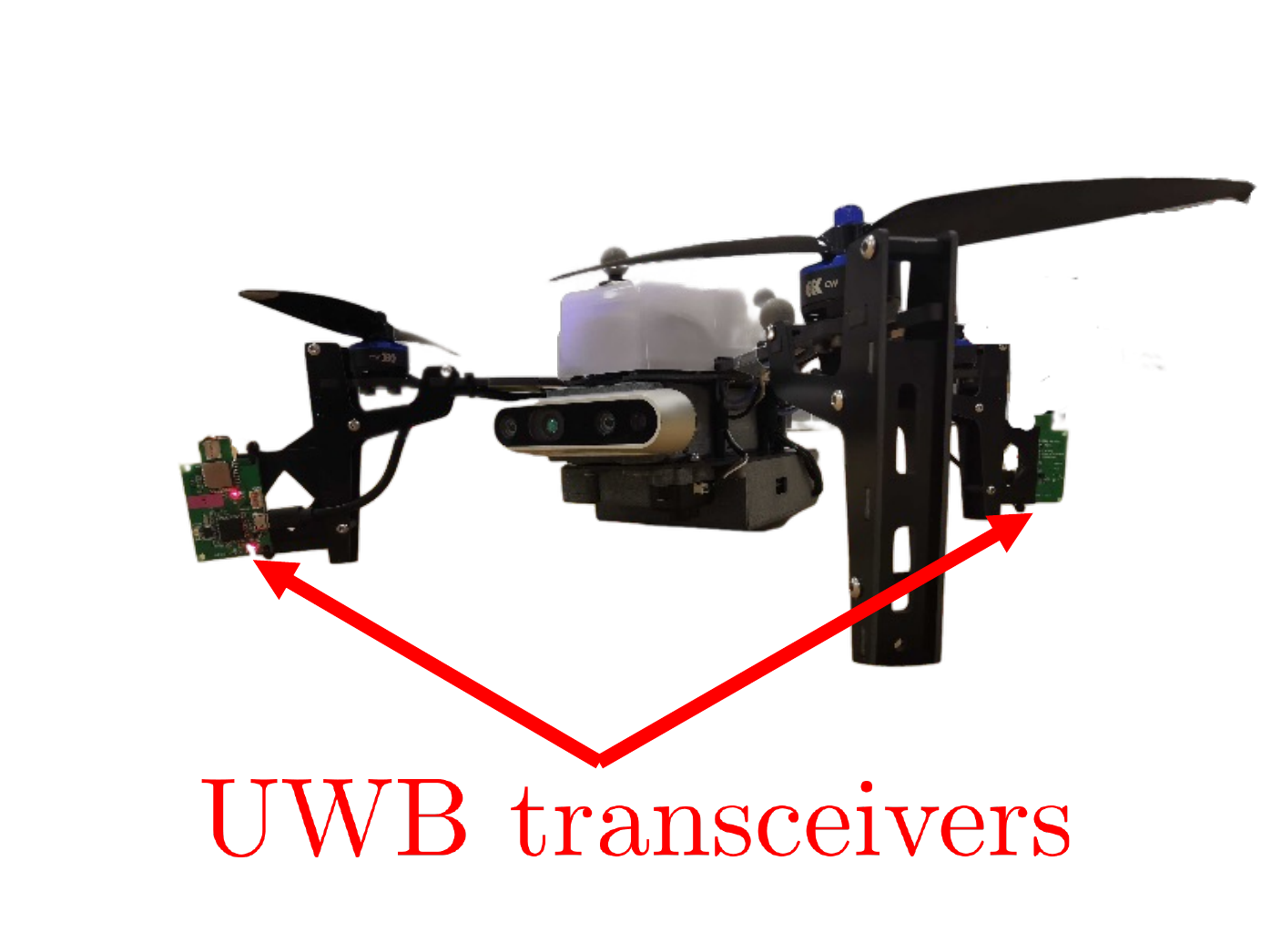}
        \label{fig:ifo_cropped_marked}}
    \end{minipage}
    \caption{The experimental set-up.}
    \label{fig:exp_setup}
\end{figure}

\IEEEPARstart{M}{ulti-robot} teams' prevalence is a direct consequence of two factors, recent advancements in available technology and demand for automating complex tasks. The former has recently been accelerated through the adoption of \emph{ultra-wideband} (UWB) radio-frequency signals as a means of \emph{ranging} and communication between robots, where ranging means obtaining distance measurements. 
UWB is 
%a particularly attractive choice as it is 
a relatively inexpensive, low-power, lightweight, and compact \jln{technology, which} allows for high-rate ranging and data transfer. 
An example of UWB boards fitted to a quadcopter is shown in Figure~\ref{fig:exp_setup}. 
Robotic teams equipped with UWB and other sensors, such as cameras and/or \emph{inertial measurement units} (IMUs), have been considered for relative pose estimation, which is a prerequisite for applications such as collision avoidance and collaborative mapping and infrastructure inspection.

\begin{figure}[t!]
    \centering
    \includegraphics[trim={20cm 0cm 20cm 0cm},clip,width=\columnwidth]{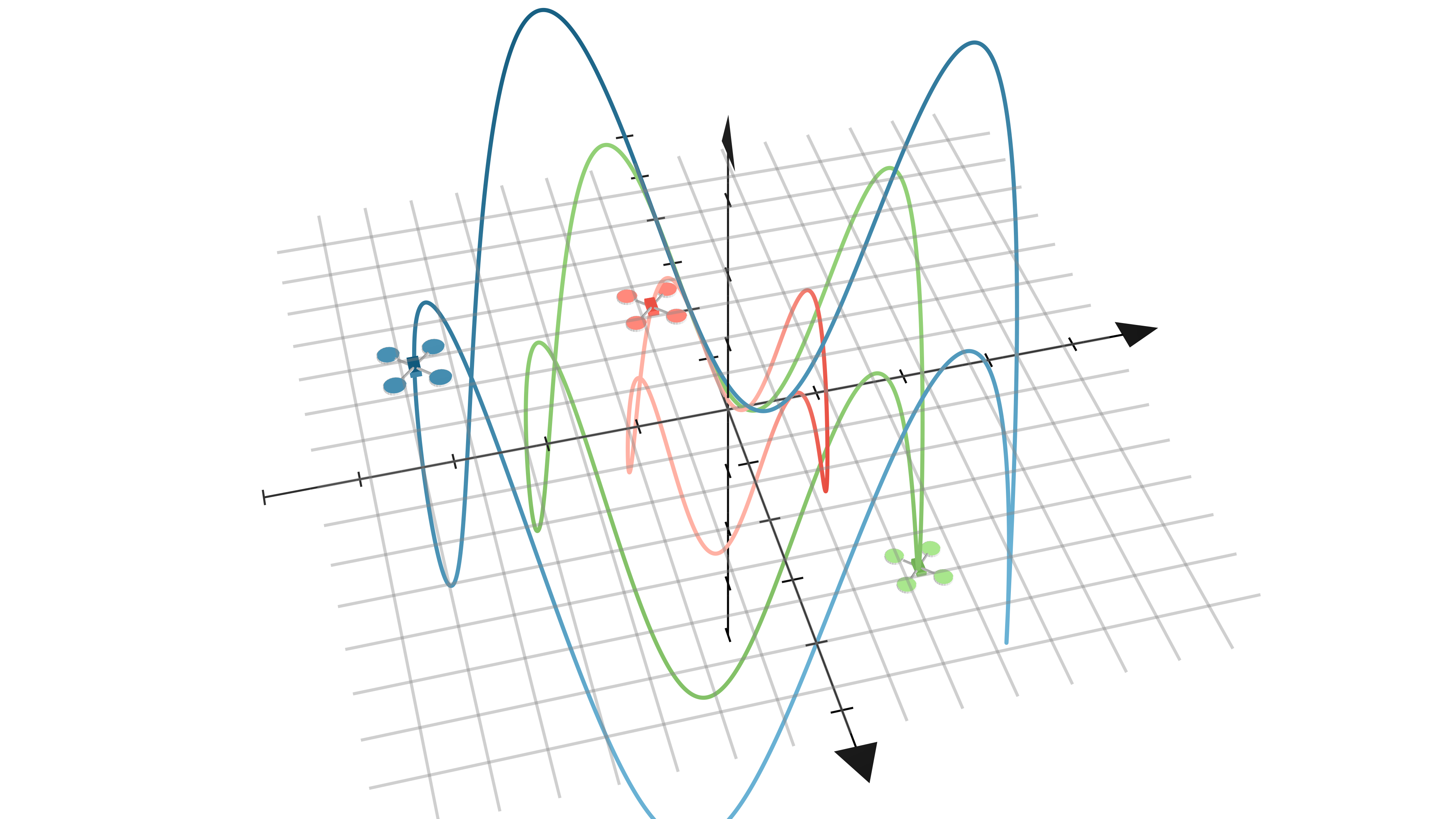}
    \caption{The trajector\jln{ies followed} by three simulated quadcopters.}
    \label{fig:sim_trajectory}
\end{figure}

Nonetheless, using UWB for relative pose estimation in multi-robot teams introduces a distinct set of problems. Firstly, UWB ranging and communication is not robust to interference, thus imposing the constraint that only one pair of transceivers can communicate at a time. This is typically addressed using \emph{time-division multiple-access} (TDMA) media-access control (MAC) protocols alongside a round-robin approach \jln{to determine} which pair communicates at each time. However, the larger the team of robots, the longer the time gaps in between a robot ranging with another. Another complication with UWB ranging is the reliance on \emph{time-of-flight} (ToF) measurements, which necessitates the presence of a clock at each UWB transceiver. However, \jln{in practice} these clocks run at different rates, and therefore require some synchronization mechanism. The importance of synchronization can be highlighted by the fact that 1 ns in synchronization error translates to $c \text{ [m/s] } \times 10^{-9} \text{ [s] } \approx 30 \text{ [cm] }$ in localization error, where $c$ is the speed of light.

Another practical issue associated with multi-robot systems is communication constraints, which limit the amount of information that can be \jln{transmitted} between robots. In filtering applications where there are for example 3 quadcopters moving randomly in 3-dimensional space as shown in Figure~\ref{fig:sim_trajectory}, IMU \jln{measurements} must be broadcasted if robots are to estimate their neighbours' \ms{relative} states directly from the raw measurements. Nonetheless, IMU measurements are typically recorded at a very high frequency, and the constraint that only one pair can be communicating at a time means that communication links between robots do not always exist. Therefore, a more efficient way of sharing odometry information is required. 

To achieve a practical relative pose estimation solution that is implementable on a robotic team, this paper addresses the aforementioned constraints. The contributions of this work are summarized as follows.
\begin{enumerate}
    \item A ranging protocol is introduced that extends classical ranging protocols by allowing neighbouring robots to passively listen to the measurements and timestamp receptions, with no assumptions or imposed constraints on the robots' hierarchy. The concept of passive listening is then utilized to provide a $(1+3n)$-fold increase in the number of measurements recorded when there are a total of $n+1$ robots each equipped with two UWB transceivers. The concept of passive listening is additionally utilized for more efficient information sharing and implementing simple MAC protocols.    
    \item Representing the extended pose state as an element of $SE_2(3)$, an on-manifold tightly-coupled simultaneous clock-synchronization and relative-pose estimator (CSRPE) is then proposed, which allows incoporating passive listening measurements in an extended Kalman filter (EKF) to improve the relative pose estimation. This provides a means for many different robots to estimate the relative poses of their neighbours relative to themselves at a high frequency.
    \item Rather than sharing high-frequency IMU readings with neighbours, the concept of preintegration \cite{Forster2017} is developed for relative pose states on $SE_2(3)$, and is used as a means of efficient IMU data logging and communication between robots. This is additionally incorporated in the CSRPE, where the theory behind filtering with delayed inputs is developed as the preintegrated IMU measurements arrive asynchronously from neighbouring robots. 
    \item The proposed algorithm is evaluated in simulation using Monte-Carlo trials and in experiments using 4 trials with 3 quadcopters equipped with two UWB transceivers each. It is shown that localization accuracy improves up to \tro{23\% when compared to a centralized scenario} and up to \tro{48\%} when compared to the case of no passive listening. 
\end{enumerate}

The remainder of the paper is organized as follows. Related work is presented in Section \ref{sec:related_work}, and Lie group and UWB preliminaries are discussed in Section \ref{sec:prelims}. The problem is formulated in Section \ref{sec:problem_formulation}, then the proposed ranging protocol is discussed in Section \ref{sec:ranging_protocol}. The relative-pose process model and preintegration on $SE_2(3)$ are discussed in Sections \ref{sec:process_model} and \ref{sec:preint}, respectively. Simulation and experimental results are discussed in Sections \ref{sec:sim} and \ref{sec:exp}, respectively, before \tro{further practical considerations are mentioned in \ref{sec:practical}}.

%% file: sections/related_work.tex
\begin{figure}[t!]
    \centering
    \includegraphics[trim={50cm 30cm 30cm 2cm},clip,width=0.7\columnwidth]{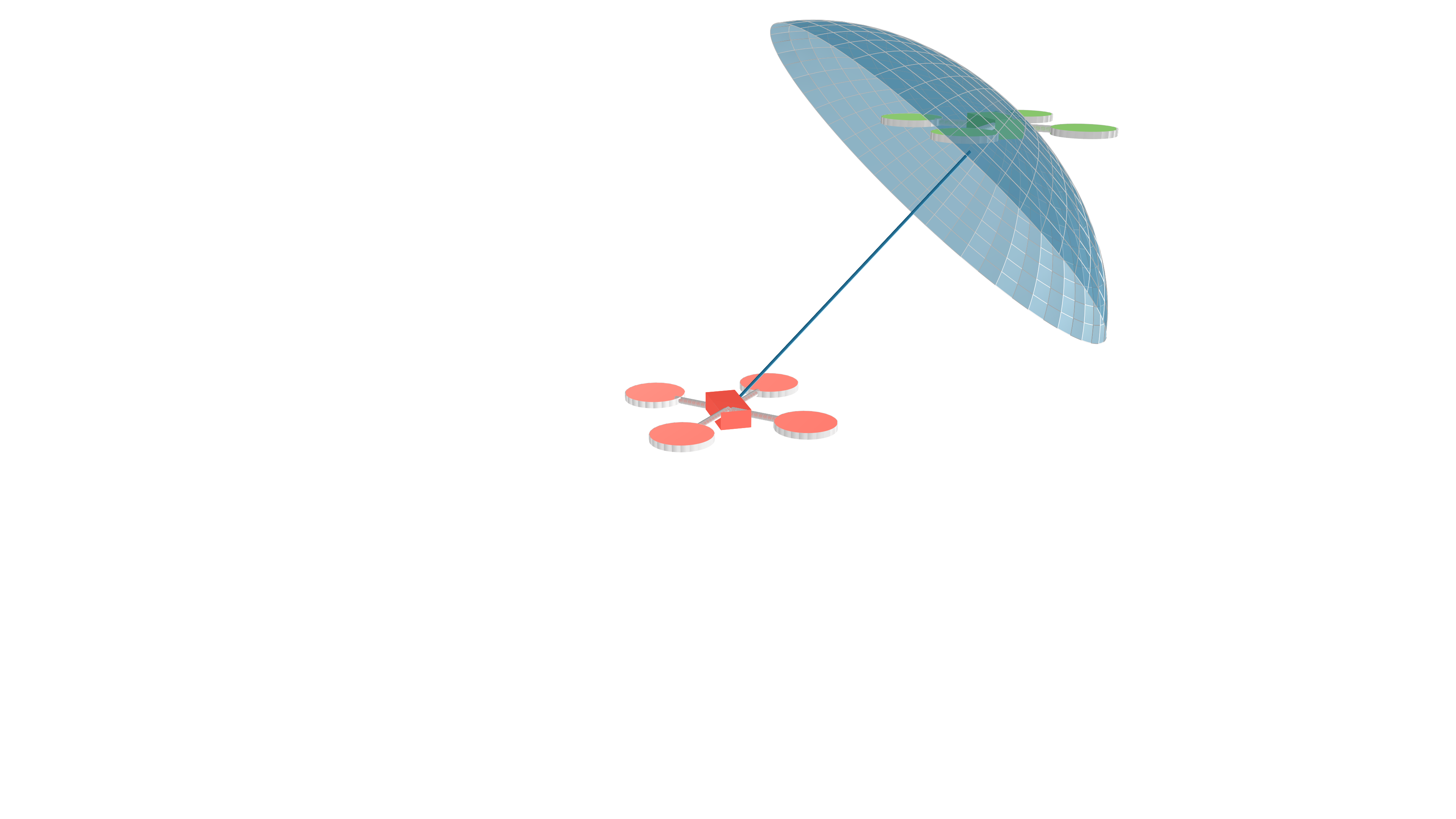}
    \caption{The distribution of the posterior position of the green robot given a position prior and a single range measurement with the red robot.}
    \label{fig:banana}
\end{figure}

The majority of UWB-based localization relies on a set of pre-localized and synchronized 
%group of 
static transceivers, or \emph{anchors}, to localize a mobile transceiver \cite{Kok2015,Mueller2015a,Jung2022}. This typically relies on the anchors ranging with the mobile transceiver using standard ranging protocols such as \emph{two-way ranging} (TWR) or \emph{time-difference-of-arrival} (TDoA) \cite{Neirynck2017,Shalaby2022a}, \cite[Chapter 7.1.4]{groves2013}. More complicated ranging protocols have been proposed in \cite{Horvath2017,Shah2019, Laadung2022} to allow multiple anchors to passively listen-in on messages 
with the mobile transceiver \ms{to localize it}.

Calibrating the clocks and location of anchors is challenging, and \cite{Ledergerber2015,Hamer2018} propose an approach where anchors actively range with one another to synchronize and localize themselves. Meanwhile, a mobile transceiver passively listens to these signals to localize itself using the anchors' estimated clock states and positions. The work in \cite{Zandian2017,Cano2019} extends this by applying a Kalman filter (KF) to the synchronization and localization problem. 
% In \cite{Badawy2021}, a \ms{localization} cost function is proposed that is invariant to the anchors' synchronization error. 
Meanwhile, in \cite{Alanwar2017}, the synchronization approach is accurate to within a few microseconds, whereas nanosecond-level accuracy is desired for localization with cm accuracy. 

Overcoming the need for a fixed infrastructure of anchors, UWB has been used more recently for teams of robots \cite{Nguyen2019, Xu2020a,Jung2021b}. In \cite{Shi2020a}, it is assumed that neighbouring robots know their poses and clock states, thus essentially behaving as mobile anchors, allowing a mobile transceiver to localize itself. The use of robots with multiple transceivers is proposed in \cite{Shalaby2021,Nguyen2018}, and in \cite{Hepp2016a} a robot equipped with 4 transceivers localizes a mobile transceiver relative to itself by having one of the 4 transceivers actively range with the target and the other 3 passively listening. 

In \cite{Shi2020,Dou2022}, a passive-listening-based ranging protocol is proposed where the network is divided into ``parent robots'' that actively range with one another and ``child robots'' that passively listen-in on these measurements. This hierarchical constraint has the limitation that parent robots cannot localize child robots \ms{and do not benefit from passive listening measurements themselves when they are not involved in a ranging transaction}. Additionally, it is suggested that the child robots use the estimated position and clock states of the parent states, which in filtering applications would lead to untracked cross-correlations that would result in poor performance \cite{Shalaby2021b}.

Furthermore, in filtering applications, the problem of communicating IMU measurements to neighbours remains unaddressed. In \cite{Allak2019,Allak2022}, scattering theory is used to send pre-computed matrices \ms{between two robots} rather than individual IMU measurements, in a manner similar to the concept of preintegration \cite{Lupton2012, Forster2017}. However, extending this to \ms{more than two robots} is challenging, particularly for preintegrated poses directly on $SE_2(3)$ \cite{Barrau2019, Brossard2021}. 

\tro{The motive behind using $SE_2(3)$ state representation for} relative pose estimation using range measurements is \tro{due to this being} 
\jln{inherently}
a nonlinear problem, \jln{which is} commonly addressed using particle filtering \cite{Gonzalez2009,Liu2017} 
to \jln{handle} non-ellipsoid-shaped distributions in Cartesian coordinates, 
see Figure \ref{fig:banana}. 
This \ms{nonlinearity} motivates the use of an on-manifold EKF, \tro{such as an EKF with states represented directly on the $SE_2(3)$ manifold,} which can represent such \tro{non-ellipsoid-shaped} distributions 
using exponential coordinates \cite{Long2013}.

%% file: sections/prelims.tex
\subsection{Notation}

Throughout this paper, a bold upper-case letter (e.g., $\mbf{X}$) denotes a matrix, a bold lower-case letter (e.g., $\mbf{x}$) denotes a column matrix, and a right arrow under the letter (e.g., $\underrightarrow{x}$) denotes a physical vector. In a 3-dimensional space, a vector $\underrightarrow{x}$ resolved in a reference frame $i$ is denoted as $\mbf{x}_i \in \mathbb{R}^3$, while the derivative of a vector $\underrightarrow{x}$ with respect to frame $i$ is denoted $^i \hspace{-2pt} \underrightarrow{\dot{x}}$.

%The position of point $z$ relative to point $w$ is denoted using a vector $\underrightarrow{r}^{zw}$. 
% The vector \tro{denoting point $z$ relative to point $w$ is}
%from point $w$ to point $z$ is denoted
% $\underrightarrow{r}^{zw}$.
The vector from point $w$ to point $z$ is denoted $\underrightarrow{r}^{zw}$.
The relative velocity and acceleration between points $z$ and $w$ with respect to frame $i$ is denoted 
\begin{align*}
    \underrightarrow{v}^{zw/i} \triangleq \hspace{4pt} ^i \hspace{-2pt} \underrightarrow{\dot{r}}^{zw}, \qquad \underrightarrow{a}^{zw/i} \triangleq \hspace{4pt} ^i \hspace{-2pt} \underrightarrow{\dot{v}}^{zw/i}.
\end{align*} 
The rotation from a reference frame $j$ to a reference frame $i$ is parametrized using a rotation matrix $\mbf{C}_{ij} \in SO(3)$. Therefore, the relationship between $\mbf{r}_i^{zw}$ and $\mbf{r}_j^{zw}$ is given by $\mbf{r}_i^{zw} \equiv \mbf{C}_{ij} \mbf{r}_j^{zw}$.

% In a 3-dimensional space, the position of point $z$ relative to point $w$ resolved in a reference frame $i$ is denoted as $\mbf{r}_i^{zw} \in \mathbb{R}^3$. \textcolor{red}{Taking the time derivative with respect to reference frame $j$, the corresponding velocity resolved in reference frame $i$ is denoted as $\mbf{v}_i^{zw/j} \in \mathbb{R}^3$. Taking the time derivative of this velocity with respect to reference frame $\ell$, the corresponding acceleration is denoted $\mbf{a}_i^{zw/j/\ell} \in \mathbb{R}^3$. Throughout this paper, a dot above a vector represents a time derivative with respect to the reference frame in which the vector is resolved in; for example, $\mbfdot{r}_i^{zw} = \mbf{v}_i^{zw/i}$.} The rotation from a reference frame $i$ to a reference frame $j$ is parametrized using a rotation matrix $\mbf{C}_{ij} \in SO(3)$. Therefore, the relationship between $\mbf{r}_i^{zw}$ and $\mbf{r}_j^{zw}$ is given by $\mbf{r}_i^{zw} \equiv \mbf{C}_{ij} \mbf{r}_j^{zw}$.

Throughout this paper, $\mbf{1}$ and $\mbf{0}$ denote identity %matrices 
and zero matrices of appropriate dimension. 
%based on their usage. 
When ambiguous, a subscript will indicate the dimension of these matrices.

\subsection{Matrix Lie Groups}

\begin{figure}
    \centering
    \includegraphics[width=\columnwidth]{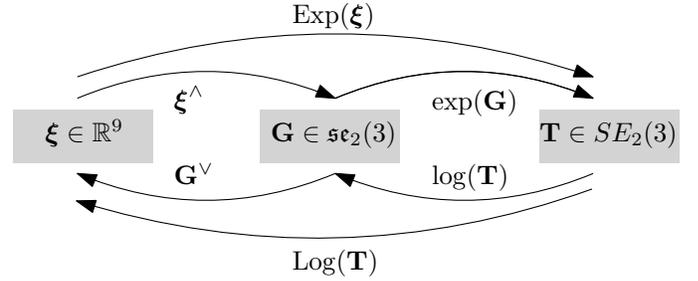}
    \caption{A summary of the operators between elements of the different spaces associated with $SE_2(3)$.}
    \label{fig:lie_project}
\end{figure}

The \emph{pose} of one rigid body relative to another is defined using the relative attitude and position $(\mbf{C}, \mbf{r})$, where all subscripts and superscripts are dropped in this section for conciseness. Meanwhile, the \emph{extended pose} of one rigid body relative to another is defined using the relative attitude, velocity, and position $(\mbf{C}, \mbf{v}, \mbf{r})$. The extended pose can be represented using an extended pose transformation matrix \cite{Barrau2019}
\beq
    \mbf{T} = \bma{ccc} \mbf{C} & \mbf{v} & \mbf{r} \\ & 1 & \\ & & 1 \ema \in SE_2(3), \nonumber
\eeq
where empty spaces represent $\mbf{0}$ entries. The corresponding matrix Lie algebra is denoted as $\mathfrak{se}_2(3) \subset \mathbb{R}^{5 \times 5}$, and elements of \jln{this} Lie algebra \tro{$\mbf{G}~\in~\mathfrak{se}_2(3)$} can also be represented as elements of $\mathbb{R}^9$ \tro{by defining a linear mapping $(\cdot)^\wedge:~\mathbb{R}^9~\to~\mathfrak{se}_2(3)$ such that $$\mbf{G} = \mbs{\xi}^\wedge, \qquad \mbs{\xi} \in \mathbb{R}^9.$$ Similarly, the inverse of $(\cdot)^\wedge$, denoted $(\cdot)^\vee:~\mathfrak{se}_2(3)~\to~\mathbb{R}^9$, is defined such that $\mbs{\xi}~=~\mbf{G}^\vee$.}

\tro{Mapping elements of the matrix Lie algebra $\mathfrak{se}_2(3)$ to the matrix Lie group $SE_2(3)$ is the \emph{exponential map} $\operatorname{exp}:~\mathfrak{se}_2(3)~\to~SE_2(3)$, thus yielding the definition $$\mbf{T} = \operatorname{exp}(\mbs{\xi}^\wedge) \triangleq \operatorname{Exp}(\mbs{\xi}),$$ where $\operatorname{Exp}:~\mathbb{R}^9~\to~SE_2(3)$ is defined for conciseness. The inverse of the exponential map is the \emph{logarithmic map} $\operatorname{log}:~SE_2(3)~\to~\mathfrak{se}_2(3)$, yielding the definition $$\mbs{\xi} = \operatorname{log}(\mbf{T})^\vee \triangleq \operatorname{Log}(\mbf{T}),$$ where $\operatorname{Log}:~SE_2(3)~\to~\mathbb{R}^9$ is defined for conciseness.} 

\tro{Owing to the fact that $SE_2(3)$ is a matrix Lie group, the exponential map and logarithmic map are the same as the matrix exponential and the matrix logarithm, respectively.} The operators between elements of the different spaces are summarized in Figure \ref{fig:lie_project}, and their expression is found in \cite{Barrau2019,Brossard2021},\cite[Chapter 9]{Barfoot2022}.
Two other useful operators on $SE_2(3)$ are the \emph{Adjoint} matrix $\operatorname{Ad}: SE_2(3) \rightarrow \mathbb{R}^{9\times 9}$ defined by
\beq
    \operatorname{Exp}(\operatorname{Ad}(\mbf{T})\mbs{\xi}) \triangleq \mbf{T}\operatorname{Exp}(\mbs{\xi})\mbf{T}^{-1},  \nonumber
\eeq
and the \emph{odot} operator $(\cdot)^\odot: \mathbb{R}^5 \rightarrow \mathbb{R}^{5 \times 9}$ defined such that 
\beq 
    \label{eq:odot}
    \mbf{p}^{\odot} \mbs{\xi} \triangleq \mbs{\xi}^\wedge \mbf{p}
\eeq
for \jln{any} vector $\mbf{p} \in \mathbb{R}^5$. 

\tro{The non-commutativity of matrix multiplication means that} matrix Lie group elements can be perturbed from the left
% \beq
    $\mbf{T} = \operatorname{Exp}(\delta \mbs{\xi}) \mbfbar{T}$ %\nonumber
% \eeq
or the right 
% \beq
    $\mbf{T} = \mbfbar{T} \operatorname{Exp}(\delta \mbs{\xi})$, %\nonumber
% \eeq
where the overbar denotes a nominal value. Additionally, the first-order approximation
\beq
    \label{eq:exp_approx}
    \operatorname{Exp}(\mbsdel{\xi}) \approx \mbf{1} + \mbsdel{\xi}^\wedge
\eeq
will often be used when linearizing nonlinear models.

\subsection{UWB Ranging and Clocks} \label{sec:prelim_uwb}

% \begin{figure}[h]
%     \centering
%     \begin{minipage}{0.45\columnwidth}%
%         \subfloat[Subfigure 1 list of figures text][Visualizing different clock rates.]{
%         \includegraphics[width=\columnwidth]{figs/ranging_protocol/viz_skew.eps}
%         \label{fig:viz_skew}}
%     \end{minipage}%
%     \qquad
%     \begin{minipage}{0.43\columnwidth}%
%         \subfloat[Subfigure 2 list of figures text][The standard DS-TWR protocol.]{
%         \includegraphics[width=\columnwidth]{figs/ranging_protocol/basic_twr.eps}
%         \label{fig:basic_twr}}
%     \end{minipage}
%     \caption{Schematics representing the passage of time and message timestamping in the clocks of Transceivers $f$ and $s$.}
%     \label{fig:twr}
% \end{figure}

UWB ranging between two transceivers relies on ToF measurements, which \jln{are deduced from}
timestamps recorded by a clock on each transceiver. \jln{However, } these clocks are %typically 
unsynchronized. % and have varying clock rates. %To visualize this, 
%Consider two transceivers denoted $p$ and $q$ equipped with two different clocks. 
%The time recorded by either clock at any instant will be different. 
%This difference is referred to as the \emph{clock offset}, and is denoted using $\tau_{pq}$, where
\jln{Denoting $t_i(t)$ as the time $t$ resolved in Transceiver $i$'s clock gives}
%for any two transceivers $p$ and $q$
\beq    \label{eq: offset general definition}
    %t_p(t) = t_q(t) + \tau_{pq}(t), 
    t_i(t) = t + \tau_{i}(t),
\eeq
\jln{where $\tau_{i}(t)$ defines the (time-varying) \emph{offset} of clock $i$.}
%\jln{where $\tau_{pq}(t)$ defines the \emph{clock offset} between $p$ and $q$ and is itself time-varying,
%due to different clock rates.}
%and $t_i$ is the time $t$ resolved in Transceiver $i$'s clock. 
%The rate of change of the clock offset, representing the difference in the clock rates, 
%is referred to as the \emph{clock skew} between the transceivers and is denoted $\gamma_{pq}$.

To obtain a range measurement, the two transceivers transmit and timestamp \jln{a sequence of} 
messages among themselves as shown between Robot 1 and Robot 2 in Figure \ref{fig:ds_twr}. 
A time instance corresponding to the $i^\text{th}$ message is denoted as $\mathtt{T}^{i}$ 
for the transmission time and $\mathtt{R}^{i}$ for the reception time, while a subscript $j$ 
denotes the time instance as timestamped by Transceiver $j$. For example, $\mathtt{T}^1_{f_0}$ 
is the timestamp corresponding to the first message transmission as recorded by Transceiver $f_0$. 
The protocol example shown between Robot 1 and Robot 2 in Figure \ref{fig:ds_twr} 
is a \jln{modified} version of the \jln{standard} double-sided two-way ranging 
(DS-TWR \jln{\cite{WPANcommittee:standar03:WPAN}}) 
protocol as presented in \cite{Shalaby2022a}, where the message shown in red represents 
an ``information message'' used to broadcast the timestamps recorded by Robot 1. 

%% file: sections/problem_formulation.tex
\begin{figure}
    \centering
    \includegraphics[width=\columnwidth]{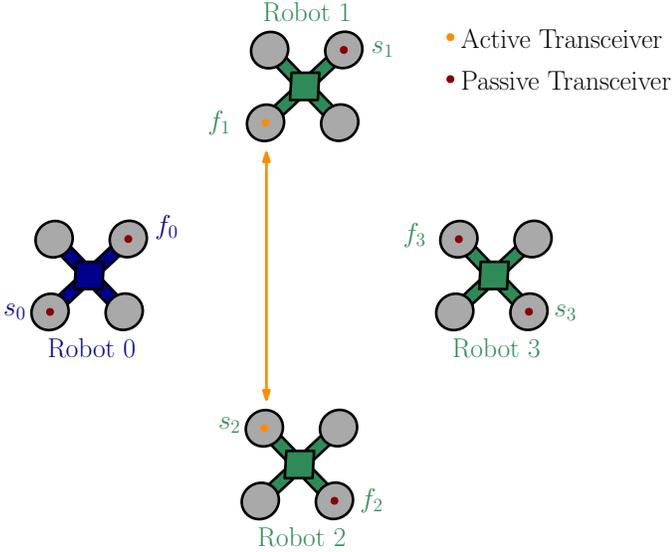}
    \caption{An example of a ranging transaction, where Transceivers $f_1$ and $s_2$ are actively ranging with one another and all other tags are passively listening.}
    \label{fig:ranging}
\end{figure}

Consider a scenario with $n+1$ robots, \jln{as shown in Figure \ref{fig:ranging} for $n=3$}. 
Throughout this paper, the perspective of one robot 
is considered, denoted without loss of generality Robot 0, as any of the $n+1$ robots can 
be considered Robot 0. 
Neighbouring robots are then referred to as Robot $i$, $i \in \{1, \ldots, n\}$. 
%An example with $n=3$ is shown in Figure \ref{fig:ranging}. 
This paper employs a ``robocentric'' viewpoint of the relative-pose state estimation problem, 
where all states are estimated relative to Robot 0 and are resolved in the body frame of that robot. 
The robots are \jln{assumed to be} rigid bodies, so any vector can be resolved 
in one of the following $n+2$ reference frames:
\begin{itemize}
    \item an (absolute) inertial frame denoted with a subscript $a$,
    \item Robot 0's body frame denoted with a subscript 0, or
    \item neighbouring Robot $i$'s body frame denoted with a subscript $i$.
\end{itemize}

Each robot is equipped with an IMU at its center, 
consisting of a 3-axis gyroscope and accelerometer. Given the use of accelerometers, 
the relative pose estimation problem involves estimating the extended pose 
of each neighbouring robot relative to Robot 0 in Robot 0's body frame. 
The extended pose of Robot $i$ is then \jln{defined} as
\beq
    \mbf{T}_{0i} = 
    \bma{ccc} 
        \mbf{C}_{0i} & \mbf{v}_0^{i0/a} & \mbf{r}_0^{i0} \\
        & 1 & \\
        & & 1
    \ema \in SE_2(3), \quad i \in \{1, \ldots, n\}, \nonumber
\eeq
where time dependence is omitted from the notation for conciseness. 
The dependence on the absolute frame $a$ is also omitted 
from the notation $\mbf{T}_{0i}$, %for the extended relative pose $\mbf{T}_{0i}$. 
%However, throughout this paper 
\jln{with the convention that}
all extended relative pose matrices in the paper are of this form, 
where the \jln{vector corresponding to the}
second component in the first row is the derivative 
with respect to the \emph{absolute} frame of the 
\jln{vector corresponding to} the third component,
\jln{irrespective of the fact that these vectors are resolved in
frame $0$.}
%third component with respect to the \emph{absolute} frame.

Each robot is also equipped with 2 UWB transceivers for relative pose observability \cite{Shalaby2021}. 
The \emph{first} and \emph{second} transceivers on Robot $j$ are denoted $f_j$ and $s_j$, 
respectively, for $j \in \{0, \ldots, n\}$. It is assumed that the vector coordinates $\mbf{r}_j^{f_j j}$ 
and $\mbf{r}_j^{s_j j}$ between the transceivers and the IMU on Robot $j$ are known, since they 
can be measured by hand or more accurately using a motion capture system. 

Denote the set of all transceivers as $\mc{C} = \{ f_0, \ldots, f_n, s_0, \ldots, s_n \}$. 
Consider the state of the clock on Transceiver $i \in \mc{C}$ relative to real time.
%which is \jln{represented} as yet another clock $r$. 
The evolution of the \jln{offset} $\tau_i(t)$ of clock $i$ is modelled using a third-order model 
in \cite{Hamer2018}. However,\cite{Cano2019} shows that a second-order model of the form
\beq
    \label{eq:clock_dynamics}
    \underbrace{\bma{c} \dot{\tau}_{i} \\ \dot{\gamma}_{i} \ema}_{\mbfdot{c}_{i}} = \underbrace{\bma{cc} 0 & 1 \\ 0 & 0 \ema}_{\mbf{A}} \underbrace{\bma{c} \tau_{i} \\ \gamma_{i} \ema}_{\mbf{c}_{i}} + \mbf{w}_{i}
\eeq
is sufficient for localization purposes, \jln{where $\gamma_{i}(t)$ is called the
\emph{clock skew}}, $\mbf{w}_{i}$ is a continuous-time zero-mean white Gaussian process noise 
with $\mathbb{E}[\mbf{w}_{i}(t_1)\mbf{w}_{i}(t_2)^\trans] = \mbc{Q}\delta(t_1-t_2)$,
\beq
    \mbc{Q} = \bma{cc} \mc{Q}^\tau & \\ & \mc{Q}^\gamma \ema, \nonumber
\eeq
$\delta(\cdot)$ is the \emph{Dirac's delta function}, and $\mc{Q}^\tau$ and $\mc{Q}^\gamma$ are the clock offset and skew process-noise \emph{power spectral densities}, respectively.

A robocentric viewpoint is also maintained for the clock states, where offsets and skews 
of all clocks are estimated relative to the clock of Transceiver $f_0$ on Robot 0. 
The clock state of Transceiver $s_0$ is then 
\beq
    \mbf{c}_{s_0 f_0} = \bma{c} \tau_{s_0 f_0} \\ \gamma_{s_0 f_0} \ema 
    \jln{\triangleq \bma{c} \tau_{s_0} - \tau_{f_0} \\ \gamma_{s_0} - \gamma_{f_0} \ema}
    \in \mathbb{R}^2, \nonumber
\eeq
while the clock state of neighbouring Robot $i$ is given by
\beq
    \ms{\mbc{X}_{i0}^{\text{c}} \triangleq} \left( \mbf{c}_{f_i f_0}, \mbf{c}_{s_i f_0} \right) \in \mathbb{R}^2 \times \mathbb{R}^2, \quad i \in \{1, \ldots, n\}, \nonumber
\eeq
where, as before, time dependence is omitted from the notation for conciseness. 
The full relative state estimate of Robot $i$ is then given by
\beq
    \mbc{X}_{i0} \triangleq (\mbf{T}_{0i}, \ms{\mbc{X}_{i0}^{\text{c}}}) \in SE_2(3) \times \mathbb{R}^2 \times \mathbb{R}^2, \nonumber
\eeq
and the full state estimated by Robot 0 is
\beq
    \mbc{X} \triangleq (\mbf{c}_{s_0 f_0}, \mbc{X}_{10}, \ldots, \mbc{X}_{n0}) \in 
    \mathbb{R}^2 \times \left(SE_2(3) \times \mathbb{R}^2 \times \mathbb{R}^2 \right)^n. \nonumber
\eeq

Communication constraints limit Robot 0's ability to estimate the state $\mbc{X}$, 
since to prevent \tro{message collision} only one pair of transceivers can communicate at a time.
\jln{As the number of transceivers increases}, this can result in poor scalability 
due to longer wait times between successive ranging measurements by a given pair.
%, as each pair of transceivers has to wait for all other pairs in the sequence of communicating pairs. 
Additionally, the rate at which transceivers communicate is typically lower 
than the rate at which IMU measurements are recorded at neighbouring robots, 
thus Robot 0 cannot collect the IMU measurements from all its neighbours 
without significant \ms{and impractical} communication overhead. Therefore, part of the problem 
is to design a scalable and practical ranging protocol that accommodates these 
communication constraints.

This paper presents an on-manifold extended Kalman filter (EKF) for estimating 
the state $\mbc{X}$ using a novel ranging protocol that allows all robots 
to listen-in on neighbours while awaiting their turn to communicate. 
%For the observability of the filter, 
It is known from \cite{Shalaby2021, Cossette2021} 
that the relative pose states are \emph{observable} from IMU and range measurements. \tro{In particular, the use of two transceivers per robot ensures the observability of the relative poses while overcoming the need for \emph{persistent excitation} or constant relative motion between the robots \cite{Shalaby2021}. This benefit comes at the added cost of an additional transceiver. Nonetheless, UWB transceivers are typically compact, lightweight, low-power, and inexpensive. In fact, the ones used in this paper as shown in Figure \ref{fig:exp_setup}, are 32 mm $\times$ 49 mm in size and weigh approximately 8 g each.} \tro{Meanwhile, assuming that the relative pose states are known since they are observable from the IMU and range measurements alone, clock-offset measurements are sufficient to ensure observability of the clock states \cite{Cano2019}}.
% and it will be shown that the clock %offset  \ms{states} are measured directly using the proposed ranging protocol and can thus be estimated. % The process model of the filter is then derived while passive listening is utilized to broadcast preintegrated IMU measurements for efficient communication between robots. 

\tro{To simplify the analysis in this paper, a complete communication graph with no packet drops or failures between the robots is assumed, which reduces the scalability of the system. Another factor impacting the scalability of the system is that} the size of the state $\mbc{X}$ increases with $n$; therefore, the number of robots that can be included in Robot 0's EKF is limited by Robot 0's computational capabilities. This paper addresses scenarios where $n$ is limited to a few robots. \tro{The complications associated with larger systems and incomplete and dynamic communication graphs are discussed in Section \ref{subsec:dynamic_graphs}.}

%However, a potential extension of this work for larger teams would be to have each robot only estimate the relative states of $m < n$ neighbouring robots.

%% file: sections/ranging_protocol.tex
\begin{figure}
    \centering
    \includegraphics[width=0.9\columnwidth]{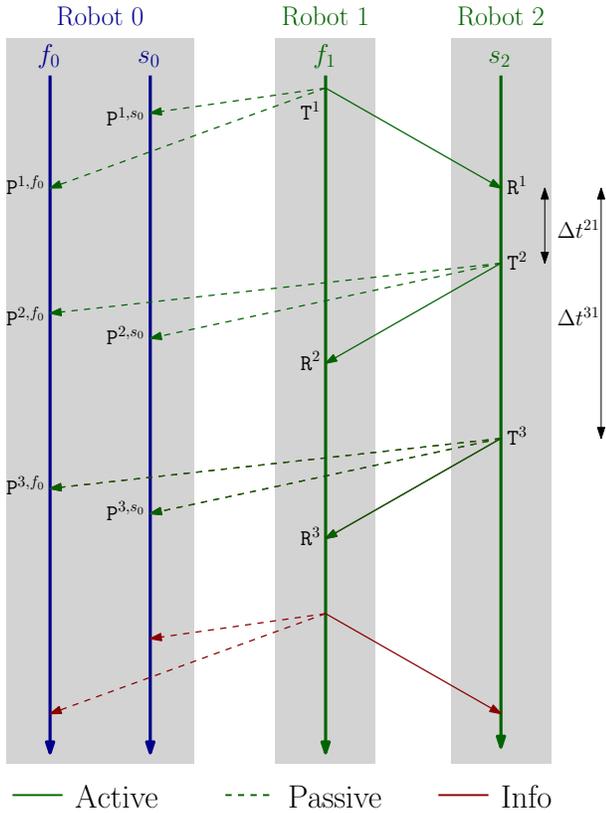}
    \caption{%Schematic showing the 
    Proposed ranging protocol when Transceiver $f_1$ is initiating a DS-TWR ranging transaction with Transceiver $s_2$. This paper proposes that all other transceivers listen in on these \jln{messages}. Shown here are Transceivers $f_0$ and $s_0$ on Robot~0 passively listening, where the time instance corresponding to the $i^\text{th}$ passive reception at Transceiver $j$ is denoted $\mathtt{P}^{i,j}$.}
    \label{fig:ds_twr}
\end{figure}

\subsection{Overview}

To address the communication constraints, a ranging protocol is proposed that involves performing \ms{DS-TWR} between all pairs of transceivers not on the same robot in sequence while leveraging passive listening measurements at all other transceivers \jln{that are} not actively ranging. This is shown in Figure \ref{fig:ds_twr} for an example where Transceiver $f_1$ is initiating a TWR \jln{transaction} with Transceiver $s_2$, and Transceivers $f_0$ and $s_0$ are passively listening. In the proposed ranging protocol, any of the $2(n+1)$ transceivers can initiate a TWR transaction with any of the $2n$ transceivers not on the same robot. 
%as the initiating one. 
In this section, the passive listening measurements are utilized in the relative-pose state estimator as a source of ranging information between the different robots. This is possible due to the tightly-coupled nature of the proposed estimator, which performs both clock synchronization and relative pose estimation, meaning that clock-offset-corrupted passive listening measurements can still be used to correct relative pose states, as cross-correlation information is available between clock states and relative pose states at all times. \tro{There are multiple advantages to passive listening in multi-robot pose estimation applications, including the availability of more measurements for state correction, providing the robots with information-broadcasting ability, and allowing the implementation of simple MAC protocols. }

\begin{algorithm}[h!]
    \caption[The algorithm for the proposed ranging protocol when Transceiver $f_1$ is initiating the ranging transaction with Transceiver $s_2$, and Transceivers $f_0$, $s_0$, $s_1$, and $f_2$ are passively listening.]{\tro{The algorithm for the proposed ranging protocol when Transceiver $f_1$ is initiating the ranging transaction with Transceiver $s_2$, and Transceivers $f_0$, $s_0$, $s_1$, and $f_2$ are passively listening. Note that in this algorithm, the color} \textcolor{blue}{blue} \tro{is reserved for passive listening measurements that are available to the reference robot Robot 0, and the color} \textcolor{teal}{teal} \tro{is reserved for passive listening measurements at other robots that are not available to Robot 0. \vspace{5pt}}} 
    \label{alg:ranging_protocol}
    \begin{algorithmic}[1]
        \renewcommand{\algorithmicrequire}{}
        \renewcommand{\algorithmicensure}{}
        \REQUIRE \vspace{5pt} \textbf{Initiating Robot (Robot 1)}: The initiating robot has an active transceiver, \textcolor{black}{$f_1$}, and a passive transceiver, \textcolor{teal}{$s_1$}. \vspace{3pt}
        % \ENSURE 
        \STATE \textcolor{black}{Transmit message to $s_2$, and timestamp $\mathtt{T}^1$ in own clock.}
        \STATE \textcolor{teal}{Timestamp passive reception $\mathtt{P}^{1,s_1}$ in own clock, for message transmitted by $f_1$.}
        \STATE \textcolor{black}{Timestamp reception $\mathtt{R}^2$ in own clock, for message transmitted by $s_2$.}
        \\ \textcolor{teal}{Timestamp passive reception $\mathtt{P}^{2,s_1}$ in own clock, for message transmitted by $s_2$.}
        \STATE \textcolor{black}{Timestamp reception $\mathtt{R}^3$ in own clock and read $\mathtt{R}^1$, $\mathtt{T}^2$, $\mathtt{T}^3$, for message transmitted by $s_2$.}
        \\ \textcolor{teal}{Timestamp passive reception $\mathtt{P}^{3,s_1}$ in own clock, for message transmitted by $s_2$.}
        \STATE \textcolor{black}{Transmit info message with $\mathtt{T}^1$, $\mathtt{R}^2$, $\mathtt{R}^3$ to $s_2$.}
        \\\hspace{-15pt}\dotfill    
    \end{algorithmic}
    \begin{algorithmic}[1]
        \renewcommand{\algorithmicrequire}{}
        \renewcommand{\algorithmicensure}{}
        \REQUIRE \textbf{Target Robot (Robot 2)}: The target robot has an active transceiver, \textcolor{black}{$s_2$}, and a passive transceiver, \textcolor{teal}{$f_2$}. \vspace{3pt}
        % \ENSURE 
        \STATE \textcolor{black}{Timestamp reception $\mathtt{R}^1$ in own clock, for message transmitted by $f_1$.}
        \\ \textcolor{teal}{Timestamp passive reception $\mathtt{P}^{1,f_2}$ in own clock, for message transmitted by $f_1$.}
        % \STATE \textcolor{black}{Wait $\Delta t^{21}$ seconds in own clock.}        
        \STATE \textcolor{black}{Transmit message to $f_1$, and timestamp $\mathtt{T}^2$ in own clock.}
        \STATE \textcolor{teal}{Timestamp passive reception $\mathtt{P}^{2,f_2}$ in own clock, for message transmitted by $s_2$.}
        \STATE \textcolor{black}{Set $\mathtt{T}^3 = \mathtt{R}^1 + \Delta t^{31}$, and wait until own clock is $\mathtt{T}^3$.}
        \STATE \textcolor{black}{Transmit message with $\mathtt{R}^1$, $\mathtt{T}^2$, $\mathtt{T}^3$ to $f_1$.}
        \STATE \textcolor{teal}{Timestamp passive reception $\mathtt{P}^{3,f_2}$ in own clock, for message transmitted by $s_2$.}
        \STATE \textcolor{black}{Read $\mathtt{T}^1$, $\mathtt{R}^2$, $\mathtt{R}^3$ from info message transmitted by $f_1$.}
        \\\hspace{-15pt}\dotfill
    \end{algorithmic}
    % \vspace{10pt}
    \begin{algorithmic}[1]
        \renewcommand{\algorithmicrequire}{}
        \renewcommand{\algorithmicensure}{}
        \REQUIRE \textbf{Passive Robot (Robot 0)}: The passive robot has two passive transceivers, \textcolor{blue}{$f_0$} and \textcolor{blue}{$s_0$}. \vspace{3pt}
        % \ENSURE 
        \STATE \textcolor{blue}{Timestamp passive reception $\mathtt{P}^{1,f_0}$ in own clock, for message transmitted by $f_1$.}
        \\ \textcolor{blue}{Timestamp passive reception $\mathtt{P}^{1,s_0}$ in own clock, for message transmitted by $f_1$.}
        \STATE \textcolor{blue}{Timestamp passive reception $\mathtt{P}^{2,f_0}$ in own clock, for message transmitted by $s_2$.}
        \\ \textcolor{blue}{Timestamp passive reception $\mathtt{P}^{2,s_0}$ in own clock, for message transmitted by $s_2$.}
        \STATE \textcolor{blue}{Timestamp passive reception $\mathtt{P}^{3,f_0}$ in own clock and read $\mathtt{R}^1$, $\mathtt{T}^2$, $\mathtt{T}^3$, for message transmitted by $s_2$.}
        \\ \textcolor{blue}{Timestamp passive reception $\mathtt{P}^{3,s_0}$ in own clock, for message transmitted by $s_2$.}
        \STATE Read $\mathtt{T}^1$, $\mathtt{R}^2$, $\mathtt{R}^3$ from info message transmitted by $f_1$.
    \end{algorithmic}
\end{algorithm}

The remainder of this section analyzes how the proposed ranging protocol can be used in a CSRPE. The particular scenario under study is the one shown in Figures \ref{fig:ranging} and \ref{fig:ds_twr}, where transceivers on two neighbouring robots are the ones actively ranging. This is the most general case, and scenarios where one of the transceivers on Robot~0 is actively ranging \jln{involve} similar \jln{but simpler} derivations. %that are in fact slightly less involved. 

\subsection{\tro{The Protocol}}

\tro{The ranging protocol proposed in this paper involves two transceivers actively ranging with one another, while all other transceivers passively listen-in on the messages. The actively ranging transceivers perform DS-TWR, as presented in \cite{Shalaby2022a}. The example shown in Figure \ref{fig:ds_twr} is an example where Transceiver $f_1$ on Robot 1 initiates a transaction with Transceiver $s_2$ on Robot 2. Both robots have another transceiver, $s_1$ and $f_2$ for Robot 1 and Robot 2, respectively, which then passively listen to all the messages transmitted between the active transceivers. Additionally, all other robots have both their transceivers passively listen to all the messages. For example, Robot 0 records passive listening measurements at both Transceivers $f_0$ and $s_0$.}

\tro{When the transceivers transmit and receive messages, whether actively or passively, the transceivers timestamp the time of transmission or reception. Each robot needs access to neighbouring robots' timestamp measurements in order to be able to compute range measurements from the transaction. For example, Robot 0 needs access to the timestamps recorded by Transceivers $f_1$ and $s_2$.} As shown in Figure \ref{fig:ds_twr}, all timestamps are \tro{made} available at Robot~0 \jln{at the end of
the transaction} \tro{by communicating all} the timestamps recorded at Robot~1 in a 
final information message shown in red, and the timestamps recorded at Robot~2 are communicated 
in the last message transmitted by Robot~2. \tro{Note, however, that passive listening measurements recorded by the other transceivers on Robot 1, Robot 2, and any other neighbouring robots are not made available to Robot 0.} \tro{The ranging protocol is outlined in Algorithm~\ref{alg:ranging_protocol} for the scenario shown in Figure~\ref{fig:ds_twr}.}

When implementing the ranging protocol, a choice has to be made on the receiving robot's 
side (in this case, Robot~2) for the delays $\Delta t^{21} \triangleq \mathtt{T}^{2} - \mathtt{R}^{1}$ and 
$\Delta t^{31} \triangleq \mathtt{T}^{3} - \mathtt{R}^{1}$. 
% Notice that these values can be %recorded 
% \jln{resolved}
% in different clocks; for example, in clock $f_0$, 
% $\Delta t^{21}_{f_0} = \mathtt{T}^{2}_{f_0} - \mathtt{R}^{1}_{f_0}$. 
These user-defined parameters affect the frequency and noise of the measurements, 
and can be chosen based on \cite{Shalaby2022b}. 
Note that $\Delta t^{32} \triangleq \Delta t^{31} - \Delta t^{21}$. %, 
%and \jln{use a subscript to refer to these quantities resolved in a specific clock, e.g.,
%$\Delta t^{21}_{f_0} \coloneqq \mathtt{T}^{2}_{f_0} - \mathtt{R}^{1}_{f_0}$}. 
Additionally, it will be assumed throughout this paper that the distances between 
transceivers and the clock skews remain constant during one ranging transaction. 
These are \jln{good} approximations for most robotic applications with %slowly-varying 
\jln{typical} clock rate \jln{variations} \cite[Chapter~7.1.4]{groves2013}, \cite{Shalaby2022b}. 

\tro{
The proposed ranging protocol has the following advantages. Given that all transceivers passively listen to neighbouring robots communicating, this proposed protocol gives robots the ability to broadcast information such as IMU measurements or estimated maps at a higher rate as any robot can obtain information communicated between two neighbouring robots. This feature will be utilized for multi-robot preintegration in Section \ref{sec:preint}. Additionally, given that each robot knows which robots are currently ranging, a simple MAC protocol can be implemented to prevent message collision between multiple robots attempting to transmit messages at the same time. %schedule which pair will range next. 
    %Passive listening allows an algorithm that combines the customizability of TDMA and the robustness of the \emph{token-passing protocol} \cite[Chapter 3.3]{miao2016}. 
    % Given that \jln{each} robot %now have access to the knowledge of 
    % \jln{knows}
    % 
    To do so, %which robots are currently ranging, 
    a user-defined sequence of ranging pairs can be made known to all robots. Each robot can then keep track of which pair in the sequence is currently ranging, and initiate a TWR transaction to a specified transceiver when it is its turn to do so. This MAC protocol is named here the \emph{common-list protocol}. %, and has the advantage over the token-passing protocol that it allows more customizability on the sequence of ranging pairs without having to allocate discrete-time windows like in TDMA.
}

\subsection{\jln{Modelling} Timestamp \jln{Measurements}}

%The timestamps 
\jln{The time instances}
shown in Figure \ref{fig:ds_twr} are only available to the robots as noisy 
%measurements 
\jln{timestamps}
and in the clocks of the transceivers rather than in the global common time. 
Therefore, %these measurements 
\jln{the timestamp measurements}
are affected by clock offsets, clock skews, and white noise. 
Modelling these effects, 
%\jln{with the offset definition in \eqref{eq: offset general definition}},
the timestamps available at Robot~1 (hereinafter, the \emph{initiating robot}) are of the form
\begin{align}
    \tilde{\mathtt{T}}_{f_1}^{1} &= \mathtt{T}^{1} + \tau_{f_1}(\mathtt{T}^1) + \eta_{f_1}^{1}, \label{eq:ts1}\\
    \tilde{\mathtt{R}}_{f_1}^{2} &= \mathtt{T}^{1} + \f{2}{c}d^{s_2 f_1} + \Delta t^{21} + \tau_{f_1}(\mathtt{R}^2) +  \eta_{f_1}^{2}, \label{eq:ts2} \\
    \tilde{\mathtt{R}}_{f_1}^{3} &= \mathtt{T}^{1} + \f{2}{c}d^{s_2 f_1} + \Delta t^{31} + \tau_{f_1}(\mathtt{R}^3) + \eta_{f_1}^{3}, \label{eq:ts3}
\end{align}
where $\tilde{(\cdot)}$ here denotes a measured value, 
%\jln{$\tau_{f_1 f_0}$ is the clock offset at time $T^1$},
$d^{s_2f_1}$ is the distance between Transceivers $s_2$ and $f_1$,
and $\eta_i^{\ell}$ is the random noise on the $\ell^\text{th}$ measurement of Transceiver $i$.
\jln{All the random noise variables on timestamps are assumed to be independent, zero-mean and with the 
same variance $\sigma^2$.}
%and $\eta_i^{\ell} \sim \mc{N} (0, \sigma^2)$ is white Gaussian measurement noise associated with the %$\ell^\text{th}$ measurement of Transceiver $i$. 
%
%Note that errors due to skew are considered during the delays $\Delta t^{21}$ and $\Delta t^{31}$ 
%but not during the time of flight $\f{1}{c}d$, since $\Delta t^{21}, \Delta t^{31} \gg \f{1}{c}d$.
%\jln{In addition, \eqref{eq:ts2}-\eqref{eq:ts3} use the approximations 
%$\gamma_{f_1 f_0} \Delta t^{21} \approx \gamma_{f_1 f_0} \Delta t^{21}_{f_0}$
%and $\gamma_{f_1 f_0} \Delta t^{31} \approx \gamma_{f_1 f_0} \Delta t^{31}_{f_0}$,
%which are justified in Appendix \ref{appx:approx_delta_t}.}

Similarly, the measurements available at Robot~2 (hereinafter, the \emph{target robot}) are of the form
\begin{align}
    \tilde{\mathtt{R}}_{s_2}^{1} &= \mathtt{T}^{1} + \f{1}{c}d^{s_2 f_1} + \tau_{s_2}(\mathtt{R}^1) + \eta_{s_2}^{1}, \label{eq:ts4} \\
    \tilde{\mathtt{T}}_{s_2}^{2} &= \mathtt{T}^{1} + \f{1}{c}d^{s_2 f_1} + \Delta t^{21} + \tau_{s_2}(\mathtt{T}^2) + \eta_{s_2}^{2}, \label{eq:ts5}\\
    \tilde{\mathtt{T}}_{s_2}^{3} &= \mathtt{T}^{1} + \f{1}{c}d^{s_2 f_1} + \Delta t^{31} + \tau_{s_2}(\mathtt{T}^3) +  \eta_{s_2}^{3}. \label{eq:ts6}
\end{align}

The timestamp measurements \eqref{eq:ts1}-\eqref{eq:ts6} correspond to the standard DS-TWR protocol, 
from which ToF pseudomeasurements can be generated. Nonetheless, additional measurements are available 
at Robot~0 (hereinafter, the \emph{passive robot}) since its transceivers $f_0$ and $s_0$ also receive 
the messages exchanged between the two actively ranging robots. 
This yields the following additional timestamp measurements at Robot~0,
\begin{align}
    \tilde{\mathtt{P}}_{i}^{1,i} &= \mathtt{T}^{1} + \f{1}{c}d^{f_1 i} + \tau_{i}(\mathtt{P}^{1,i}) + \eta_{i}^{1}, \label{eq:ts7} \\
    \tilde{\mathtt{P}}_{i}^{2,i} &= \mathtt{T}^{1} + \f{1}{c}d^{s_2 f_1} + \f{1}{c}d^{s_2 i} + \Delta t^{21} + \tau_{i}(\mathtt{P}^{2,i}) + \eta_{i}^{2}, \label{eq:ts8} \\
    \tilde{\mathtt{P}}_{i}^{3,i} &= \mathtt{T}^{1} + \f{1}{c}d^{s_2 f_1} + \f{1}{c}d^{s_2 i} + \Delta t^{31} + \tau_{i}(\mathtt{P}^{3,i}) + \eta_{i}^{3}, \label{eq:ts9}
\end{align}
where $i \in \{ f_0, s_0 \}$. % and $\tau_{f_0 f_0} = \gamma_{f_0 f_0} = 0$. 
Similarly, each neighbouring robot not involved in the ranging transaction records 
\jln{its} own passive listening measurements at its \jln{two} transceivers. 
However, these are not shared with other robots as this would require each robot 
to take its turn transmitting a message.

In the case where Robot~0 is not involved in the ranging transaction and just listens in passively, there are 12 available timestamp measurements at Robot~0, 6 sent by neighbouring robots, and 3 passive-listening timestamps per transceiver on Robot~0. However, when one of the transceivers $f_0$ or $s_0$ is involved in the ranging transaction, only 9 timestamp measurements are available. 
%
% Additionally, in the case where Transceiver $f_0$ is an active transceiver, the equations \eqref{eq:ts1}-\eqref{eq:ts6} are simpler as the clock states are being modelled relative to this transceiver. For example, if Transceiver $f_0$ initiates, the timestamp measurements \eqref{eq:ts1}-\eqref{eq:ts3} \jln{become}
% \begin{align*}
%     \tilde{\mathtt{T}}_{f_0}^{1} &= \mathtt{T}_{f_0}^{1} + \eta_{f_0}^{1},\\
%     \tilde{\mathtt{R}}_{f_0}^{2} &= \mathtt{T}_{f_0}^{1} + \f{2}{c}d^{s_2 f_0} + \Delta t_{f_0}^{21} + \eta_{f_0}^{2}, \\
%     \tilde{\mathtt{R}}_{f_0}^{3} &= \mathtt{T}_{f_0}^{1} + \f{2}{c}d^{s_2 f_0} + \Delta t_{f_0}^{31} + \eta_{f_0}^{3},
% \end{align*}
% where clock states no longer appear.

\subsection{Pseudomeasurements as a Function of the State}

% \begin{figure}
%     \centering
%     \includegraphics[width=0.3\columnwidth]{figs/ranging_protocol/known_vector.eps}
%     \caption{An example of the position of a transceiver relative to the IMU of Robot~0. Throughout this paper, it is assumed that the IMU is at the center of the robot.}
%     \label{fig:known_vector}
% \end{figure}

To use the timestamp measurements \eqref{eq:ts1}-\eqref{eq:ts9} in the CSRPE, they must be rewritten as a function of the state being estimated. In this subection, pseudomeasurements based on the timestamps available at Robot 0 after one TWR transaction are formulated to get models that are only a function of the states being estimated, as well as the known vectors between the transceivers and the IMUs resolved in the robot's body frame.

First, notice that the distance $d^{s_2 f_1}$ between transceivers in \eqref{eq:ts1}-\eqref{eq:ts6} can be written as a function of the estimated states, 
\begin{align}
    d^{s_2 f_1} &= \norm{\mbf{r}_0^{s_2 f_1}} \nonumber \\
    &= \norm{\mbf{r}_0^{s_2 0} - \mbf{r}_0^{f_1 0}} \nonumber \\
    &= \norm{(\mbf{C}_{02} \mbf{r}_2^{s_2 2} + \mbf{r}_0^{2 0}) - (\mbf{C}_{01} \mbf{r}_1^{f_1 1} + \mbf{r}_0^{1 0})} \nonumber \\
    &= \norm{\mbs{\Pi} \left(\mbf{T}_{02} \mbftilde{r}_2^{s_2 2} - \mbf{T}_{01} \mbftilde{r}_1^{f_1 1} \right)}, \label{eq:dist}
\end{align}
where $\norm{\cdot}$ is the Euclidean norm, $\mbs{\Pi} = \bma{ccc} \mbf{1}_3 & \mbf{0}_{3 \times 2} \ema \in \mathbb{R}^{3 \times 5}$, and
\beq
    \mbftilde{r} = \bma{ccc} 
        \mbf{r}^\trans & 0 & 1
    \ema^\trans. \nonumber
\eeq
\jln{To design the} EKF, the linearization of \eqref{eq:dist} with respect to the state is shown in Appendix \ref{appx:linearize_range}.

Therefore, pseudomeasurements can be formed that are only a function of the distance between the transceivers, 
the clock states \jln{(relative to $f_0$)}, and the white \jln{timestamping} noise. 
The \textbf{first pseudomeasurement} is the standard ToF measurement associated with DS-TWR \cite{Shalaby2022a}, which from timestamps \eqref{eq:ts1}-\eqref{eq:ts6} can be written as
\begin{align}
    y^\text{tof} &= \f{1}{2} \left( \left( \tilde{\mathtt{R}}_{f_1}^{2} - \tilde{\mathtt{T}}_{f_1}^{1} \right) - \f{\tilde{\mathtt{R}}_{f_1}^{3} - \tilde{\mathtt{R}}_{f_1}^{2}}{\tilde{\mathtt{T}}_{s_2}^{3} - \tilde{\mathtt{T}}_{s_2}^{2}} \left( \tilde{\mathtt{T}}_{s_2}^{2} - \tilde{\mathtt{R}}_{s_2}^{1} \right) \right) \nonumber \\
    &\approx \f{1}{c} d^{s_2 f_1} + \f{1}{2} \left( \eta_{f_1}^{2} - \eta_{f_1}^{1} - \eta_{s_2}^{2} + \eta_{s_2}^{1} \right). \label{eq:y_r}
\end{align}
The relation \eqref{eq:y_r} is obtained under the following approximations. First,
clock skews $\gamma_i$ are assumed constant over the duration of the transaction, \tro{where the transaction}
% which 
is in the order of a few milliseconds, so that during the transaction
\[
\tau_i(t') - \tau_i(t) \approx \gamma_{i} (t'-t),
\]
for any time instances $t,t'$ and Transceiver $i$.
Second, $\Delta t^{21}$, which like $\Delta t^{32}$ is in the order of a few 
hundreds of microseconds, is much greater than $\f{d}{c}$, 
and since clock skews are also small (in the order 
of a few parts-per-million \cite{Neirynck2017}), then to first order
%$\gamma_{i} (\mathtt{T}^2-\mathtt{R}^1) \approx \gamma_i \Delta t^{21}$.}
$\gamma_{i} (\mathtt{R}^2-\mathtt{T}^1) \approx \gamma_i \Delta t^{21}$.
Third,
\beq
    \f{(1 + \gamma_{f_1}) \Delta t^{32} + \eta_{f_1}^{3} - \eta_{f_1}^{2}}{(1 + \gamma_{s_2}) \Delta t^{32} + \eta_{s_2}^{3} - \eta_{s_2}^{2}} \approx \f{(1 + \gamma_{f_1})}{(1 + \gamma_{s_2})} \nonumber
\eeq
\jln{because the timestamping noise, in the order of a few hundred picoseconds at most, is much
smaller than $\Delta t^{32}$.}
\jln{Finally,}
\beq
    \f{(1 + \gamma_{f_1})}{(1 + \gamma_{s_2})} \left( \eta_{s_2}^{2} - \eta_{s_2}^{1} \right) \approx \eta_{s_2}^{2} - \eta_{s_2}^{1}, \nonumber
\eeq
%which simplifies the derivation significantly, 
\jln{to first order, because the clock skews and timestamping noise are both small.}

The \textbf{second pseudomeasurement} is a direct clock offset measurement between the initiating and target transceivers, which from timestamps \eqref{eq:ts1}-\eqref{eq:ts6} can be written as
\begin{align}
    y^\tau &= \f{1}{2} \left( \left( \tilde{\mathtt{R}}_{f_1}^{2} + \tilde{\mathtt{T}}_{f_1}^{1} \right) - \f{\tilde{\mathtt{R}}_{f_1}^{3} - \tilde{\mathtt{R}}_{f_1}^{2}}{\tilde{\mathtt{T}}_{s_2}^{3} - \tilde{\mathtt{T}}_{s_2}^{2}} \left( \tilde{\mathtt{T}}_{s_2}^{2} - \tilde{\mathtt{R}}_{s_2}^{1} \right) - 2\tilde{\mathtt{R}}_{s_2}^{1} \right) \nonumber \\
    &\approx \tau_{f_1 f_0} - \tau_{s_2 f_0} + \f{1}{2} \left( \eta_{f_1}^{2} + \eta_{f_1}^{1} - \eta_{s_2}^{2} - \eta_{s_2}^{1} \right), \label{eq:y_tau}
\end{align}
\jln{using the fact that $\tau_{f_1} - \tau_{s_2} = \tau_{f_1 f_0} - \tau_{s_2 f_0}$.}
\jln{Here and in the following, clock offsets are evaluated at time $\mathtt{T}^1$, which is omitted from the notation.}
This model is somewhat similar to the measurement model proposed in \cite{Cano2019}, but involves an additional term to correct the effect of the clock skew on the measured offset. \tro{In fact, the form of the first two pseudomeasurements is chosen to cancel out the terms $\f{1}{2}(1+\gamma_{f_1})\Delta t^{21}$ and $-\f{1}{2}(1+\gamma_{s_2})\Delta t^{21}$ by multiplying the latter with $\f{1+\gamma_{f_1}}{1+\gamma_{s_2}}$.}

The \textbf{third pseudomeasurement} is associated with the first passive-listening timestamp, which is a function of the distance between the passive robot and the initiating robot, as well the clock offset between the two transceivers. Using timestamps \eqref{eq:ts1} and \eqref{eq:ts7} for $i \in \{f_0, s_0\}$, 
and $\tau_{f_0 f_0} \triangleq 0$,
this is written as
\begin{align}
    y^\text{p,1}_{i} &= \tilde{\mathtt{P}}_{i}^{1,i} - \tilde{\mathtt{T}}_{f_1}^{1} = \f{1}{c}d^{f_1 i} + \tau_{i f_0} - \tau_{f_1 f_0} + \eta_{i}^{1} - \eta_{f_1}^{1}.
    \label{eq:y_p1}
\end{align}

The \textbf{fourth pseudomeasurement} is similar to the third one, with an additional skew-correction component to model the passage of time $\Delta t^{21}$ between the first and second signal in two clocks with different clock rates. Using timestamps \eqref{eq:ts5} and \eqref{eq:ts8} for $i \in \{f_0, s_0\}$, and $\gamma_{f_0 f_0} \triangleq 0$, this is modelled as
\begin{align}
    y^\text{p,2}_{i} &= \tilde{\mathtt{P}}_{i}^{2,i} - \tilde{\mathtt{T}}_{s_2}^{2} \nonumber \\
    &= \f{1}{c}d^{s_2 i} + \tau_{i f_0} - \tau_{s_2 f_0} \nonumber \\ &\hspace{30pt} + (\gamma_{i f_0} - \gamma_{s_2 f_0}) \Delta t^{21} + \eta_{i}^{2} - \eta_{s_2}^{2} \label{eq:y_p2}.
\end{align}
\jln{using the fact that 
$\gamma_{i} - \gamma_{s_2} = \gamma_{i f_0} - \gamma_{s_2 f_0}$.
The exact delay $\Delta t^{21}$ appearing in \eqref{eq:y_p2} is in fact unknown, 
as delay values are enforced by the transceivers in their own clocks.
%, so in the scenario presented in Figure \ref{fig:ds_twr} the only measured values 
%are $\Delta t^{21}_{s_2}$ and $\Delta t^{31}_{s_2}$. 
Nonetheless, to first order, the corresponding term can be replaced by
\[
    (\gamma_{i f_0} - \gamma_{s_2 f_0})\Delta t^{21} \approx (\gamma_{i f_0} - \gamma_{s_2 f_0})(\tilde{\mathtt{T}}_{s_2}^{2} - \tilde{\mathtt{R}}_{s_2}^{1}). 
\]}

Lastly, the \textbf{fifth pseudomeasurement} is similar to the fourth pseudomeasurement, but modelling the evolution of the clocks over a longer time window $\Delta t^{31}$. Using timestamps \eqref{eq:ts6} and \eqref{eq:ts9} for $i \in \{f_0, s_0\}$, this is modelled as
\begin{align}
    y^\text{p,3}_{i} &= \tilde{\mathtt{P}}_{i}^{3,i} - \tilde{\mathtt{T}}_{s_2}^{3} \nonumber \\
    &= \f{1}{c}d^{s_2 i} + \tau_{i f_0} - \tau_{s_2 f_0} \nonumber \\ &\hspace{30pt} + (\gamma_{i f_0} - \gamma_{s_2 f_0}) \Delta t^{31} + \eta_{i}^{3} - \eta_{s_2}^{3}. \label{eq:y_p3}
\end{align}
\jln{As before, $\Delta t^{31}$ is unknown, but to first order 
\[
(\gamma_{i f_0} - \gamma_{s_2 f_0})\Delta t^{31} \approx (\gamma_{i f_0} - \gamma_{s_2 f_0})(\tilde{\mathtt{T}}_{s_2}^{3} - \tilde{\mathtt{R}}_{s_2}^{1}).
\]}

Note the last three pseudomeasurements are per listening transceiver $i$, and therefore there are a total of 8 pseudomeasurements available at Robot~0 if it is not involved in the ranging transaction, or 5 pseudomeasurements if one of the transceivers on Robot~0 is active. \tro{The additional pseudomeasurements available at the listening transceivers results in a $(1+3n)$-fold increase in the total number of distinct measurements when considering a centralized approach where passive listening measurements from all robots are available, and a $(\f{1}{2}+2n)$-fold increase in the number of distinct measurements when considering the perspective of an individual robot that does not have access to passive listening measurements at other robots. For example, for 5 neighbouring robots, this results in a 16-fold and an 11.5-fold increase in the number of measurements, respectively. The former is purely due to passive listening measurements, while the latter is due to passive listening measurements as well as the ability to obtain direct ToF measurements between two neighbouring robots. The proof of this claim is  given in Appendix \ref{appx:fold}.}

% Additionally, note that the delay intervals $\Delta t^{21}$ and $\Delta t^{31}$ appear in the pseudomeasurements \eqref{eq:y_p2}-\eqref{eq:y_p3} as if the length of the interval is known in the clock of Transceiver $f_0$. These are in fact unknown values, as the delay values are programmed into all transceivers manually, which then enforce these delays in their own clocks, so in the scenario presented in Figure \ref{fig:ds_twr} the only measured values are $\Delta t^{21}_{s_2}$ and $\Delta t^{31}_{s_2}$. Nonetheless, the approximation
% \begin{align*}
%     (\gamma_{i f_0} - \gamma_{s_2 f_0})\Delta t^{21}_{f_0} &\approx (\gamma_{i f_0} - \gamma_{s_2 f_0})(\tilde{\mathtt{T}}_{s_2}^{2} - \tilde{\mathtt{R}}_{s_2}^{1}), \\
%     (\gamma_{i f_0} - \gamma_{s_2 f_0})\Delta t^{31}_{f_0} &\approx (\gamma_{i f_0} - \gamma_{s_2 f_0})(\tilde{\mathtt{T}}_{s_2}^{3} - \tilde{\mathtt{R}}_{s_2}^{1})
% \end{align*}
% can be made as explained in Appendix \ref{appx:approx_delta_t}.

%\subsection{Cross-correlation Between Pseudomeasurements}
\subsection{Pseudomeasurements' \jln{Covariance Matrix}}

Given that the pseudomeasurements are a function of the same measured timestamps, cross-correlations between the pseudomeasurements exist and must be correctly modelled in the filter. Computing the variance of the pseudomeasurements \eqref{eq:y_r}-\eqref{eq:y_p3} is straightforward, and can be summarized as
\begin{align*}
    \mathbb{E}\left[ (y^\text{tof} - \bar{y}^\text{tof})^2 \right] &= \sigma^2, \quad
    \mathbb{E}\left[ (y^\tau - \bar{y}^\tau)^2 \right] = \sigma^2, \\
    \mathbb{E}\left[ (y^\text{p,j}_i - \bar{y}^\text{p,j}_i)^2 \right] &= 2\sigma^2, \quad j \in \{1,2,3\},
\end{align*}
where an overbar denotes a noise-free value. Meanwhile, the cross-correlation between the ToF and offset measurements can be computed as
% \begin{align*}
    $\mathbb{E}\left[ (y^\text{tof} - \bar{y}^\text{tof})(y^\tau - \bar{y}^\tau) \right] = 0$
% \end{align*}
as the noise values are of alternating signs. Lastly, the cross-correlations between the passive listening measurements and the ToF measurements can be shown to be
\begin{align*}
    \mathbb{E}\left[ (y^\text{p,1}_i - \bar{y}^\text{p,1}_i)(y^\text{tof} - \bar{y}^\text{tof}) \right] &= \f{1}{2}\sigma^2, \\
    \mathbb{E}\left[ (y^\text{p,2}_i - \bar{y}^\text{p,2}_i)(y^\text{tof} - \bar{y}^\text{tof}) \right] &= \f{1}{2}\sigma^2, \\
    \mathbb{E}\left[ (y^\text{p,3}_i - \bar{y}^\text{p,3}_i)(y^\text{tof} - \bar{y}^\text{tof}) \right] &= 0,
\end{align*}
while the cross-correlations with offset measurements are the same but with an opposite sign for the correlation with $y^\text{p,2}_i$. Passive listening measurements of different transceivers are also correlated. Stacking all the pseudomeasurements into one column matrix gives the random measurement vector 
\begin{equation*}
    \mbf{y} = \bma{cccccccc} y^\text{tof} & y^\tau & y^\text{p,1}_{f_0} & y^\text{p,2}_{f_0} & y^\text{p,3}_{f_0} & y^\text{p,1}_{s_0} & y^\text{p,2}_{s_0} & y^\text{p,3}_{s_0} \ema^\trans,
\end{equation*}
with mean $\mbfbar{y}$ and covariance matrix $\mbf{R}$, where
%distributed as per $\mbf{y} \sim \mc{N} (\mbfbar{y}, \mbf{R})$, where
\beq
    \mbf{R} = \bma{ccc}
        \sigma^2 \mbf{1}_2 & \f{1}{2} \sigma^2 \mbf{D} & \f{1}{2} \sigma^2 \mbf{D} \\
        \f{1}{2} \sigma^2 \mbf{D}^\trans & 2\sigma^2 \mbf{1}_3 & \sigma^2 \mbf{1}_3  \\
        \f{1}{2} \sigma^2 \mbf{D}^\trans & \sigma^2 \mbf{1}_3 & 2\sigma^2 \mbf{1}_3
    \ema, \nonumber
\eeq
and   
\beq
    \mbf{D} = \bma{ccc}
        1 & 1 & 0 \\
        1 & -1 & 0
    \ema. \nonumber
\eeq
The measurement vector $\mbf{y}$ and its covariance $\mbf{R}$ are used in the correction step of an on-manifold EKF, where they are fused with the process model derived in the next section.

%% file: sections/process_model.tex
To derive the process model, a Lie group referred to here as $DE_2(3)$ ($D$ stands for \emph{Delta}) with matrices of the form
\beq
    \label{eq:U}
    \mbf{U} = \bma{ccc}
        \mbf{C} & \mbf{v} & \mbf{r} \\
        & 1 & \Delta t \\
        & & 1
    \ema \in DE_2(3)
\eeq
is introduced, where $\mbf{C} \in SO(3)$, $\mbf{v}, \mbf{r} \in \mathbb{R}^3$, and $\Delta t \in \mathbb{R}$. The inverse of $\mbf{U}$ in \eqref{eq:U} is 
\beq
    \mbf{U}^{-1} = \bma{ccc}
        \mbf{C}^\trans & -\mbf{C}^\trans \mbf{v} & - \mbf{C}^\trans(\mbf{r}- \Delta t \mbf{v}) \\
        & 1 & -\Delta t \\
        & & 1
    \ema \in DE_2(3). \nonumber
\eeq
Meanwhile, the adjoint operator satisfying 
\beq
    \operatorname{Exp}(\operatorname{Ad}(\mbf{U})\mbs{\xi}) \triangleq \mbf{U}\operatorname{Exp}(\mbs{\xi})\mbf{U}^{-1}, \quad \operatorname{Exp}(\mbs{\xi}) \in SE_2(3) \nonumber
\eeq
is given by 
\beq
    \operatorname{Ad}(\mbf{U}) = \bma{ccc}
        \mbf{C} & \mbf{0} & \mbf{0} \\
        \mbf{v}^\times \mbf{C} & \mbf{C} & \mbf{0} \\
        - (\Delta t \mbf{v} - \mbf{r})^\times \mbf{C} & -\Delta t \mbf{C} &  \mbf{C}
    \ema, \nonumber
\eeq
where, for $\mbf{v} = \bma{ccc} v_1 & v_2 & v_3 \ema^\trans \in \mathbb R^3$, 
\begin{align*}
    \mbf{v}^\times = \bma{ccc} 
        0 & -v_3 & v_2 \\
        v_3 & 0 & -v_1 \\
        -v_2 & v_1 & 0
    \ema.
\end{align*}

% $(\cdot)^\times: \mathbb{R}^3 \rightarrow \mathfrak{so}(3)$ is the skew-symmetric cross-product matrix operator in $\mathbb{R}^3$, and $\mathfrak{so}(3)$ is the Lie algebra associated with the Special Orthogonal Group $SO(3)$. 

Additionally, following the terminology in \cite[Chapter 9]{Barfoot2022}, 
a \emph{time machine} is a matrix $\mbf{M}$ of the form
\begin{align*}
    \mbf{M} = \bma{ccc}
        \mbf{1} & & \\
        & 1 & \Delta t \\
        & & 1
    \ema \in \mathbb{R}^{5 \times 5},
\end{align*}
where $\Delta t \in \mathbb{R}$. This allows writing $\mbf{U}$ in \eqref{eq:U} as the product of two matrices, 
\beq
    \mbf{U} = \underbrace{\bma{ccc}
        \mbf{1} & & \\
        & 1 & \Delta t \\
        & & 1
    \ema}_{\mbf{M}} \underbrace{\bma{ccc}
        \mbf{C} & \mbf{v} & \mbf{r} \\
        & 1 &  \\
        & & 1
    \ema}_{\mbf{T} \in SE_2(3)}. \nonumber
\eeq
It can be shown that $\mbf{M}$ is in itself an element of a Lie group closed under matrix multiplication.

This section first extends the results in \cite[Chapter 9]{Barfoot2022} to address relative extended pose states. The clock-state process model is then derived. These are then used alongside the ranging protocol presented in Section \ref{sec:ranging_protocol} in the CSRPE.

\subsection{Deriving the Extended-Pose Process Model} \label{subsec:derive_process_model}

The on-manifold relative-pose kinematic model is \jln{first} derived in continuous-time as a function of the IMU measurements. The process model for the relative attitude between Robot 0 and Robot $i$ is
\begin{align}
    \mbfdot{C}_{0i} = \mbf{C}_{0i} \left( \mbs{\omega}_i^{i0} \right)^\times, \label{eq:C_process_raw}
\end{align}
where $\mbs{\omega}_i^{i\jln{0}}$ is the angular velocity of Robot $i$'s body frame relative to Robot $0$'s body frame, resolved in Robot $i$'s body frame. However, the gyroscopes on Robots 0 and $i$ measure $\mbs{\omega}_0^{0a}$ and $\mbs{\omega}_i^{ia}$, respectively. Therefore, \eqref{eq:C_process_raw} is rewritten as
\begin{align}
    \mbfdot{C}_{0i} &= \mbf{C}_{0i} \left( \mbs{\omega}_i^{ia} - \mbf{C}_{0i}^\trans \mbs{\omega}_0^{0a} \right)^\times \nonumber \\
    &= - \mbf{C}_{0i} \left( \mbf{C}_{0i}^\trans \mbs{\omega}_0^{0a} \right)^\times + \mbf{C}_{0i} \left( \mbs{\omega}_i^{ia} \right)^\times \nonumber \\
    &= - \left( \mbs{\omega}_0^{0a} \right)^\times \mbf{C}_{0i} + \mbf{C}_{0i} \left( \mbs{\omega}_i^{ia} \right)^\times. \label{eq:C_process}
\end{align}

Meanwhile, using the \emph{transport theorem} \cite[Chapter 2.10]{Rao2006}, the process model for the relative velocity of Robot $i$ relative to Robot $0$ is 
\begin{align}
    %^0 \hspace{-2pt} \underrightarrow{\dot{v}}^{i0/a} &= \underrightarrow{a}^{i0/a} 
    %- \underrightarrow{\omega}^{0a} \times \underrightarrow{v}^{i0/a} \nonumber \\
    ^0 \hspace{-2pt} \underrightarrow{\dot{v}}^{i0/a} &= - \underrightarrow{\omega}^{0a} \times \underrightarrow{v}^{i0/a} + \underrightarrow{a}^{iw/a} - \underrightarrow{a}^{0w/a}, \label{eq:v_process_raw}
\end{align}
where $w$ is \jln{any} point fixed to the reference frame $a$. 
Denoting \jln{the} \emph{specific forces} \jln{measured by the accelerometers} as
\begin{align*}
    \underrightarrow{\alpha}^0 \triangleq \underrightarrow{a}^{0w/a} - \underrightarrow{g}, \qquad \underrightarrow{\alpha}^i \triangleq \underrightarrow{a}^{iw/a} - \underrightarrow{g},
\end{align*}
where $\underrightarrow{g}$ is the gravity vector, \eqref{eq:v_process_raw} can be written as
\begin{align}
    ^0 \hspace{-2pt} \underrightarrow{\dot{v}}^{i0/a} &= - \underrightarrow{\omega}^{0a} \times \underrightarrow{v}^{i0/a} + \underrightarrow{\alpha}^i - \underrightarrow{\alpha}^0. \label{eq:v_process_unresolved}
\end{align}

Similarly, the transport theorem gives the following process model for the position of Robot $i$ relative to Robot $0$ 
\begin{align}
    ^0 \hspace{-2pt} \underrightarrow{\dot{r}}^{i0} &= - \underrightarrow{\omega}^{0a} \times \underrightarrow{r}^{i0} + \underrightarrow{v}^{i0/a}. \label{eq:r_process_unresolved}
\end{align}

Lastly, resolving \eqref{eq:v_process_unresolved} and \eqref{eq:r_process_unresolved} in the body frame of Robot 0 and writing these equations as a function of the accelerometer-measured quantities $\mbs{\alpha}^0_0$ and $\mbs{\alpha}^i_i$ yields
\begin{align}
    ^0 \mbfdot{v}_0^{i0/a} &= - \left(\mbs{\omega}_0^{0a}\right)^\times \mbf{v}_0^{i0/a} + \mbf{C}_{0i} \mbs{\alpha}_i^i - \mbs{\alpha}_0^0, \label{eq:v_process} \\
    ^0 \mbfdot{r}_0^{i0} &= - \left(\mbs{\omega}_0^{0a}\right)^\times \mbf{r}_0^{i0} + \mbf{v}_0^{i0/a}. \label{eq:r_process}
\end{align}

% Meanwhile, using the \emph{transport theorem} \cite[Chapter 2.10]{Rao2006}, the process model for the relative velocity of Robot $i$ relative to Robot $0$ is 
% \begin{align}
%     \mbfdot{v}_0^{i0/a} &= - \left(\mbs{\omega}_0^{0a}\right)^\times \mbf{v}_0^{i0/a} + \mbf{a}_0^{i0/a/a} \nonumber \\
%     %
%     &= - \left(\mbs{\omega}_0^{0a}\right)^\times \mbf{v}_0^{i0/a} + \mbf{C}_{0i} \mbf{a}_i^{iw/a/a} - \mbf{a}_0^{0w/a/a}. \label{eq:v_process_raw}
% \end{align}
% The body-frame acceleration is typically measured using an accelerometer, which can be modelled in the absence of noise as
% \begin{align*}
%     \mbs{\alpha}_0^0 = \mbf{a}_0^{0w/a/a} - \mbf{C}_{a0}^\trans \mbf{g_a}, \qquad \mbs{\alpha}_i^i = \mbf{a}_i^{iw/a/a} - \mbf{C}_{ai}^\trans \mbf{g_a},
% \end{align*}
% where $\mbf{g}_a$ is the gravity vector resolved in the absolute frame. Therefore, \eqref{eq:v_process_raw} can be written as
% \begin{align}
%     \mbfdot{v}_0^{i0/a} &= - \left(\mbs{\omega}_0^{0a}\right)^\times \mbf{v}_0^{i0/a} + \mbf{C}_{0i} \mbs{\alpha}_i^i + \mbf{C}_{a0}^\trans \mbf{g}_a - \mbs{\alpha}_0^0 - \mbf{C}_{a0}^\trans \mbf{g}_a \nonumber \\
%     %
%     &= - \left(\mbs{\omega}_0^{0a}\right)^\times \mbf{v}_0^{i0/a} + \mbf{C}_{0i} \mbs{\alpha}_i^i - \mbs{\alpha}_0^0. \label{eq:v_process}
% \end{align}

% Lastly, again using the transport theorem, the process model for the relative position of Robot $i$ relative to Robot $0$ is 
% \begin{align}
%     \mbfdot{r}_0^{i0} &= - \left(\mbs{\omega}_0^{0a}\right)^\times \mbf{r}_0^{i0} + \mbf{v}_0^{i0/a}. \label{eq:r_process}
% \end{align}

Combining \eqref{eq:C_process}, \eqref{eq:v_process}, and \eqref{eq:r_process}, 
the \jln{extended} relative-pose process model for Robot $i$ can be written compactly 
%as a function of the extended pose matrix $\mbf{T}_{0i}$ 
as
\begin{align}
    \mbfdot{T}_{0i} &= \bma{ccc}
        \mbfdot{C}_{0i} & ^0\mbfdot{v}_0^{i0/a} & ^0\mbfdot{r}_0^{i0} \\
        & 0 & \\
        & & 0 
    \ema \nonumber \\
    &= - \bma{ccc}
        \left(\mbs{\omega}_0^{0a}\right)^\times & \mbs{\alpha}_0^0 &  \\
        & & 1 \\
        & & 0 
    \ema \mbf{T}_{0i} \nonumber \\ &\hspace{62pt} + \mbf{T}_{0i} \bma{ccc}
        \left(\mbs{\omega}_i^{ia}\right)^\times & \mbs{\alpha}_i^i &  \\
        & & 1 \\
        & & 0 
    \ema \nonumber \\
    &\triangleq - \mbftilde{U}_0 \mbf{T}_{0i} + \mbf{T}_{0i} \mbftilde{U}_i, \label{eq:process_model_ct}
\end{align}
\jln{with the matrices $\mbftilde{U}_0$ and $\mbftilde{U}_i$ containing the
IMU measurements for Robot $0$ and Robot $i$, respectively.} 

\subsection{\jln{Discrete-Time} Extended-Pose Process Model}

In order to discretize \eqref{eq:process_model_ct}, the common assumption is made 
that accelerations and angular velocities are constant between IMU measurements, 
which is justified by the fact that IMU measurements typically occur at 
a high frequency ($\sim$100-1000 Hz). Consequently, since \eqref{eq:process_model_ct} is a \emph{differential Sylvester equation}, and setting the initial condition to be $\mbf{T}_{0i,k}$ at time-step $k$, a closed-form solution exists of the form \cite{Behr2019}
\beq
    \label{eq:process_model_dt}
    \mbf{T}_{0i,k+1} = \underbrace{\operatorname{exp} (\mbftilde{U}_{0,k} \Delta t)^{-1}}_{\mbf{U}_{0,k}^{-1}} \mbf{T}_{0i,k} \underbrace{\operatorname{exp} (\mbftilde{U}_{i,k} \Delta t)}_{\mbf{U}_{i,k}},
\eeq
where $\Delta t$ is the time interval between the IMU measurements at time-steps $k$ and $k+1$.

Following a similar derivation as in \cite[Chapter 9]{Barfoot2022}, expanding the matrix exponential is shown in Appendix \ref{appx:U} to yield a closed-form matrix of the form
\begin{align*}
    &\mbf{U}_{0,k} = \bma{ccc}
        \operatorname{Exp}(\mbs{\Omega}_{0,k}) & \Delta t \mbf{J}_l\left(\mbs{\Omega}_{0,k}\right) \mbs{\alpha}_{0,k}^0 & \f{\Delta t^2}{2} \mbf{N}\left(\mbs{\Omega}_{0,k}\right) \mbs{\alpha}_{0,k}^0 \\
        & 1 & \Delta t \\
        & & 1
    \ema
\end{align*} 
where $\mbs{\Omega}_{0,k} \triangleq \mbs{\omega}_{0,k}^{0a} \Delta t$ and $\mbf{J}_l$ is the left Jacobian of $SO(3)$. Both $\mbf{J}_l$ and $\mbf{N}$ are defined in Appendix \ref{appx:U}. Note that $\mbf{U}_{0,k}$ is an element of the aforementioned Lie group $DE_2(3)$. Similarly, $\mbf{U}_{i,k} \in DE_2(3)$ is of the same form as $\mbf{U}_{0,k}$ with the inputs being that of neighbouring Robot $i$ instead.

\subsection{Linearizing the Extended-Pose Process Model} \label{subsec:process_model_linearize}

To perform uncertainty propagation computations for the extended-pose states, 
the process model is now linearized. Throughout this paper, the state 
is perturbed on the left, as it yields \jln{simpler} Jacobians. Nonetheless, 
a similar derivation can be done by perturbing the state on the right. 

Perturbing \eqref{eq:process_model_dt} with respect to the state yields
\begin{align*}
    \operatorname{Exp} (\mbsdel{\xi}_{0i,k+1}) \mbfbar{T}_{0i,k+1} &= \bar{\mbf{U}}_{0,k}^{-1} \operatorname{Exp} (\mbsdel{\xi}_{0i,k}) \mbfbar{T}_{0i,k} \bar{\mbf{U}}_{i,k} \\
    &= \operatorname{Exp} ( \operatorname{Ad}(\bar{\mbf{U}}_{0,k}^{-1})\mbsdel{\xi}_{0i,k}) \bar{\mbf{U}}_{0,k}^{-1} \mbfbar{T}_{0i,k} \bar{\mbf{U}}_{i,k}.
\end{align*}
Cancelling out nominal terms and taking the $\operatorname{Log}(\cdot)$ of both sides results in the linearized model
\beq
    \label{eq:process_model_cov_state}
    \mbsdel{\xi}_{0i,k+1} = \operatorname{Ad}(\bar{\mbf{U}}_{0,k}^{-1})\mbsdel{\xi}_{0i,k}.
\eeq

To perturb \eqref{eq:process_model_dt} with respect to the input noise, the aforementioned concept of time machines is used. The input matrix $\mbf{U}_{0,k}$ can be written as
\begin{align}
    &\mbf{U}_{0,k} \nonumber \\
    &= \mbf{M} \bma{ccc}
        \operatorname{Exp}(\mbs{\Omega}_{0,k}) & \Delta t \mbf{J}_l\left(\mbs{\Omega}_{0,k}\right) \mbs{\alpha}_{0,k}^0 & \f{\Delta t^2}{2} \mbf{N}\left(\mbs{\Omega}_{0,k}\right) \mbs{\alpha}_{0,k}^0 \\
        & 1 & \\
        & & 1
    \ema \nonumber \\
    &= \mbf{M} \operatorname{Exp} \left( \bma{c}
        \mbs{\omega}_{0,k}^{0a} \Delta t \\
        \mbs{\alpha}_{0,k}^0 \Delta t \\
        \f{\Delta t^2}{2} \mbf{J}_l\left(\mbs{\Omega}_{0,k}\right)^{-1} \mbf{N}\left(\mbs{\Omega}_{0,k}\right) \mbs{\alpha}_{0,k}^0
    \ema \right) \nonumber \\
    &= \mbf{M} \operatorname{Exp} \Bigg( \underbrace{\bma{cc}
        \Delta t \mbf{1}  & \\
        & \Delta t \mbf{1} \\
        & \f{\Delta t^2}{2} \mbf{J}_l\left(\mbs{\Omega}_{0,k}\right)^{-1} \mbf{N}\left(\mbs{\Omega}_{0,k}\right)
    \ema}_{\mbf{V}_{0,k}} \underbrace{\bma{c}
        \mbs{\omega}_{0,k}^{0a} \\
        \mbs{\alpha}_{0,k}^0
    \ema}_{\mbf{u}_{0,k}} \Bigg) \nonumber \\
    &\triangleq \mbf{M} \operatorname{Exp} (\mbf{V}_{0,k} \mbf{u}_{0,k}), \label{eq:u_tilde_factored}
\end{align}
where $\mbf{u}_{0,k} \in \mathbb{R}^6$ is Robot 0's IMU measurements or \emph{input} at time-step $k$. Taking the perturbation of \eqref{eq:u_tilde_factored} with respect to the input yields
\begin{align}
    \mbf{U}_{0,k} &\approx \mbf{M} \operatorname{Exp} (\mbfbar{V}_{0,k} (\mbfbar{u}_{0,k} + \mbfdel{u}_{0,k})) \nonumber \\
    &\approx \mbf{M} \operatorname{Exp} (\mbfbar{V}_{0,k} \mbfbar{u}_{0,k}) \operatorname{Exp} (\mbc{J}_l(-\mbfbar{V}_{0,k} \mbfbar{u}_{0,k})\mbfbar{V}_{0,k} \mbfdel{u}_{0,k}) \nonumber \\
    &= \bar{\mbf{U}}_{0,k} \operatorname{Exp} (\mbc{J}_l(-\mbfbar{V}_{0,k} \mbfbar{u}_{0,k})\mbfbar{V}_{0,k} \mbfdel{u}_{0,k}) \nonumber \\
    &\triangleq  \bar{\mbf{U}}_{0,k} \operatorname{Exp} (\mbf{L}_{0,k} \mbfdel{u}_{0,k})
\end{align}
where input noise perturbations in $\mbf{V}_{0,k}$ are neglected as the term $ \f{\Delta t^2}{2} \mbf{J}_l\left(\mbs{\Omega}_{0,k}\right)^{-1} \mbf{N}\left(\mbs{\Omega}_{0,k}\right)$ is small when the measurements 
are obtained using a high-rate IMU, $\mbf{L}_{0,k} \triangleq \mbc{J}_l(-\mbfbar{V}_{0,k} \mbfbar{u}_{0,k})\mbfbar{V}_{0,k}$, 
and $\mbc{J}_l(\cdot)$ is the left Jacobian of $SE_2(3)$ \ms{\cite[Eq. (94)]{Brossard2021}}. Similarly, 
\begin{equation}
    \mbf{U}_{i,k} = \mbf{M} \operatorname{Exp} (\mbf{V}_{i,k} \mbf{u}_{i,k}) \approx  \bar{\mbf{U}}_{i,k} \operatorname{Exp} (\mbf{L}_{i,k} \mbfdel{u}_{i,k}). \label{eq:Ui_perturbation}
\end{equation}
Therefore, left-perturbing the state process model \eqref{eq:process_model_dt} with respect to the input noise yields
\begin{align*}
    &\operatorname{Exp} (\mbsdel{\xi}_{0i,k+1}) \mbfbar{T}_{0i,k+1} \nonumber \\
    &= \operatorname{Exp} (-\mbf{L}_{0,k} \mbfdel{u}_{0,k}) \bar{\mbf{U}}_{0,k}^{-1} \mbfbar{T}_{0i,k} \bar{\mbf{U}}_{i,k} \operatorname{Exp} (\mbf{L}_{i,k} \mbfdel{u}_{i,k}) \\
    &= \operatorname{Exp} (-\mbf{L}_{0,k} \mbfdel{u}_{0,k}) \operatorname{Exp} (\operatorname{Ad}(\mbfbar{T}_{0i,k+1})\mbf{L}_{i,k} \mbfdel{u}_{i,k}) \mbfbar{T}_{0i,k+1},
\end{align*}
which can then be simplified to give
\begin{align}
    \label{eq:process_model_cov_noise}
    \mbsdel{\xi}_{0i,k+1} &= -\mbf{L}_{0,k} \mbfdel{u}_{0,k} + \operatorname{Ad}(\mbfbar{T}_{0i,k+1})\mbf{L}_{i,k} \mbfdel{u}_{i,k}.
\end{align}

It is worth mentioning that cross-correlations develop between relative pose states 
\jln{for} all neighbours, because the noisy IMU measurements of Robot 0 are used 
to propagate all the relative pose states. These cross-correlations can be tracked using the models \eqref{eq:process_model_cov_state} and \eqref{eq:process_model_cov_noise}.

\subsection{\jln{Discrete-Time} Clock-State Process Model} \label{subsec:discrete_clock_process_model}

The state dynamics for every clock is modelled as in \eqref{eq:clock_dynamics}. Nonetheless, the clock states relative to real-time are unknown and unobservable. Therefore, clocks are modelled relative to clock $f_0$, thus giving dynamics of the form
\beq
    \label{eq:process_model_c_ct}
    %\mbfdot{c}_{cf_0} = \mbf{A} \mbf{c}_{cf_0} + \bma{cc} -\mbf{1} & \mbf{1} \ema \bma{c} \mbf{w}_{f_0r} \\ %\mbf{w}_{cr} \ema
    \mbfdot{c}_{i f_0} = \mbf{A} \mbf{c}_{i f_0} + \bma{cc} -\mbf{1} & \mbf{1} \ema \bma{c} \mbf{w}_{f_0} \\ \mbf{w}_{i} \ema
\eeq
for $i \in \mc{C} \backslash \{f_0\}$. 
Discretizing \eqref{eq:process_model_c_ct} yields \cite[Chapter 4.7]{Farrell2008}
\beq
    \label{eq:process_model_c discrete-time}
    \mbf{c}_{i f_0,k+1} = \mbf{A}^\text{d} \, \mbf{c}_{i f_0,k} + \mbf{w}_{i f_0,k}, 
\eeq
where 
\beq
    \mbf{A}^\text{d} = \operatorname{exp} (\mbf{A} \Delta t) = \bma{cc} 1 & \Delta t \\ & 1 \ema,  \nonumber
\eeq
$\mbf{w}_{i f_0,k} \sim \mc{N} \left( \mbf{0}, \mbf{Q}^\text{d} \right)$, and
\begin{align*}
    \mbf{Q}^\text{d} &= 2 \bma{cc} 
        \Delta t \mc{Q}^\tau + \f{1}{3} \Delta t^3 \mc{Q}^\gamma & \f{1}{2} \Delta t^2 \mc{Q}^\gamma \\
        \f{1}{2} \Delta t^2 \mc{Q}^\gamma & \Delta t \mc{Q}^\gamma
    \ema.
\end{align*}
Since the same noise $\mbf{w}_{f_0}$ \jln{appears in \eqref{eq:process_model_c_ct} 
for all $i \in \mc{C} \backslash \{f_0\}$, the process noise vectors
$\mbf{w}_{i f_0,k}$
in \eqref{eq:process_model_c discrete-time} are jointly Gaussian but
correlated, and one can show that their cross-covariance is
\[
\mathbb E \left[ \mbf{w}_{i f_0,k} \, \mbf{w}_{j f_0,k}^\trans \right] = \f{1}{2}\mbf{Q}^\text{d},
\]
for all $i, j \in \mc{C} \backslash \{f_0\}, i \neq j$.}

% %the evolution of the clock states is in fact correlated. 
% \jln{these states are correlated}.
% It can be shown that the augmented process noise vector for any two 
% clocks %$c_i, c_j \in \mc{C} \backslash \{f_0\}, c_i \neq c_j$ 
% $i, j \in \mc{C} \backslash \{f_0\}, i \neq j$
% is distributed as
% \beq
%     \bma{c} \mbf{w}_{i f_0,k} \\ \mbf{w}_{j f_0,k} \ema \sim \mc{N} \bigg( \mbf{0}, \bma{cc} \mbf{Q}^\text{d} & \f{1}{2}\mbf{Q}^\text{d} \\ \f{1}{2}\mbf{Q}^\text{d} & \mbf{Q}^\text{d} \ema \bigg). \nonumber
% \eeq

%% file: sections/preintegration.tex
\subsection{Need for Preintegration} \label{subsec:need_for_preintegration}

\begin{figure}[ht]
    \centering
    \begin{minipage}{\columnwidth}%
        \centering
        \subfloat[Subfigure 1 list of figures text][RMI communication without passive listening.]{
        \includegraphics[width=\columnwidth]{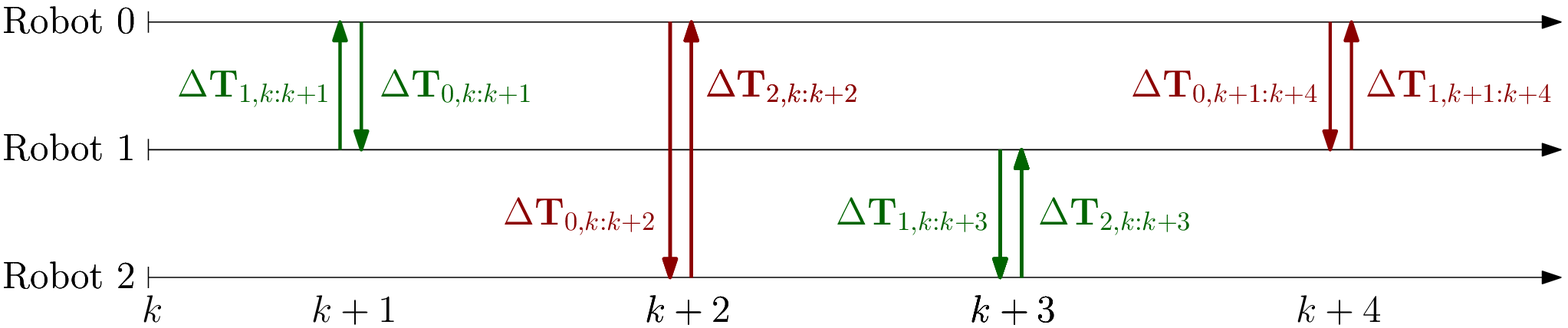}
        \label{fig:preint_no_passive}}
    \end{minipage}\\
    \vspace{10pt}%
    \begin{minipage}{\columnwidth}%
        \centering
        \subfloat[Subfigure 2 list of figures text][RMI communication with passive listening, \ms{where passive listening messages are shown using the grey arrows}.]{
        \includegraphics[width=\columnwidth]{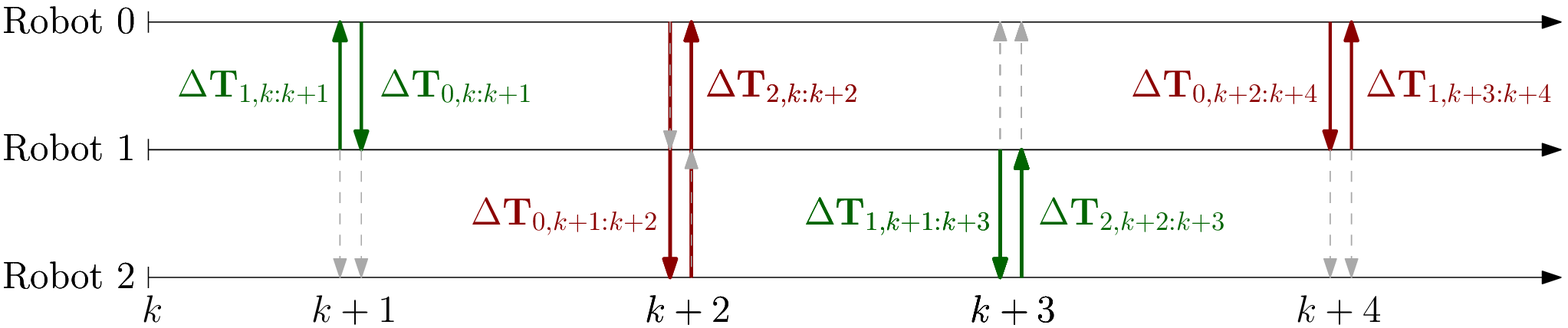}
        \label{fig:preint_w_passive}}
    \end{minipage}
    \caption{Communicated RMIs with and without passive listening over a window of \ms{4 ranging transactions, where $\Delta \mbf{T}_{i,\ell:m}$ is the RMI associated with the IMU measurements of Robot $i$ from time-step $\ell$ to time-step $m-1$.}}
    \label{fig:preint}
\end{figure}

When considering Robot 0's perspective, the estimated relative-pose state 
is updated using \eqref{eq:process_model_dt} and the corresponding 
\jln{error} covariance matrix using \eqref{eq:process_model_cov_state} 
and \eqref{eq:process_model_cov_noise}. 
Therefore, Robot 0 needs the IMU measurements of neighbouring robots at every time-step 
in order to update its estimated state of its neighbours. This is limiting since robots cannot broadcast their IMU measurements at the same rate as they are recorded due to \tro{the possibility of message collision if multiple robots attempt to broadcast at the same time}. Additionally, to allow DS-TWR transactions to occur at the highest rate possible, the IMU information should ideally be transmitted using the ranging messages presented in Section \ref{sec:ranging_protocol}.
% communication \jln{links} between robots do not exist at all times since 1) the ranging 
% frequency is lower than the frequency in which IMU measurements are recorded, 
% and 2) only one pair of robots can communicate at any point of time to prevent 
% UWB interference. 

\ms{In this section, the concept of preintegration is proposed to compactly encode} the IMU measurements of a neighbouring Robot $i$ over a window between two consecutive ranging instances 
% are preintegrated into 
\ms{using} one \emph{relative motion increment} (RMI), which is then sent over when Robot $i$ ranges 
with one of its neighbours. \jln{However, as} illustrated in Figure \ref{fig:preint_no_passive},
without passive listening the RMIs of Robot $i$ become available to Robot 0 only 
when Robot 0 and Robot $i$ communicate. Given that RMIs are computed iteratively 
as new IMU measurements arrive, each robot needs to keep track of one RMI per neighbour. 
For example, looking at Figure \ref{fig:preint_no_passive} at time-step $k+3$, 
Robot 1 would be communicating the RMI of IMU measurements in the window $k$ to $k+3$ 
to Robot 1, while also tracking a separate RMI for the window starting at $k+1$ 
to be sent to Robot 0 \ms{at time-step $k+4$}. 

\jln{On the other hand, passive listening over UWB lets two actively ranging
robots broadcast their RMIs to all other robots, as}
%broadcasting the RMIs using UWB
%can now have access to the communicated RMI between two actively ranging neighbours. 
%This is 
shown in Figure \ref{fig:preint_w_passive}. This has the advantage that IMU information 
of neighbours becomes available faster at all robots, and the robots computing RMIs only need 
to track one RMI at all times since all neighbours are up-to-date with the most recently 
communicated RMI.

\subsection{Relative Motion Increments}

% Consider the case where Robot 0 is estimating the relative state of Robot $i$. The state propagation given by \eqref{eq:process_model_dt} cannot be computed directly at every time-step as the input measurements of the neighbouring robot are not always available. Information from Robot $i$ can only be communicated when Robot $i$ is an active robot. If this occurs at non-adjacent time-steps $\ell$ and $m$, then at any time-step $\ell < k < m$ \eqref{eq:process_model_dt} cannot be computed directly. 

\ms{Consider the case where Robot $i$ is an active robot only at non-adjacent time-steps $\ell$ and $m$.} From \eqref{eq:process_model_dt}, the relative pose state at time-step $m$ can 
be computed from the relative pose state at time-step $\ell$ as
\begin{align}
    \label{eq:process_model_ell_to_m}
    \mbf{T}_{0i,m} = \left(\prod_{k = \ell}^{m-1} \mbf{U}_{0,k}\right)^{-1} \mbf{T}_{0i, \ell} \prod_{k = \ell}^{m-1} \mbf{U}_{i,k}.
\end{align}
The inputs of Robot 0 are available at Robot 0 as soon as the measurements occur, 
therefore the first \jln{term} of \eqref{eq:process_model_dt} can be computed directly 
at every time-step. \jln{On the other hand}, the inputs of Robot $i$ from 
time-step $\ell$ to $m-1$ will only be available when the robot actively shares 
it at time-step $m$. Rather than sharing the individual IMU measurements, 
Robot $i$ can simply send
\beq
    % \label{eq:RMI}
    \ms{\Delta \mbf{T}_{i,\ell:m}} = \prod_{k = \ell}^{m-1} \mbf{U}_{i,k} \in DE_2(3), \nonumber
\eeq
which \ms{is an RMI of the inputs of Robot $i$ in the window $\ell$ to $m$}. The process model representing time-propagation between non-adjacent time-steps can then be rewritten as 
\beq
    \label{eq:process_model_w_rmi}
    \mbf{T}_{0i,m} = \left(\prod_{k = \ell}^{m-1} \mbf{U}_{0,k}\right)^{-1} \mbf{T}_{0i, \ell} \Delta \mbf{T}_{i,\ell:m}.
\eeq
This is a feature of the process model \eqref{eq:process_model_ell_to_m} being reliant on the inputs of Robot $i$ in a separable way, meaning that the inputs of Robot $i$ can simply be post-multiplied in \eqref{eq:process_model_ell_to_m}. 
Robot $i$ \jln{computes its RMI} iteratively, starting with $\Delta \mbf{T}_{i,\ell:\ell} = \mbf{1}$,
and updating it when a new input measurement arrives as
\beq
    \label{eq:RMI_iter}
    \Delta \mbf{T}_{i,\ell:k+1} = \Delta \mbf{T}_{i,\ell:k} \mbf{U}_{i,k}.
\eeq

In order to linearize the RMI to be used in an EKF, a perturbation of the form
\beq
    \Delta \mbf{T}_{i,\ell:m} = \Delta \mbfbar{T}_{i,\ell:m} \operatorname{Exp} (\mbfdel{w}_{i,\ell:m}) \nonumber
\eeq
is defined, where $\mbfdel{w}_{i,\ell m} \in \mathbb{R}^9$ is some unknown noise parameter 
associated with the RMI, which is a consequence of the noise associated with every input measurement. 
Despite $\Delta \mathbf{T}_{i, \ell:m}$ being an element of $DE_2(3)$, the above $\operatorname{Exp}$ 
is the exponential operator in $SE_2(3)$. Additionally, a right-perturbation is chosen to match 
the perturbation on $\mbf{U}$ derived in \eqref{eq:Ui_perturbation}, which \jln{simplifies} the 
subsequent derivation, but a left-perturbation could also have been chosen.

Perturbing \eqref{eq:RMI_iter} with respect to the RMI itself then yields
\begin{align*}
    \Delta \mbfbar{T}_{i,\ell:k+1} \operatorname{Exp} (\mbfdel{w}_{i,\ell:k+1}) &= \Delta \mbfbar{T}_{i,\ell:k} \operatorname{Exp} (\mbfdel{w}_{i,\ell:k}) \bar{\mbf{U}}_{i,k} \\
    &= \Delta \mbfbar{T}_{i,\ell:k} \bar{\mbf{U}}_{i,k} \operatorname{Exp} ( \operatorname{Ad}(\bar{\mbf{U}}_{i,k}^{-1}) \mbfdel{w}_{i,\ell:k}),
\end{align*}
which can be simplified to give
\begin{align}
    \label{eq:rmi_cov_rmi}
    \mbfdel{w}_{i,\ell:k+1} = \operatorname{Ad}(\bar{\mbf{U}}_{i,k}^{-1}) \mbfdel{w}_{i,\ell: k}.
\end{align}

Meanwhile, perturbing the RMI relative to the input noise using \eqref{eq:Ui_perturbation} yields
\begin{align*}
    \Delta \mbfbar{T}_{i,\ell:k+1} \operatorname{Exp} (\mbfdel{w}_{i,\ell:k+1}) &= \Delta \mbfbar{T}_{i,\ell:k} \bar{\mbf{U}}_{i,k} \operatorname{Exp} (\mbf{L}_{i,k} \delta \mbf{u}_{i,k}), 
\end{align*}
which can also be simplified to give
\beq
    \label{eq:rmi_cov_noise}
    \mbfdel{w}_{i,\ell:k+1} = \mbf{L}_{i,k} \delta \mbf{u}_{i,k}.
\eeq

\subsection{An Asynchronous-Input Filter}

%Again, 
Taking advantage of the separability of the process model in the neighbour's input measurements, 
an asynchronous-input filter can be designed. The key idea here is to use two process models, one of the form
\beq    
    \label{eq:process_model_no_comms}
    \mbc{T}_{0i,k+1} = \mbf{U}_{0,k}^{-1} \mbf{T}_{0i,k}, \qquad \mbc{T}_{0i,k+1} \in DE_2(3)
\eeq
at $\ell < k < m - 1$ when there is no input information from Robot $i$, and another of the form
\beq
    \label{eq:process_model_w_comms}
    \mbf{T}_{0i,m} = \mbf{U}_{0,m-1}^{-1} \mbc{T}_{0i,m-1} \Delta \mbf{T}_{i,\ell:m}
\eeq
when propagating from $k = m-1$ to $m$ as Robot $i$ communicates the RMI $\Delta \mbf{T}_{i,\ell:m}$. 
Note that $\mbc{T}$ denotes an intermediate state estimate that is not an element of $SE_2(3)$. 
Only when the IMU \jln{measurements} of the neighbouring robot are incorporated does the estimated state 
restore its original $SE_2(3)$ form. 

Given that \eqref{eq:process_model_no_comms} is of the same form as \eqref{eq:process_model_dt} with $\mbf{U}_{i,k} = \mbf{1}$, linearization is straightforward and follows Section \ref{subsec:process_model_linearize}, 
\beq
    \label{eq:state_linearz_no_comms}
    \mbsdel{\xi}_{0i, k+1} = \operatorname{Ad} (\bar{\mbf{U}}_{0,k}^{-1}) \mbsdel{\xi}_{0i, k} - \mbf{L}_{0,k} \mbfdel{u}_{0,k}.
\eeq
Similarly, \eqref{eq:process_model_w_comms} is of the same form as \eqref{eq:process_model_dt} with $\mbf{U}_{i,k} = \Delta \mbf{T}_{i, \ell:m}$, so the linearization with respect to the state is the same as \eqref{eq:state_linearz_no_comms}, giving
\begin{align}
\label{eq:state_linearz_w_preint}
    \mbsdel{\xi}_{0i, m} &= \operatorname{Ad} (\bar{\mbf{U}}_{0,m-1}^{-1}) \mbsdel{\xi}_{0i, m-1} \nonumber \\&\hspace{25pt} - \mbf{L}_{0,m-1} \mbfdel{u}_{0,m-1} + \operatorname{Ad}(\mbfbar{T}_{0i,m}) \mbfdel{w}_{i, \ell:m}.
\end{align}
\ms{A summary of the proposed on-manifold EKF is shown in Algorithm \ref{alg:ekf}.} 

\begin{algorithm}[h!]
    \caption{Algorithm for one time-step of the proposed on-manifold EKF running on Robot 0.}
    \label{alg:ekf}
    \begin{algorithmic}[1]
    \renewcommand{\algorithmicrequire}{}
    \renewcommand{\algorithmicensure}{}
    \REQUIRE \ms{The following is the pseudocode for Robot 0's EKF at time-step $k$. Let $\ell_p$ denote the last time Robot $p \in \{0, \ldots, n\}$ communicated with one of its neighbours. Therefore, at time-step $k-1$, Robot 0 has the RMI $\Delta \mbf{T}_{0, l_0:k-1}$, an intermediate estimate of neighbouring robots' relative poses, $\hat{\mbc{T}}_{0q,k-1}, q \in \{1, \ldots, n\}$, as well as an estimate of the relative clock states. Robot 0 additionally gets an IMU measurement, allowing it to compute $\mbf{U}_{0,k-1}$. The EKF is then as follows.} \vspace{3pt}
    % \ENSURE
    \ms{\STATE Propagate RMI using $\Delta \mbf{T}_{0, l_0:k} = \Delta \mbf{T}_{0, l_0:k-1} \mbf{U}_{0,k-1}$ and its covariance using \eqref{eq:rmi_cov_rmi}, \eqref{eq:rmi_cov_noise}.
    \IF{ranging with neighbour $i$}
        \STATE Communicate $\Delta \mbf{T}_{0, l_0:k}$ and its covariance.
        \STATE Generate 5 pseudomeasurements using \eqref{eq:y_r}-\eqref{eq:y_p3}.
        \STATE Propagate the relative pose state estimates in time using
            \begin{align*}
                \check{\mbc{T}}_{0p,k} &= \mbf{U}_{0,k-1}^{-1} \hat{\mbc{T}}_{0p,k-1}, \qquad p \in \{1, \ldots, n\}, p \neq i, \\
                \mbfcheck{T}_{0i,k} &= \mbf{U}_{0,k-1}^{-1} \hat{\mbc{T}}_{0i,k-1} \Delta \mbf{T}_{i,\ell_i:k},
            \end{align*}
            and the clock state estimates using Section~\ref{subsec:discrete_clock_process_model}.
        \STATE Propagate the state-error covariances using \eqref{eq:state_linearz_no_comms}, \eqref{eq:state_linearz_w_preint} and Section~\ref{subsec:discrete_clock_process_model}.
        \STATE Do an on-manifold EKF correction step \cite{Sola2018} using the pseudomeasurements to get $\hat{\mbc{T}}_{0p,k}$ and $\mbfhat{T}_{0i,k}$.
        \STATE Initiate a new RMI $\Delta \mbf{T}_{0, k:k} = \mbf{1}$ with covariance $\mbf{0}$.
    \ELSIF{neighbours $i$ and $j$ are ranging}
        \STATE Generate 8 pseudomeasurements using \eqref{eq:y_r}-\eqref{eq:y_p3}.
        \STATE Propagate the relative pose state estimates in time using
            \begin{align*}
                \check{\mbc{T}}_{0p,k} &= \mbf{U}_{0,k-1}^{-1} \hat{\mbc{T}}_{0p,k-1}, \quad p \in \{1, \ldots, n\}, p \neq i, j, \\
                \mbfcheck{T}_{0i,k} &= \mbf{U}_{0,k-1}^{-1} \hat{\mbc{T}}_{0i,k-1} \Delta \mbf{T}_{i,\ell_i:k}, \\
                \mbfcheck{T}_{0j,k} &= \mbf{U}_{0,k-1}^{-1} \hat{\mbc{T}}_{0j,k-1} \Delta \mbf{T}_{j,\ell_j:k},
            \end{align*}
            and the clock state estimates using Section~\ref{subsec:discrete_clock_process_model}.
        \STATE Propagate the state-error covariances using \eqref{eq:state_linearz_no_comms}, \eqref{eq:state_linearz_w_preint} and Section~\ref{subsec:discrete_clock_process_model}.
        \STATE Do an on-manifold EKF correction step \cite{Sola2018} using the pseudomeasurements to get $\hat{\mbc{T}}_{0p,k}$, $\mbfhat{T}_{0i,k}$, and $\mbfhat{T}_{0j,k}$.
    \ELSIF{no one is ranging}
        \STATE Propagate the relative pose state estimates in time using
            \begin{align*}
                \check{\mbc{T}}_{0p,k} &= \mbf{U}_{0,k-1}^{-1} \hat{\mbc{T}}_{0p,k-1}, \qquad p \in \{1, \ldots, n\},
            \end{align*}
            and the clock state estimates using Section~\ref{subsec:discrete_clock_process_model}.
        \STATE Propagate the state-error covariances using \eqref{eq:state_linearz_no_comms} and Section~\ref{subsec:discrete_clock_process_model}.
    \ENDIF}
    \end{algorithmic}
\end{algorithm}

\subsection{Equivalence to the No Communication Constraint Case}

\jln{In the absence of} communication constraint, each robot would have access to all its neighbours' IMU measurements 
at all times. As explained in Section \ref{subsec:need_for_preintegration}, this is not possible, %which then necessitates preintegration. 
\jln{so that preintegration is needed.}
It is shown in \eqref{eq:process_model_w_rmi} that the state can be propagated using RMIs in a manner equivalent 
to the case with no communication constraint. In this subsection, it is shown that \jln{computing the uncertainty propagation for} 
the state is also equivalent \jln{in both cases}, despite the Jacobians used %to compute uncertainty 
\jln{being} different. 
%for the case with preintegration and the case with no communication constraints. 
This is in fact a consequence of the structure of the Jacobians when perturbing the state from the left.

\subsubsection{No Communication Constraints}

When there are no communication constraints and IMU measurements of neighbours are available at all times, 
the models shown in Section \ref{sec:process_model} can be used to propagate the state. The covariance of the state 
is propagated using \eqref{eq:process_model_cov_state} and \eqref{eq:process_model_cov_noise}, which for two 
nonadjacent timestamps $\ell$ and $m$ would \ms{be written as} 
\begin{align}
    &\mbsdel{\xi}_{0i,m} \nonumber \\
    &= \operatorname{Ad}(\Delta \mbf{T}_{0,\ell:m})^{-1}\mbsdel{\xi}_{0i,\ell} 
    %\nonumber  \\
    %
    %&\hspace{12pt} 
    - \sum_{k=\ell}^{m-1}\operatorname{Ad}(\Delta \mbf{T}_{0,k+1:m})^{-1} \mbf{L}_{0,k} \mbfdel{u}_{0,k} \nonumber  \\
    &\hspace{12pt} + \sum_{k=\ell}^{m-1}\operatorname{Ad}(\Delta \mbf{T}_{0,k+1:m}^{-1} \mbfbar{T}_{0i,k+1})\mbf{L}_{i,k} \mbfdel{u}_{i,k} \label{eq:cov_propagation_l_to_m}.
\end{align}

\subsubsection{With Preintegration}

First, the uncertainty of the RMI can be computed using \eqref{eq:rmi_cov_rmi} and \eqref{eq:rmi_cov_noise} as
\beq 
    \mbfdel{w}_{i,\ell:m} = \sum_{k=\ell}^{m-1} \operatorname{Ad}(\Delta \mbf{T}_{i,k+1:m})^{-1} \mbf{L}_{i,k} \delta \mbf{u}_{i,k}. \nonumber
\eeq

Note that the RMI gets communicated at time-step $m$, so from timestep $\ell$ to $m-1$ the state propagation occurs only with the IMU measurements of Robot 0 as shown in \eqref{eq:process_model_no_comms}. The uncertainty propagation from timestamp $\ell$ to $m-1$ then follows as per \eqref{eq:state_linearz_no_comms}, \ms{which can be written as} 
\begin{align*}
    \mbsdel{\xi}_{0i, m-1} &= \operatorname{Ad}(\Delta \mbf{T}_{0,\ell:m-1})^{-1} \mbsdel{\xi}_{0i, \ell} \nonumber  \\
    &\hspace{12pt} - \sum_{k=\ell}^{m-2}\operatorname{Ad}(\Delta \mbf{T}_{0,k+1:m-1})^{-1} \mbf{L}_{0,k} \mbfdel{u}_{0,k}.
\end{align*}
Meanwhile, propagating the uncertainty from timestamp $m-1$ to $m$ using the RMI as shown in \eqref{eq:process_model_w_comms} 
then follows as per \eqref{eq:state_linearz_w_preint} to give
\begin{align*}
    &\mbsdel{\xi}_{0i, m} \\
    &= \operatorname{Ad}(\Delta \mbf{T}_{0,\ell:m})^{-1}\mbsdel{\xi}_{0i,\ell} 
    %\nonumber  \\
    %%
    %&\hspace{12pt} 
    - \sum_{k=\ell}^{m-1} \operatorname{Ad}(\Delta \mbf{T}_{0,k+1:m})^{-1} \mbf{L}_{0,k} \mbfdel{u}_{0,k} \nonumber \\
    %
    %&\hspace{10pt} + \operatorname{Ad}(\Delta \mbf{T}_{0,\ell:m}^{-1} \mbfbar{T}_{0i,\ell} \Delta \mbf{T}_{i,\ell:m}^{-1}) %\mbfdel{w}_{i,\ell:m} \\
    &\hspace{10pt} + \operatorname{Ad}(\Delta \mbf{T}_{0,\ell:m}^{-1} \mbfbar{T}_{0i,\ell} \Delta \mbf{T}_{i,\ell:m}) \mbfdel{w}_{i,\ell:m} \\
    &= \operatorname{Ad}(\Delta \mbf{T}_{0,\ell:m})^{-1}\mbsdel{\xi}_{0i,\ell} 
    %\nonumber  \\
    %
    %&\hspace{12pt} 
    - \sum_{k=\ell}^{m-1} \operatorname{Ad}(\Delta \mbf{T}_{0,k+1:m})^{-1} \mbf{L}_{0,k} \mbfdel{u}_{0,k} \nonumber \\
    &\hspace{10pt} + \sum_{k=\ell}^{m-1} \operatorname{Ad}(\Delta \mbf{T}_{0,k+1:m}^{-1} \Delta \mbf{T}_{0,\ell:k+1}^{-1} \mbfbar{T}_{0i,\ell} %\Delta \mbf{T}_{i,\ell:k+1}^{-1}) \mbf{L}_{i,k} \delta \mbf{u}_{i,k},
    \Delta \mbf{T}_{i,\ell:k+1}) \mbf{L}_{i,k} \delta \mbf{u}_{i,k},
\end{align*}
which, \jln{using \eqref{eq:process_model_w_rmi}}, simplifies to be exactly equal to \eqref{eq:cov_propagation_l_to_m}.

\subsection{Communication Requirements}

The proposed multi-robot preintegration approach provides an alternative efficient way of communicating odometry 
information as compared to communicating the individual IMU measurements. 
%Nonetheless, 
When sending IMU measurements, no covariance information is required as the covariance matrix is typically 
a fixed value that can be assumed common among all robots if they all share the same kind of IMU. 
Meanwhile, when sending an RMI, %information, 
\jln{the components of} a corresponding $9 \times 9$ positive-definite symmetric matrix 
representing its computed uncertainty must also be sent, as this is not constant but rather 
a function of the individual inputs.

Each IMU measurement consists of 6 single-precision floats, 3 for the gyroscope and 3 for the accelerometer readings, 
for a total of 24 bytes. Meanwhile, each RMI can be represented using 10 single-precision floats and the corresponding 
covariance matrix using the upper triangular part of the $9 \times 9$ matrix, which requires communicating an additional 
45 single-precision floats. Therefore, sending one RMI and its covariance matrix requires over 220 bytes of information. 
Therefore, unless an RMI replaces more than 9 IMU measurements, it is sometimes more %communicationally-
\jln{efficient to communicate} the raw IMU measurements. 
Nonetheless, using the proposed multi-robot preintegration framework has the following advantages \ms{(in addition to the discussion in Section \ref{subsec:need_for_preintegration})}.
\begin{itemize}
    \item It overcomes the need for variable amount of communication, as the RMI and its covariance 
    matrix are of fixed length but a varying number of IMU readings might be accumulated in between 
    two instances of a robot ranging. This consequently eases implementation and provides a more reliable system.
    \item It provides robustness to loss of communication, as a robot re-establishing communication 
    with its neighbours after a few seconds would not be able to send over all the accumulated IMU information.
    \item It reduces the amount of processing required at neighbours, as the input matrices $\mbf{U}_{i,k}$ 
    are pre-multiplied at Robot $i$ on behalf of all its neighbours.
    \item It overcomes the need to know the noise distribution of the neighbours' IMUs, which would 
    be useful if not all robots had the same IMU.
    \item It allows easy integration with IMU-bias estimators and approaches that dynamically tune 
    the covariance of the IMU measurements, without needing to send the bias estimates or the tuned 
    covariances over UWB. 
\end{itemize}

Additionally, UWB protocols by default allow 128 bytes of information to be sent per message 
transmission \cite{802-15-4a}, for a total of 256 bytes per transceiver in each TWR instance. 
Given that each transceiver only needs to send 2 bytes of frame-control data per signal 
(thus 4 bytes of frame-control data in total) \cite{802-15-4a} and a total of 3 single-precision 
timestamps (thus 12 bytes of timestamps), there is enough room for the 220 bytes required to send an RMI. 
Note that if more information is required, some modules such as DW1000 allow up to 1024 
bytes of data per message transmission \cite{dw1000}.

%% file: sections/simulation.tex
\begin{table}
    \renewcommand{\arraystretch}{1.2}
    \footnotesize
    \caption{Simulation parameters \ms{based on the ICM-20689 IMU and the DWM1000 UWB transceiver}.}
    \label{tab:sim_params}
    \centering
    \begin{tabular}{c|c}
    % \hline
    \bfseries Specification & \bfseries Value\\
    \hline
    Accelerometer std. dev. [m/s$^2$] & 0.023 \\
    Gyroscope std. dev. [rad/s] & 0.0066 \\
    IMU rate [Hz] & 250 \\
    UWB timestamping std. dev. [ns] & 0.33 \\
    UWB rate [Hz] & 125 \\
    Clock offset PSD [ns$^2$/Hz] & 0.4 \\
    Clock skew PSD [ppb$^2$/Hz] & 640 
    % \hline
    \end{tabular}
\end{table}

\begin{figure*}[b!]
    \centering
    \begin{minipage}{0.45\textwidth}%
        \centering
        \subfloat[Subfigure 1 list of figures text][Centralized.]{
        \includegraphics[width=\textwidth]{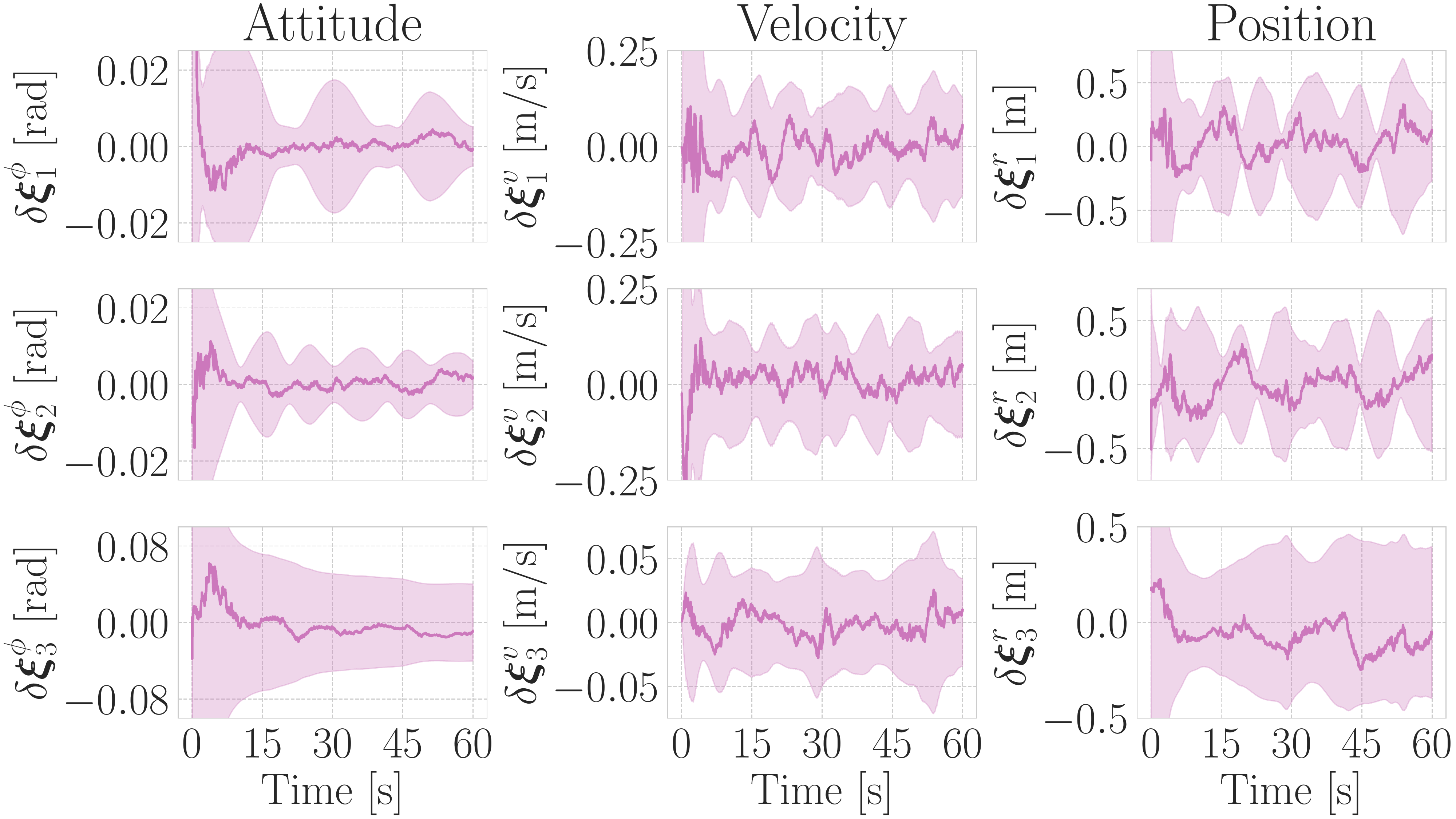}
        \label{fig:sim_pose_3sigma_c}}
    \end{minipage}\quad%
    \begin{minipage}{0.45\textwidth}%
        \centering
        \subfloat[Subfigure 2 list of figures text][With passive listening.]{
        \includegraphics[width=\textwidth]{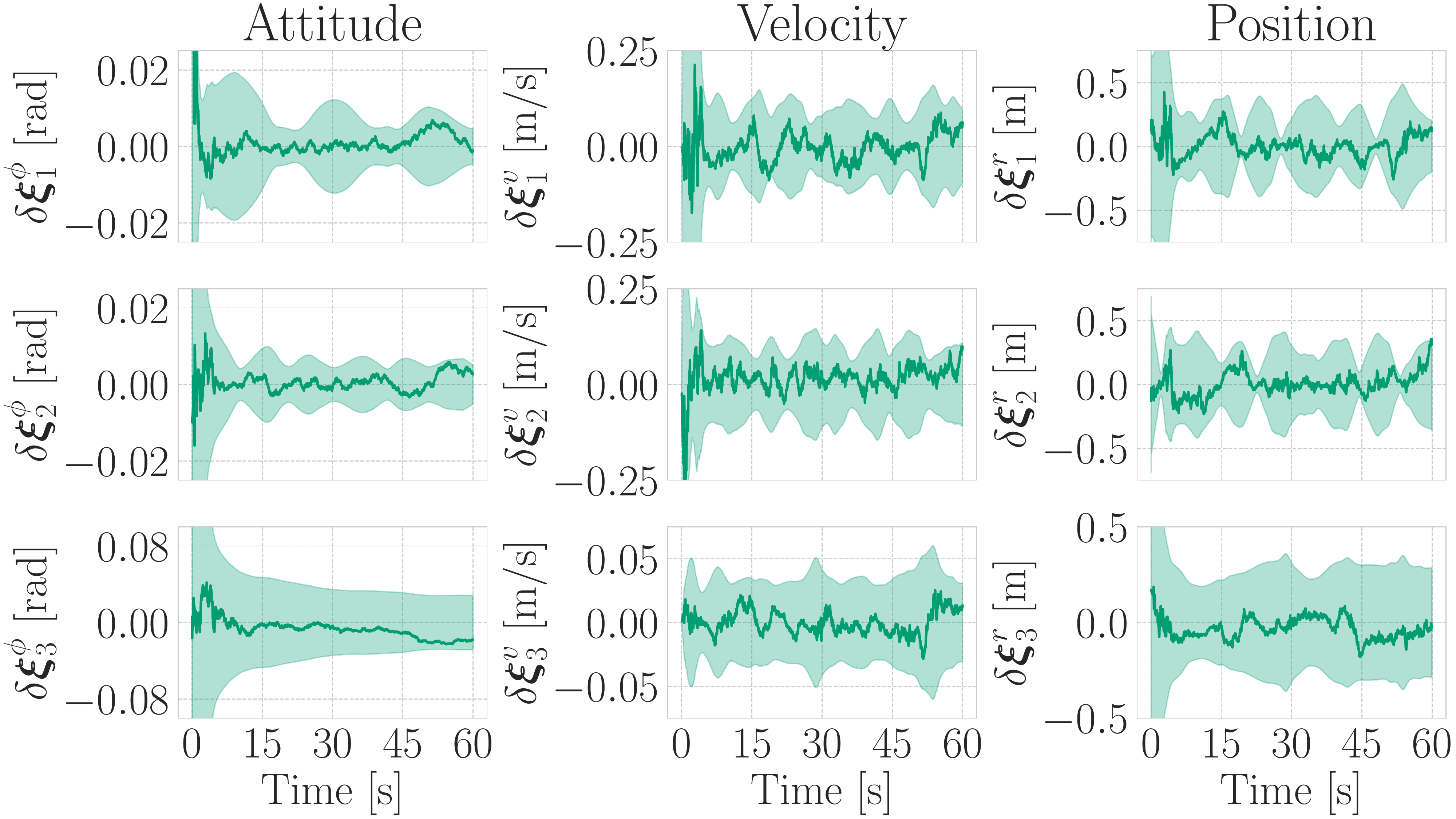}
        \label{fig:sim_pose_3sigma_wp_z}}
    \end{minipage}
    \caption{\tro{Error plots and $\pm3\sigma$ bounds (shaded region) for Robot 0's estimate of Robot 1's relative pose for Simulation S1, comparing the centralized and proposed approaches.}}
    \label{fig:sim_pose_centralized}
\end{figure*}

To evaluate the benefits of using passive listening on the estimation accuracy of relative pose states, the clock dynamics and quadcopter kinematics have been simulated. The clocks' evolution is modelled relative to a ``global time'' using the simulating computer's own clock, while the absolute-state quadcopter kinematics are simulated relative to some inertial frame. Noisy IMU and timestamp measurements are then modelled and fed into the CSRPE algorithm to estimate the relative clock and pose states.

To evaluate the proposed approach, 3 datasets are simulated.
\begin{enumerate}
    \item \textbf{S1}: A single run with \tro{4 quadcopters},
    \item \textbf{S2}: 100 Monte-Carlo trials with \tro{3 to 7 quadcopters}, and
    \item \textbf{S3}: 500 Monte-Carlo trials with \tro{$4$ quadcopters}.
\end{enumerate}
The trajectory of the quadcopters in the case of \tro{a system with 3 robots} is shown in Figure~\ref{fig:sim_trajectory}, and the simulation parameters are shown in Table~\ref{tab:sim_params}. \tro{The simulated trajectories are 60 seconds long and each quadcopter covers a distance between 60 m and 218 m, with a maximum speed of 5.5 m/s. The maximum and mean angular velocities are 1 rad/s and 0.3 rad/s, respectively.}
\jln{Following a periodic sequence, each pair of transceivers performs 
in turn a ranging transaction, except for pairs of transceivers 
on the same robot.}
%All the transceivers range with one another one pair at a time in sequence, 
%while transceivers on the same robot do not range with one another. 
The proposed algorithm is then %implemented using 
\jln{tested on}
each dataset \tro{and %is 
compared to two scenarios.
\begin{enumerate}
    \item \textbf{Centralized}: A hypothetical centralized scenario where each robot has access to range measurements between neighbours in the absence of passive listening. This differs from the proposed framework in that the pseudomeasurements associated with passive listening do not exist, and that this is practically impossible without passive listening or some other communication media. This serves as the benchmark on what is the best achievable estimator using existing methods. 
    \item \textbf{No passive listening}: A decentralized approach but in the absence of passive listening, meaning that robots do not have access to the passive listening pseudomeasurements nor range measurements between neighbours. This serves as the benchmark on what is currently a practically implementable solution without requiring a central processor or additional communication media.
\end{enumerate}
}

\begin{figure*}[t!]
    \centering
    \begin{minipage}{0.45\textwidth}%
        \centering
        \subfloat[Subfigure 1 list of figures text][No passive listening.]{
        \includegraphics[width=\textwidth]{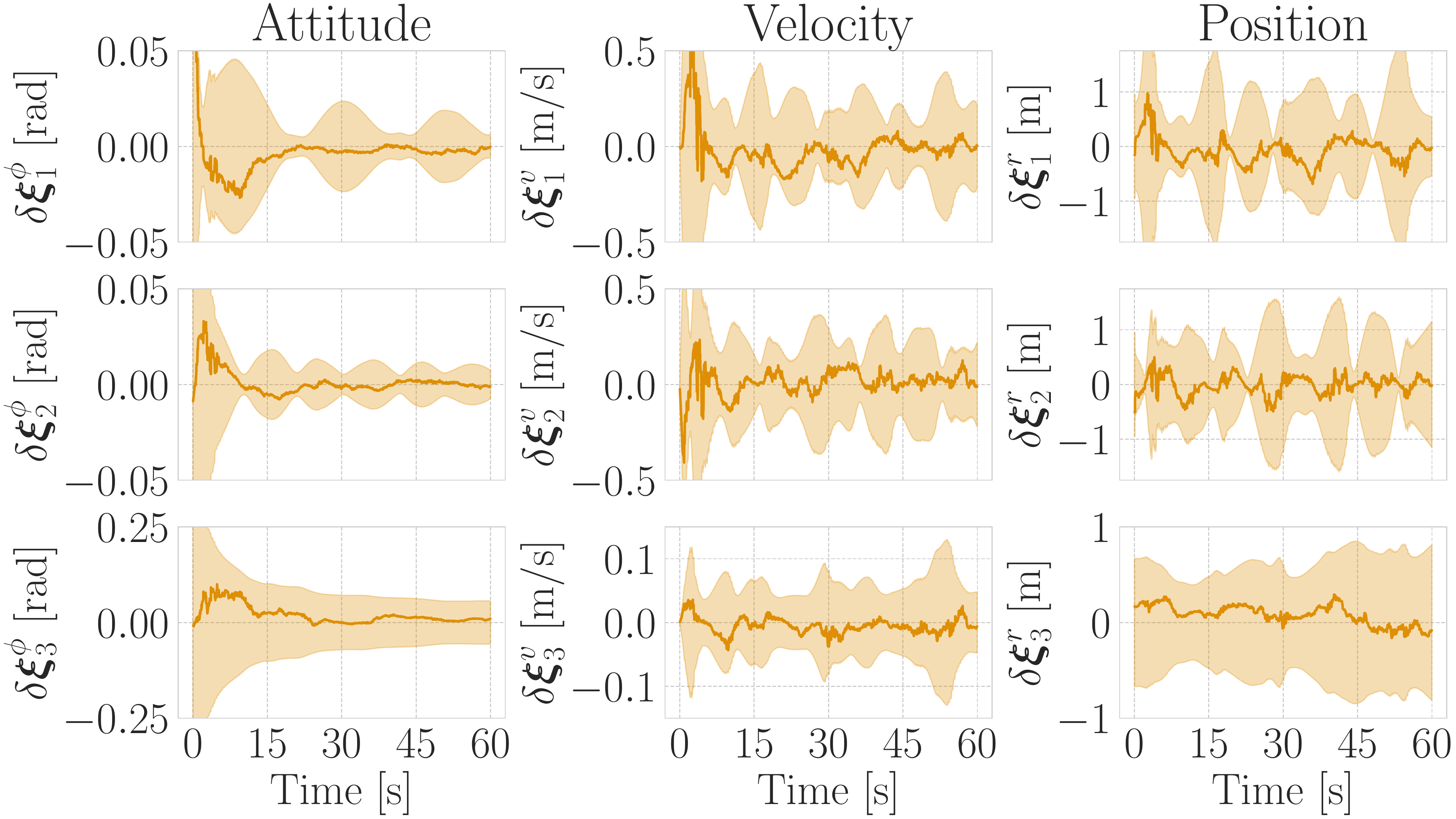}
        \label{fig:sim_pose_3sigma_np}}
    \end{minipage}\quad%
    \begin{minipage}{0.45\textwidth}%
        \centering
        \subfloat[Subfigure 2 list of figures text][With passive listening.]{
        \includegraphics[width=\textwidth]{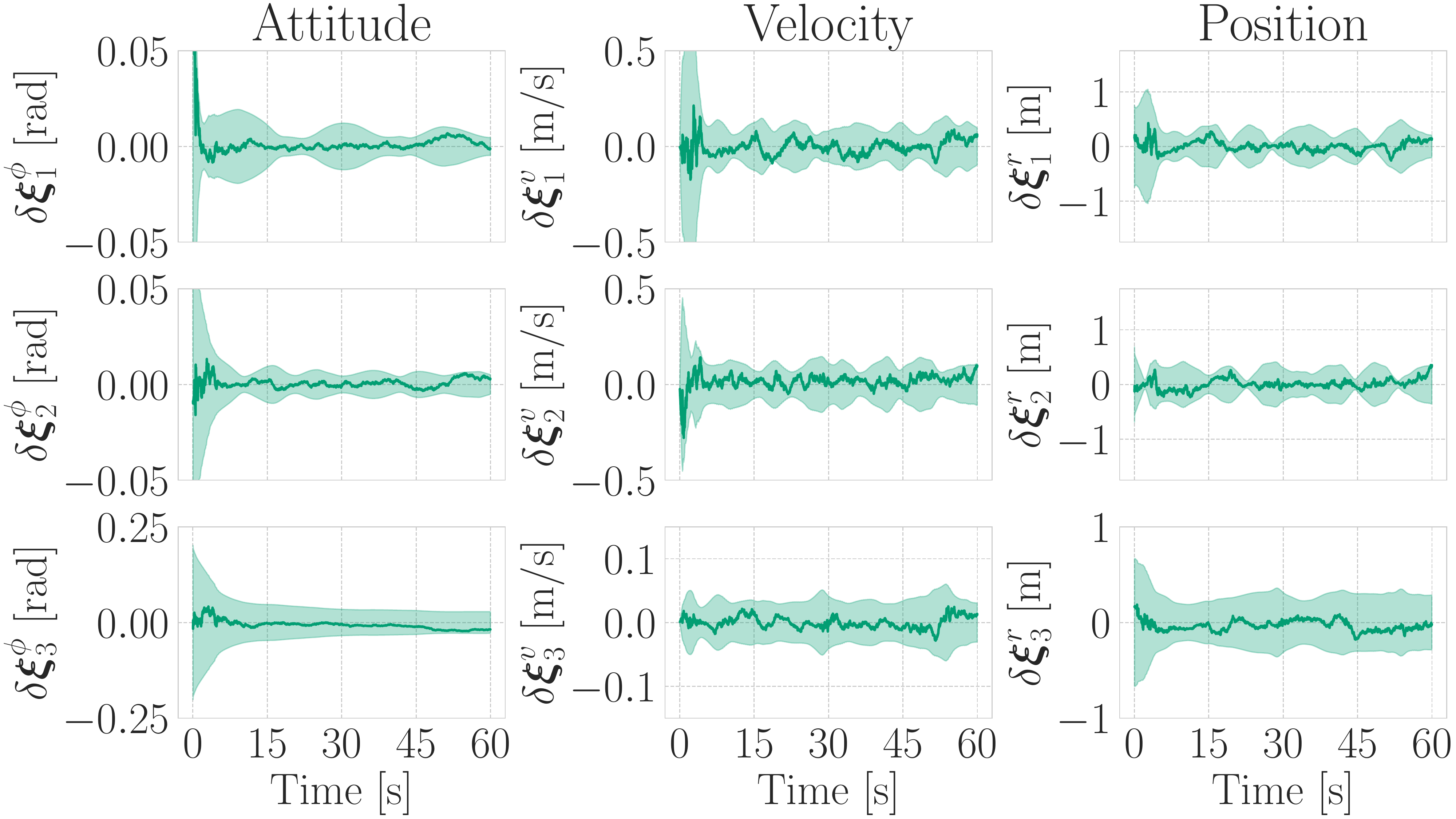}
        \label{fig:sim_pose_3sigma_wp}}
    \end{minipage}
    \caption{Error plots and $\pm3\sigma$ bounds (shaded region) for Robot 0's estimate of Robot 1's relative pose for Simulation S1, \tro{comparing the decentralized no-passive-listening and proposed approaches. The right figure is a zoomed-out version of Figure \ref{fig:sim_pose_3sigma_wp_z}.}}
    \label{fig:sim_pose}
\end{figure*}

\begin{figure*}[t!]
    \centering
    \begin{minipage}{0.45\textwidth}%
        \centering
        \subfloat[Subfigure 1 list of figures text][No passive listening.]{
        \includegraphics[width=\textwidth]{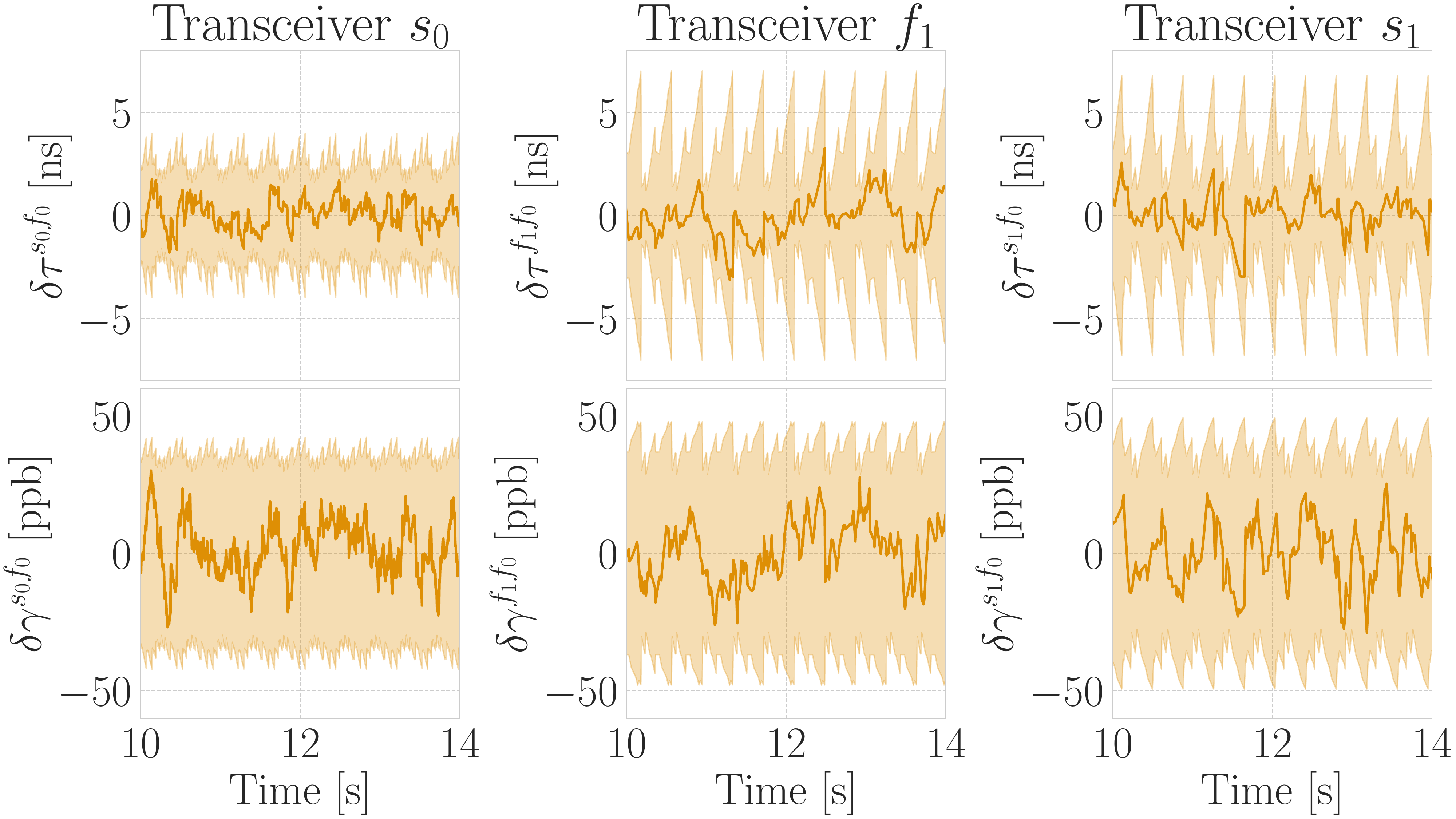}
        \label{fig:sim_clock_3sigma_np}}
    \end{minipage}\quad%
    \begin{minipage}{0.45\textwidth}%
        \centering
        \subfloat[Subfigure 2 list of figures text][With passive listening.]{
        \includegraphics[width=\textwidth]{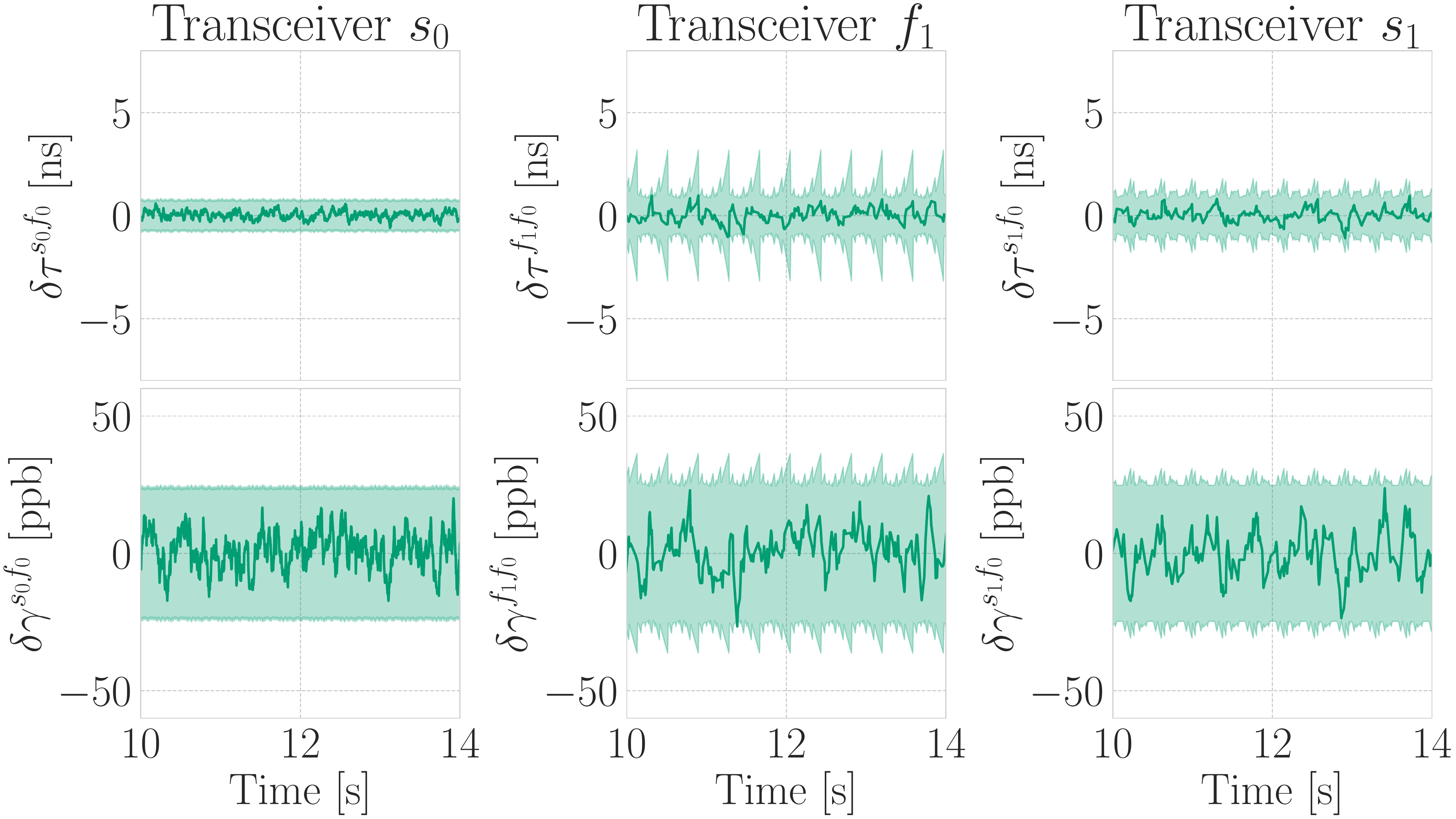}
        \label{fig:sim_clock_3sigma_wp}}
    \end{minipage}
    \caption{Error plots and $\pm3\sigma$ bounds (shaded region) for Robot 0's estimate of the clock states of Transceivers $s_0$, $f_1$, and $s_1$ relative to Transceiver $f_0$ for Simulation S1. \ms{These plots are zoomed in to a window of 4 seconds to show clearly the cycle of expanding and contracting uncertainty in the clock estimates as the transceiver alternates between active ranging and passive listening.}}
    \label{fig:sim_clock}
\end{figure*}

The evaluation is based on the following three criteria.
\begin{enumerate}
    \item Accuracy: The accuracy of the proposed algorithm as compared to the case with no passive listening is quantified using error plots and the \emph{root-mean-squared-error} (RMSE), which for the \ms{pose} estimation error \ms{$\mbf{e}_k = \operatorname{Log}(\mbfhat{T}_k\mbf{T}_k^{-1})$} is computed as 
        \beq
            \text{RMSE} \triangleq \sqrt{\f{1}{N+1} \sum_{k=0}^{N} \ms{\mbf{e}^\trans_k \mbf{e}_k}} \nonumber
        \eeq
        for $N+1$ time-steps.
    \item Precision: The precision of the proposed algorithm is quantified using $\pm 3\sigma$-bound regions \ms{about the estimate}, which represent a 99.73\% confidence bound \jln{under a Gaussian distribution assumption}.
    \item Consistency: A consistent estimator is an estimator with a modelled precision that reflects the true precision of its estimate. In more specific terms, a consistent estimator outputs a covariance matrix on its estimate that is representative of the true uncertainty of that estimate. Consistency is evaluated using the \emph{normalized-estimation-error-squared} (NEES) test \cite[Section 5.4]{barshalom2002}.
\end{enumerate}

\subsection{Estimation Accuracy and Precision}

\begin{figure}
    \centering
    \includegraphics[width=0.85\columnwidth]{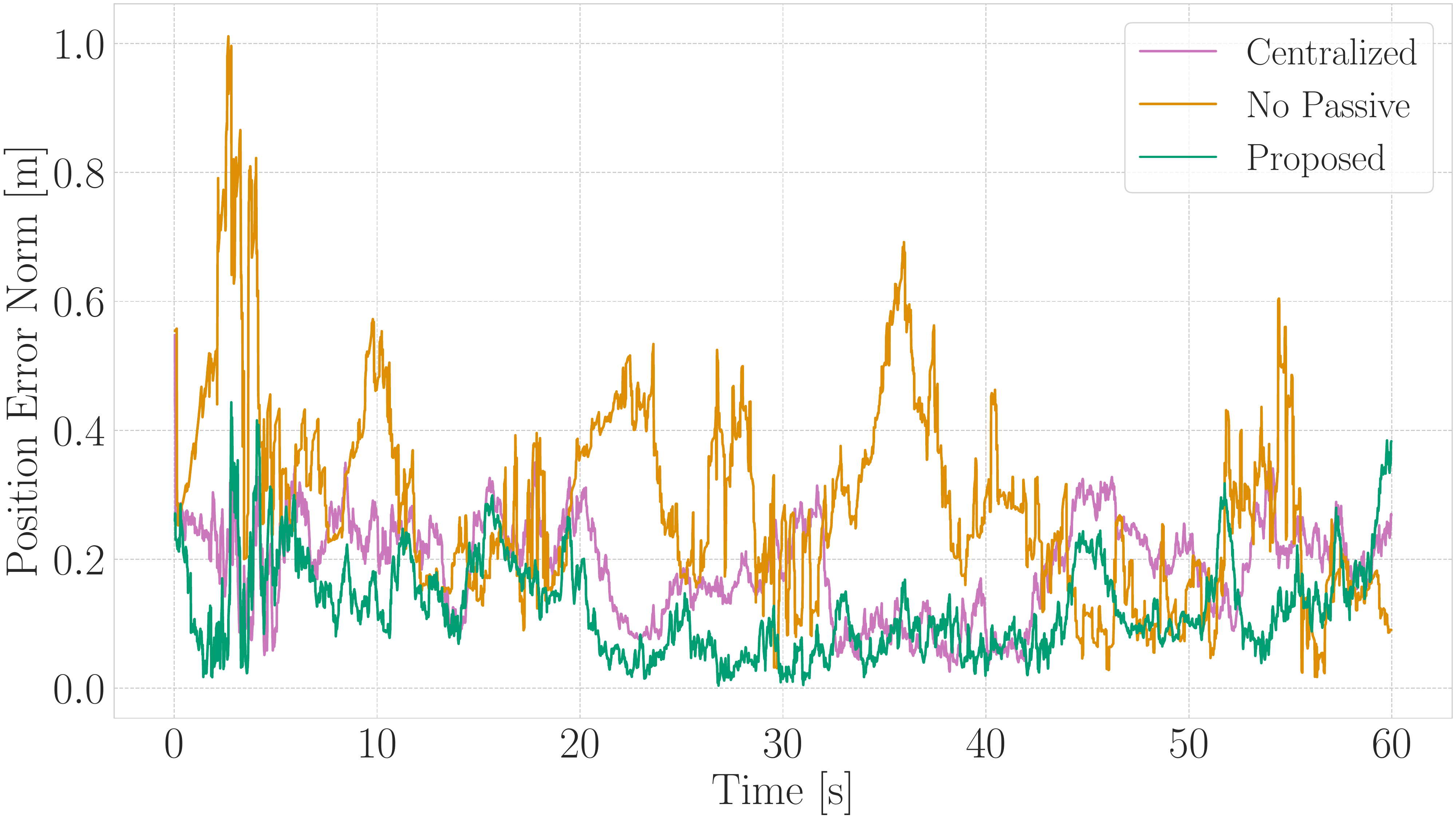}
    \caption{\tro{The error norm for Robot 0's estimate of Robot 1's relative pose for Simulation S1.}}
    \label{fig:sim_rmse}
\end{figure}

The error plots for the relative pose estimate of Robot 1 relative to Robot 0 in Simulation S1 are shown in \tro{Figures \ref{fig:sim_pose_centralized} and \ref{fig:sim_pose}. Passive listening reduces the positioning RMSE by 29.4\% from 0.204 m to 0.144 m as compared to the centralized approach, and by 55.96\% from 0.327 m to 0.144 m when compared to the case of no passive listening. Additionally, passive listening \jln{produces at almost every time-step} a position error with smaller norm, as shown in Figure \ref{fig:sim_rmse}. The proposed estimator is also significantly more confident in its estimate, as shown by the covariance bounds in Figures \ref{fig:sim_pose_centralized} and \ref{fig:sim_pose}. }

This improvement in localization performance can be attributed to more measurements and stronger cross-correlation between the different states when passive listening measurements are available. As shown in Figure \ref{fig:sim_clock}, passive listening results in the clock state of a transceiver
%Transceiver $i$ 
not drifting significantly in between instances where this transceiver is ranging. 
This brings down the clock offset RMSE of Transceiver $f_1$ for example by 59.31\% from 1.155 ns to 0.470 ns \tro{when compared to the case with no passive listening}.

The improvement in performance can also be seen %in more scenarios 
\jln{as the number of robots is increased,}
%when there is a varying number of neighbours, 
as shown in Table \ref{tab:sim_rmse} for the Simulation S2. Because only one pair of transceivers can communicate 
at a time, in the absence of passive listening \jln{the rate at which each transceiver participates in a ranging transaction
decreases with the number of transceivers,}
%participates in fewer ranging instances as the number of transceivers increase. 
%Therefore, the number of measurements obtained from every neighbour is reduced, 
and \jln{as a result the overall} localization performance degrades. 
% as the number of robots increases. %since the UWB communication space gets overcrowded. 
With passive listening \jln{on the other hand}, adding robots does not result 
in longer periods without measurements \ms{and the measurement rate per robot remains the same}. 
%windows of no measurements 
In fact, \ms{it turns out that adding robots in the presence of passive listening produces better performance} due to spatial variations in the range-measurement sources \cite{Cossette2022}. \tro{This is also the case for the centralized estimator.}

\begin{figure*}[t!]
    \centering
    \begin{minipage}{0.9\columnwidth}%
        \centering
        \subfloat[Subfigure 1 list of figures text][Position RMSE.]{
        \includegraphics[width=\textwidth]{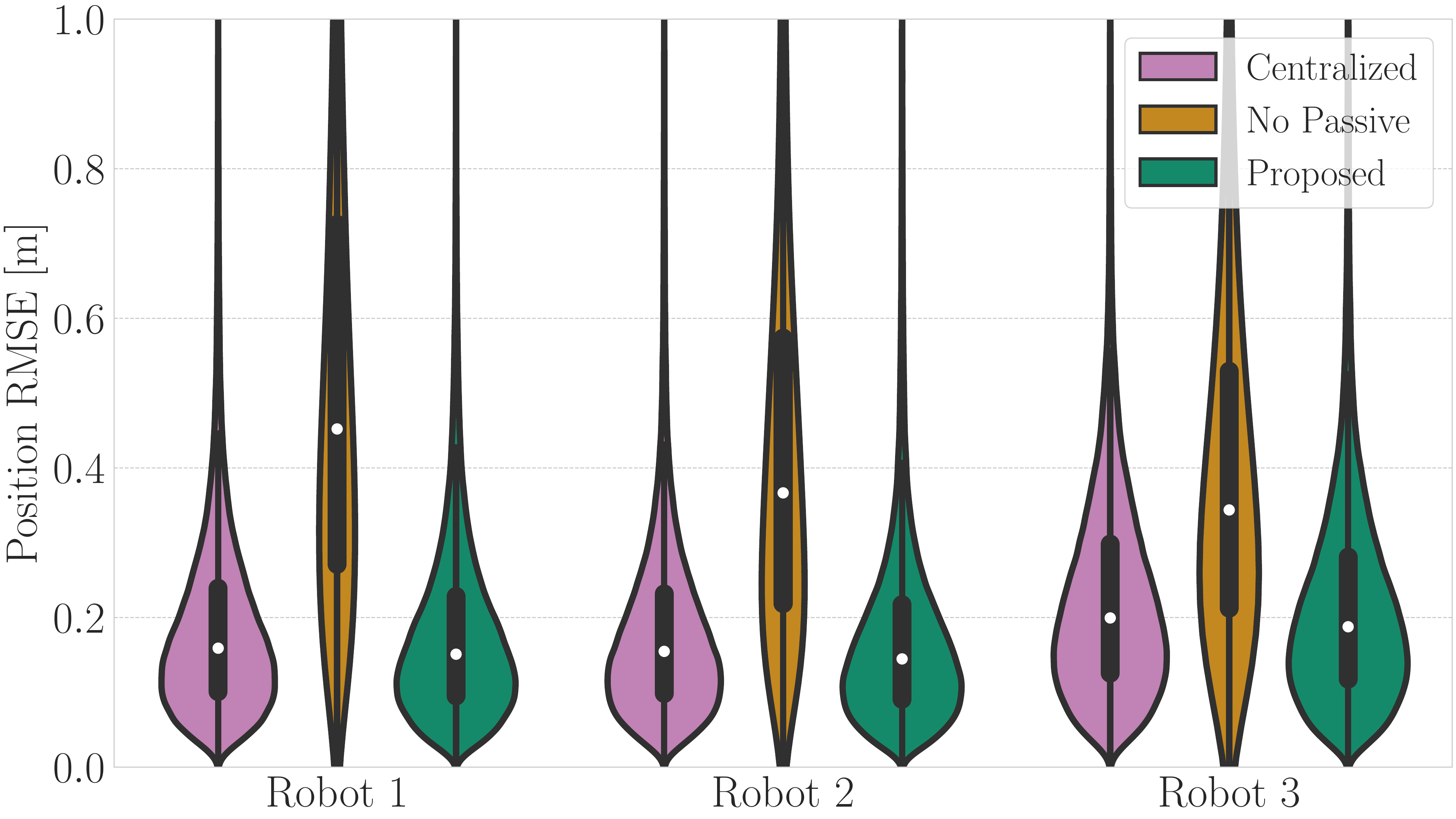}
        \label{fig:pos_rmse_violin}}
    \end{minipage}%
    \qquad
    \begin{minipage}{0.9\columnwidth}%
        \centering
        \subfloat[Subfigure 2 list of figures text][Attitude RMSE.]{
        \includegraphics[width=\textwidth]{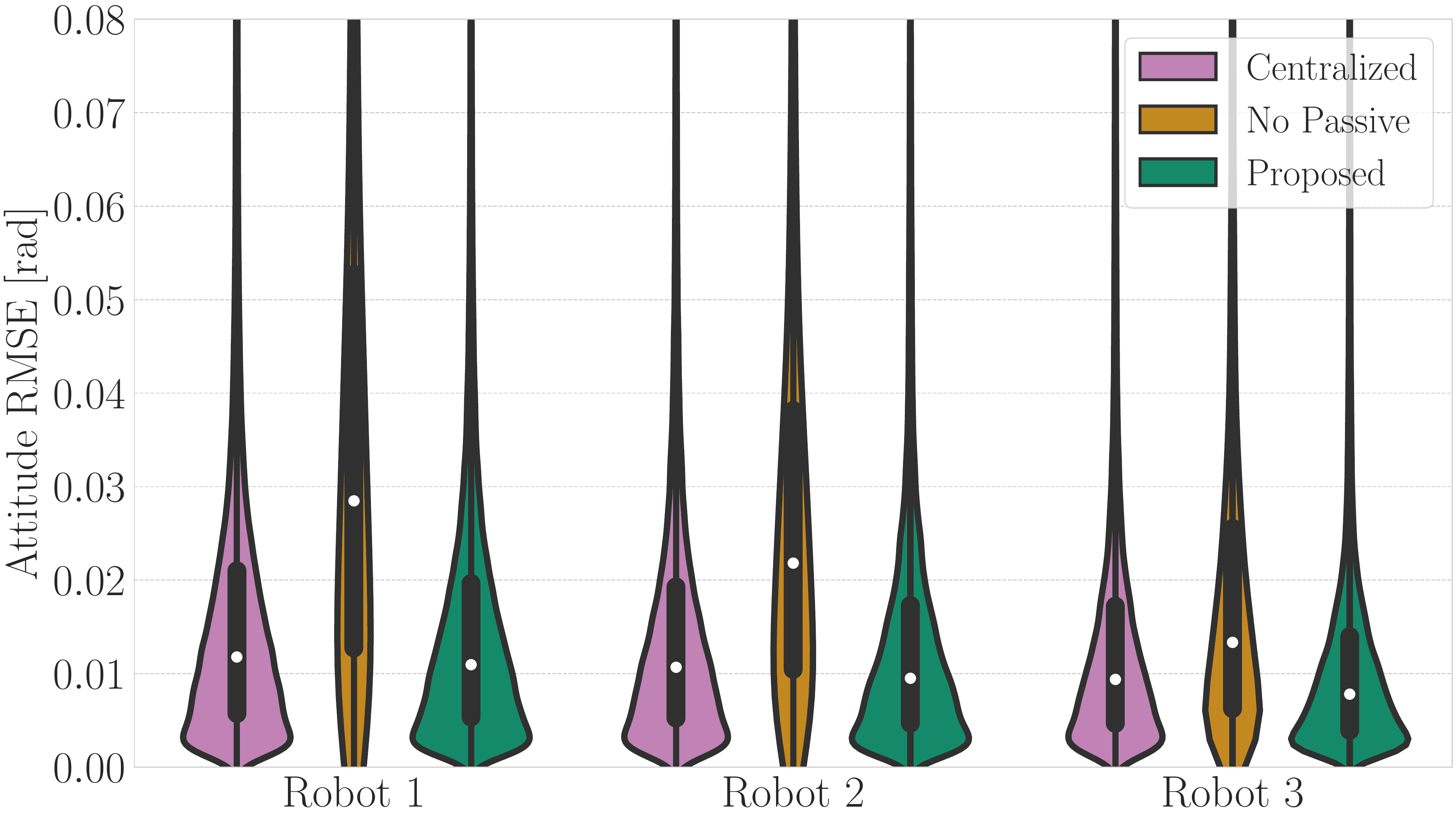}
        \label{fig:att_rmse_violin}}
    \end{minipage}\quad%
    \caption{\tro{Violin and box plots showing the distribution of the position and attitude RMSEs for Simulation S3. The envelope shows the relative frequency of RMSE values. The box plot shows the median as a white dot, while the first and third quartile of the data are represented using the lower and upper bound of the thick black bar, respectively.}}
    \label{fig:rmse_violin}
\end{figure*}

To provide further insight into the contribution of passive 
listening measurements on the behaviour of the estimator, 
the distribution of the RMSEs of the position and attitude 
estimates of all robots in Simulation S3 are visualized 
in Figure \ref{fig:rmse_violin}. \tro{Not only does the proposed approach significantly outperform the no passive listening approach, but it matches the centralized approach, which is typically the best possible solution under an assumption of the availability of a central processor. In fact, the proposed framework slightly outperforms the standard centralized approach due to the availability of additional pseudomeasurements.}

\newcolumntype{D}{>{\centering\arraybackslash}m{1cm}}
\newcolumntype{E}{>{\centering\arraybackslash}m{2.5cm}}
\newcolumntype{F}{>{\centering\arraybackslash}m{3.8cm}}
% \newcolumntype{D}{>{\centering\arraybackslash}m{1.5cm}}
\begin{table}[h!]
    \footnotesize
    \renewcommand{\arraystretch}{1.2}
    \caption{\tro{The average RMSE (aRMSE) for all trials of Robot 0's estimate of neighbouring robots' relative pose for Simulation S2. The percentage change is $\f{\text{Proposed} - \text{Comparison}}{\text{Comparison}}$, where the Comparison is either Centralized or No Passive.}}
    \label{tab:sim_rmse}
    \centering
    \begin{tabular}{D|DDD|DD}
    % \hline
    & \multicolumn{3}{F}{\bfseries Position aRMSE averaged over all Robots [m]} & \multicolumn{2}{E}{\bfseries Percentage change [\%]}\\
    \hline 
    \bfseries Number of Robots & \bfseries Centr. & \bfseries No Passive & \bfseries Proposed & \bfseries Centr. & \bfseries No Passive\\
    \hline
    3 & 0.277 & 0.486 & \bfseries 0.263 & -5.05 & -45.88 \\
    4 & 0.231 & 0.574 & \bfseries 0.222 & -3.90 & -61.32 \\
    5 & 0.220 & 0.662 & \bfseries 0.211 & -4.09 & -68.13 \\
    6 & 0.220 & 0.737 & \bfseries 0.199 & -9.55 & -73.00 \\
    7 & 0.186 & 0.917 & \bfseries 0.165 & -11.29 & -82.01
    % \hline
    \end{tabular}
\end{table}

\subsection{Consistency}

Given that the estimator is an EKF, consistency cannot be guaranteed due to linearization and discretization errors. Nonetheless, the proposed on-manifold framework can characterize banana-shaped error distributions that result from range measurements as shown in Figure \ref{fig:banana} more efficiently. Consequently, the error distribution appears to be well-characterized by the estimator as shown in Figures \ref{fig:sim_pose_centralized}, \ref{fig:sim_pose}, and \ref{fig:sim_clock}, as the error trajectory typically lies within the $\pm 3\sigma$ bounds. 

\begin{figure}
    \centering
    \includegraphics[width=0.85\columnwidth]{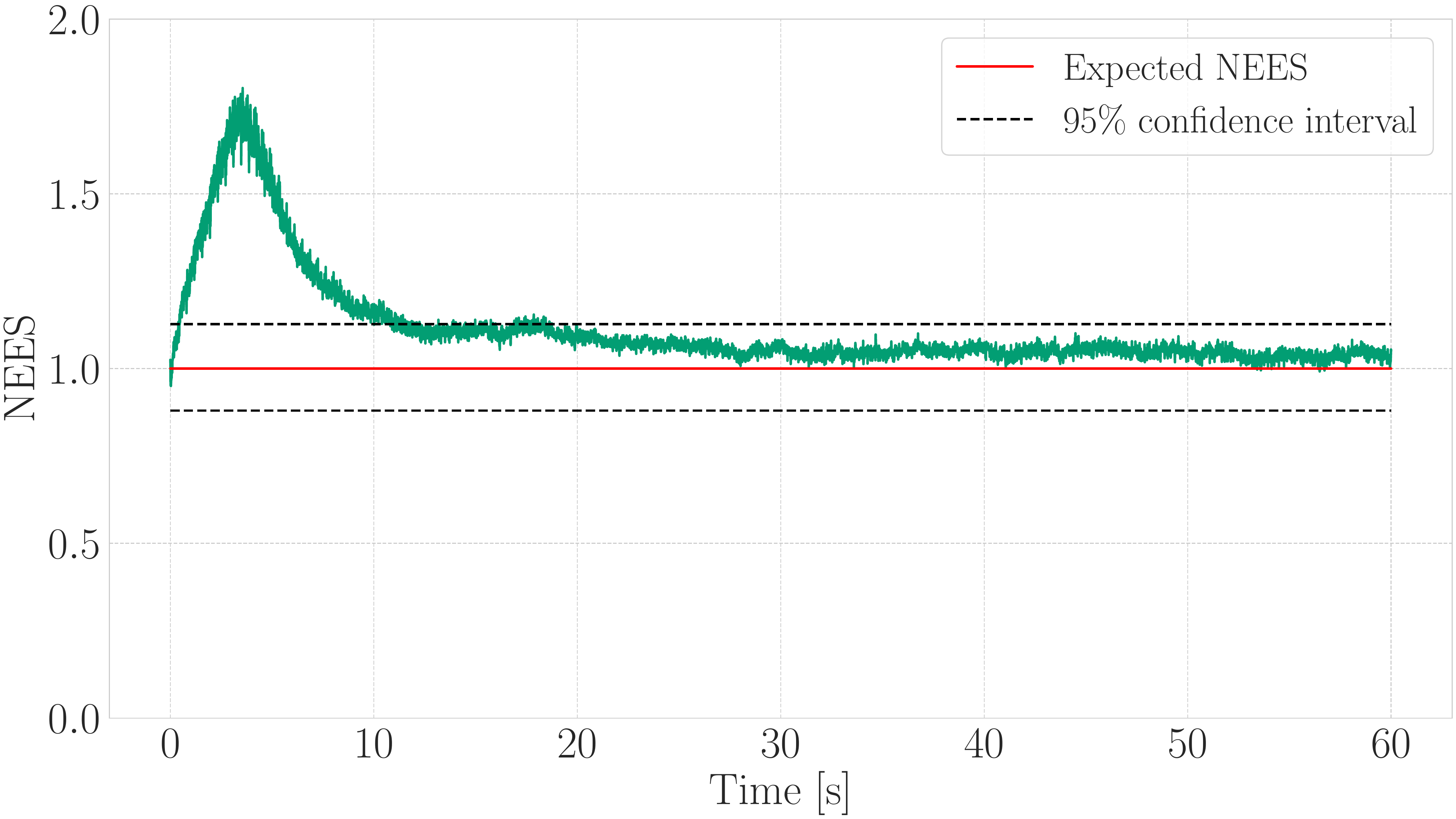}
    \caption{500-trial NEES plot for the proposed estimator on Simulation S3.}
    \label{fig:nees}
\end{figure}

A better evaluation of the consistency of the estimator is a NEES test, which is performed over the 500 trials of Simulation S3 
and is shown in Figure \ref{fig:nees}. During the first few seconds when the quadcopters are taking off from the ground, 
their geometry and low speeds result in a weakly-observable system \cite{Cossette2021}, which results in overconfidence 
of the estimator as linearization-based filters can correct in unobservable directions \cite{Huang2010, Huang2011}. 
Nonetheless, the estimator then converges towards consistency, %albeit not fully consistent 
\jln{although it is never perfectly consistent} due to linearization and discretization errors, which is a feature of EKFs. 
This can be solved by slightly inflating the associated covariance matrices used in the filter.

%% file: sections/experiment.tex
\begin{figure*}[t!]
    \centering
    \begin{minipage}{0.39\textwidth}%
        \centering
        \includegraphics[width=\textwidth]{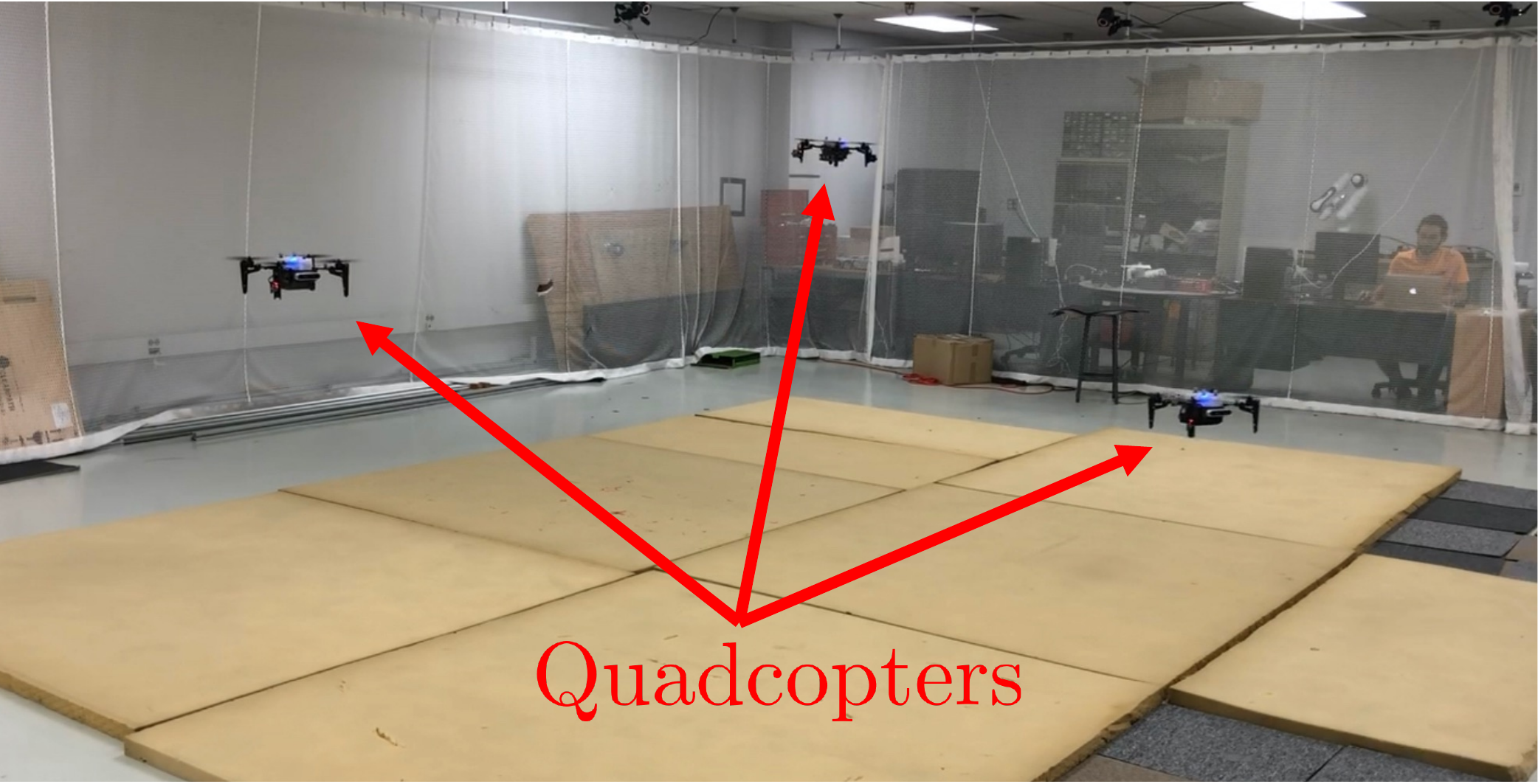}
        \label{fig:flying_drones_marked}
    \end{minipage}%
    ~
    \begin{minipage}{0.29\textwidth}%
        \centering
        \includegraphics[trim={25cm 10cm 32cm 10cm},clip,width=\textwidth]{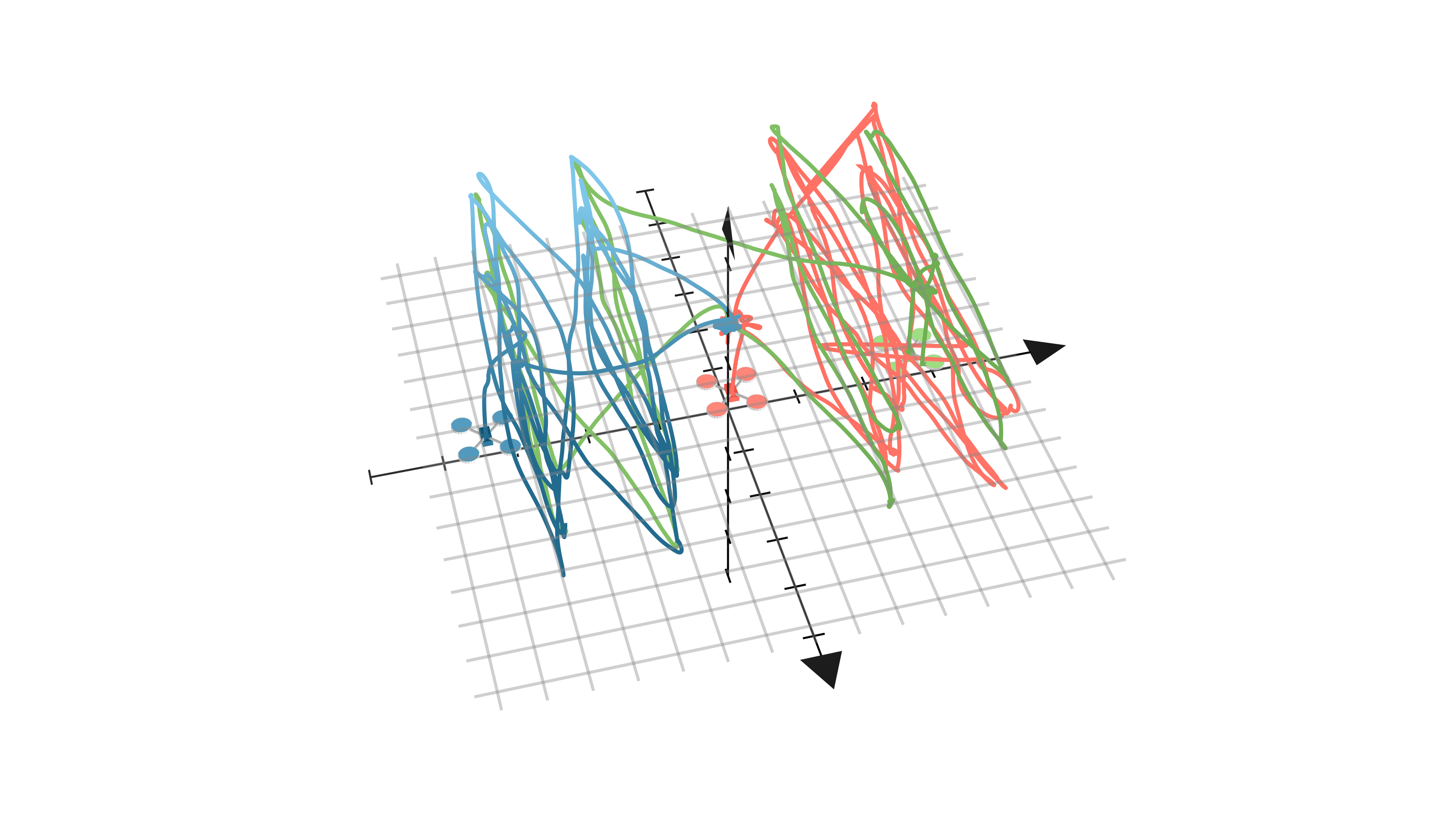}
        \label{fig:exp_trajectory_1}
    \end{minipage}
    ~
    \begin{minipage}{0.29\textwidth}%
        \centering
        \includegraphics[trim={30cm 10cm 25cm 10cm},clip,width=\textwidth]{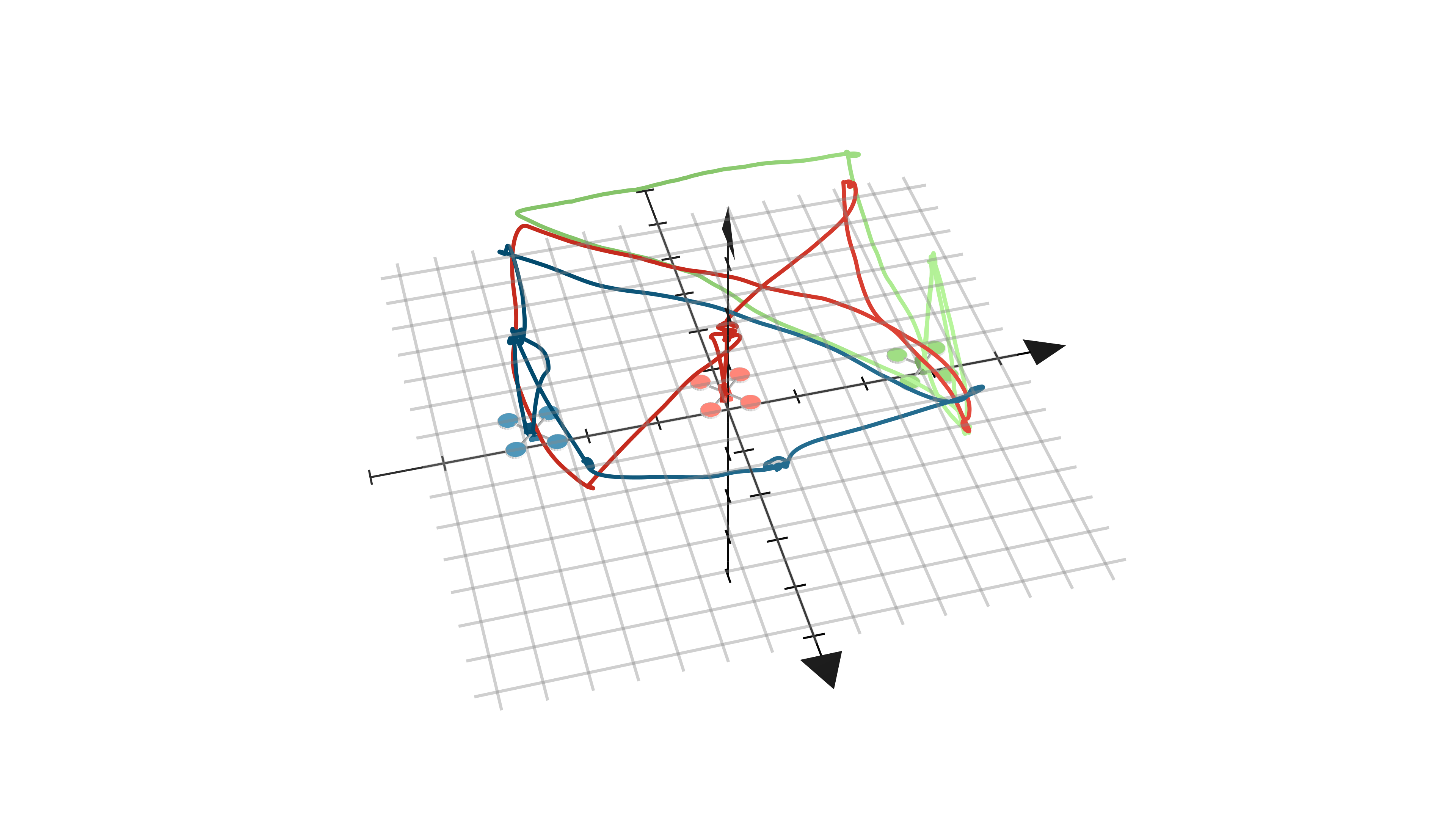}
        \label{fig:exp_trajectory_2}
    \end{minipage}
    \caption{(Left) 3 quadcopters in the experimental space. (Middle) The experimental trajectory for Trial 1, where each colour represents the trajectory of a different quadcopter and the grid represents a roughly 5 m $\times$ 5 m area. (Right) The experimental trajectory for Trial 2.}
    \label{fig:3_drones}
\end{figure*}

\begin{figure}[t!]
    \centering
    \begin{minipage}{0.9\columnwidth}%
        \centering
        \subfloat[Subfigure 1 list of figures text][Centralized.]{
        \includegraphics[width=\textwidth]{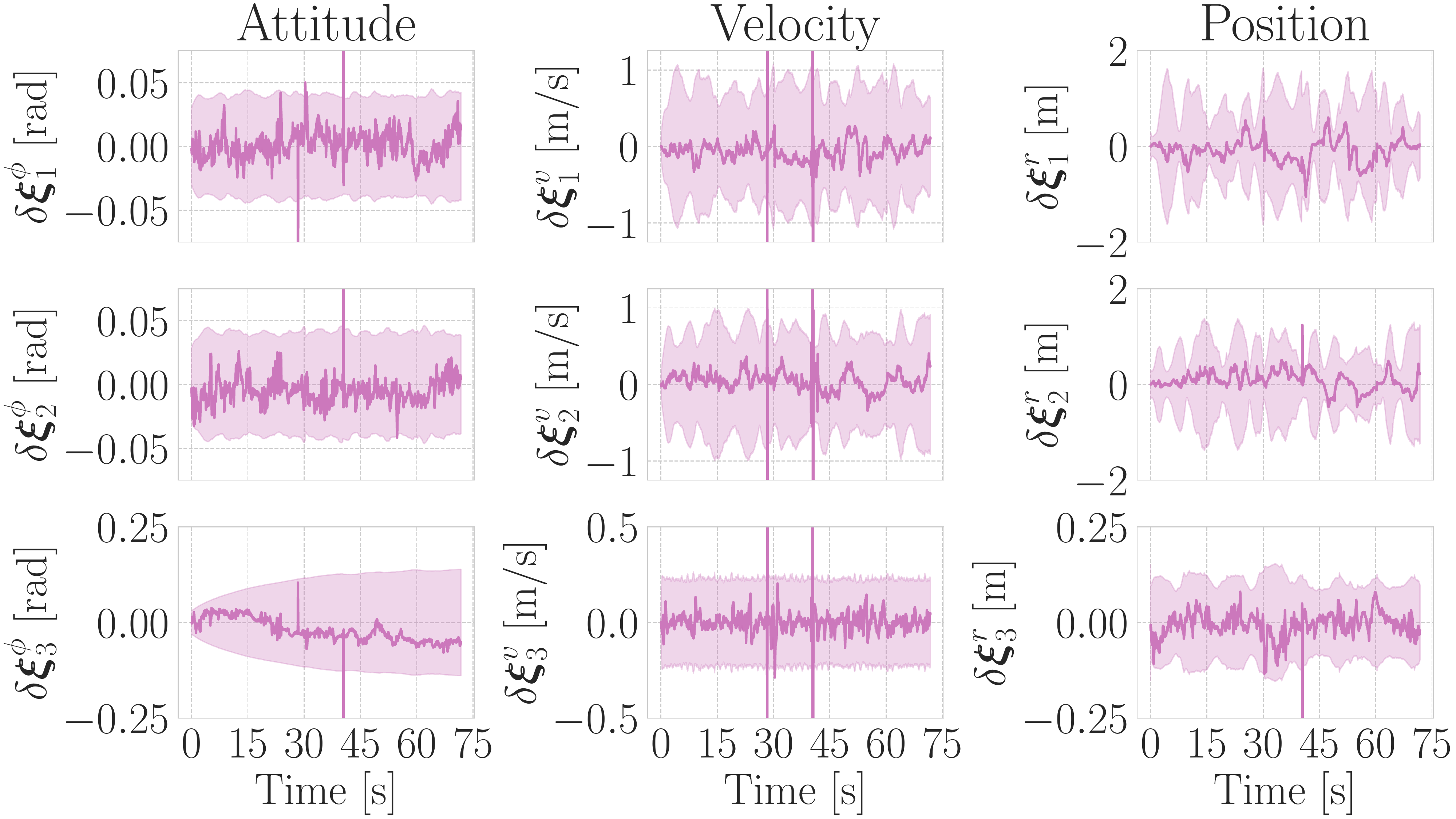}
        \label{fig:exp_pose_3sigma_centralized}}
    \end{minipage}
    \begin{minipage}{0.9\columnwidth}%
        \centering
        \subfloat[Subfigure 1 list of figures text][No passive listening.]{
        \includegraphics[width=\textwidth]{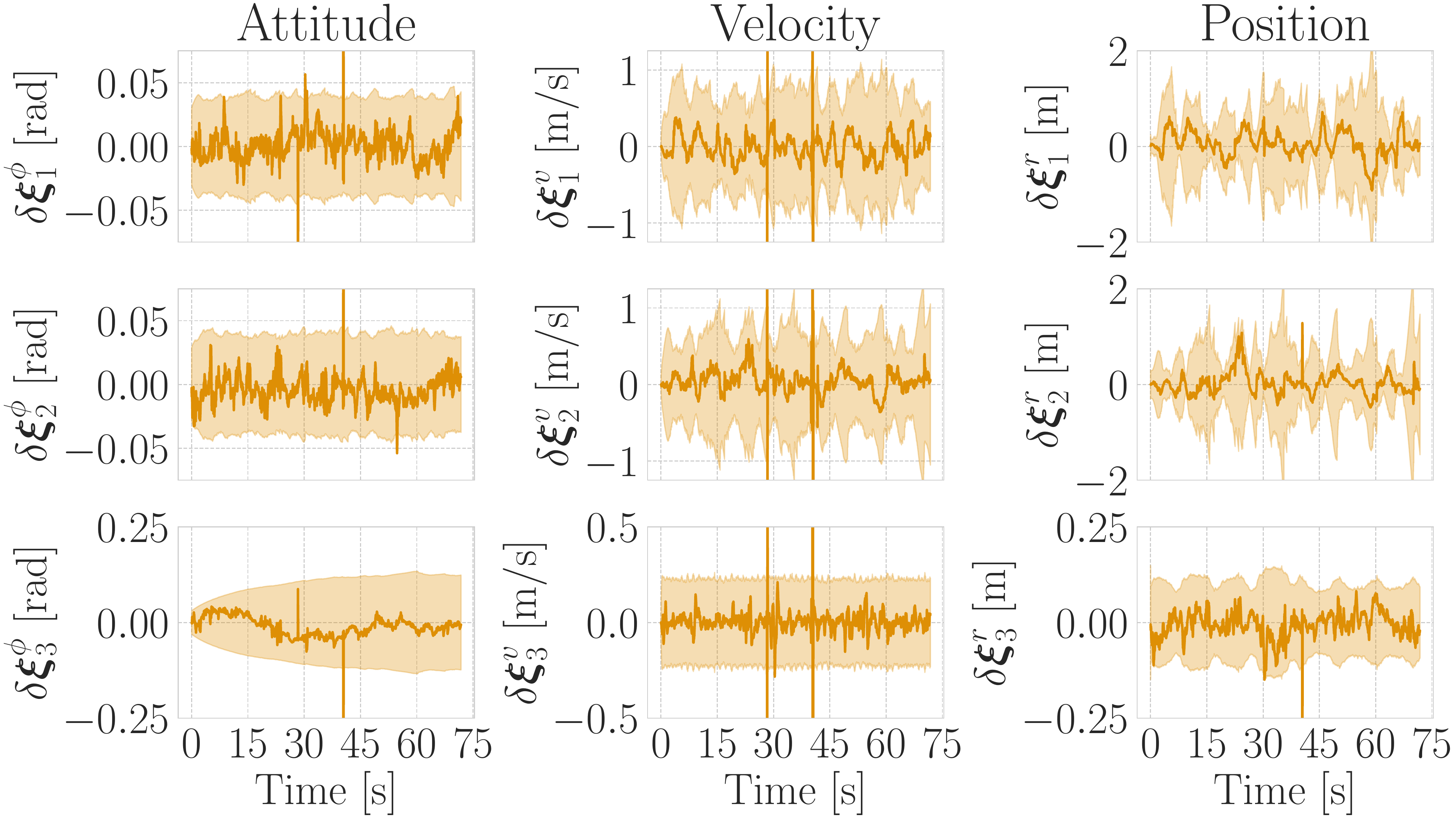}
        \label{fig:exp_pose_3sigma_np}}
    \end{minipage}
    \begin{minipage}{0.9\columnwidth}%
        \centering
        \subfloat[Subfigure 2 list of figures text][With passive listening.]{
        \includegraphics[width=\textwidth]{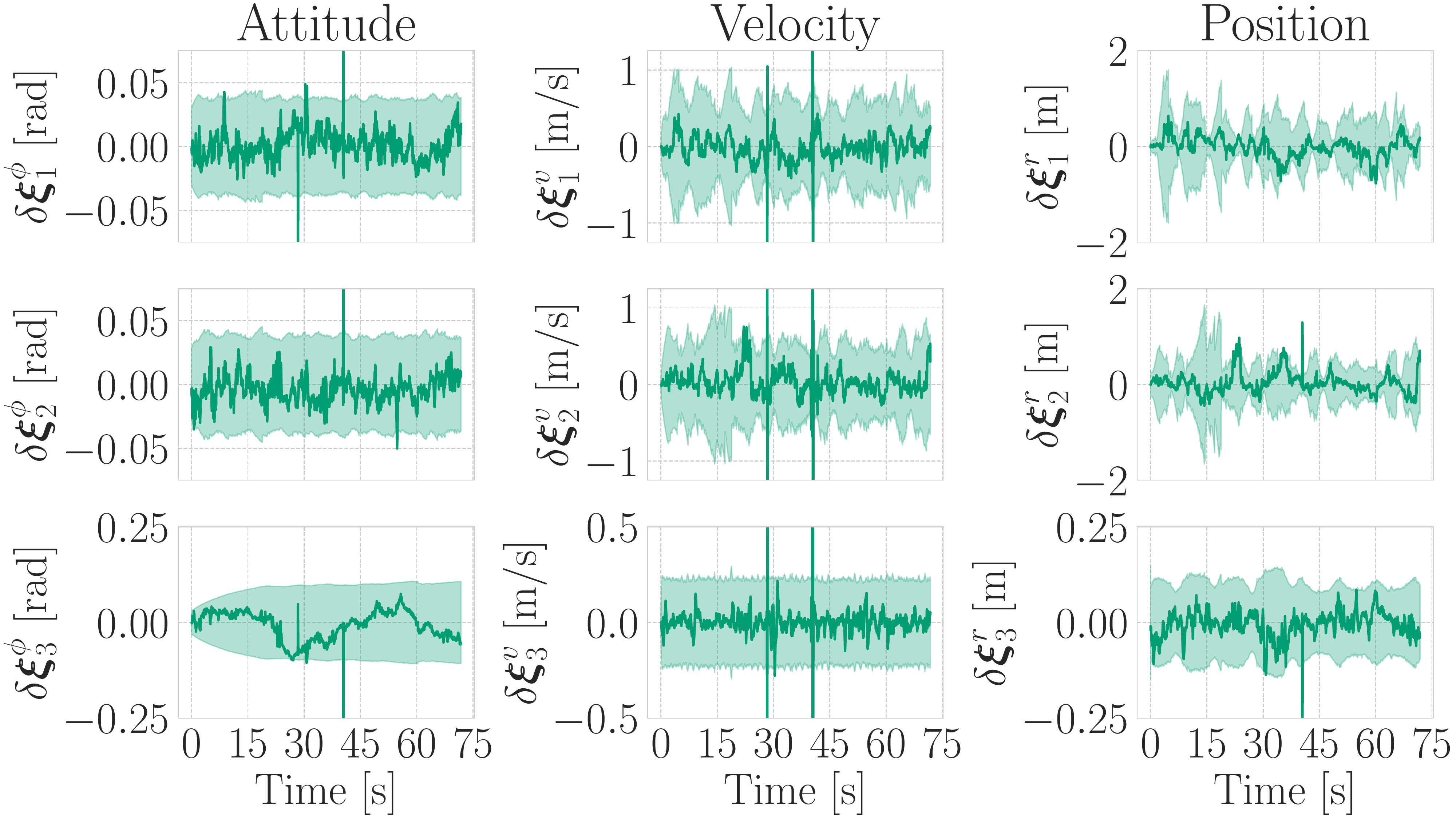}
        \label{fig:exp_pose_3sigma_wp}}
    \end{minipage}
    \caption{\tro{Error plots and $\pm3\sigma$ bounds (shaded region) for Robot 0's estimate of Robot 1's relative pose for experimental trial 1.}}
    \label{fig:exp_pose}
\end{figure}

The proposed approach is tested on multiple experimental trials. The ranging protocol discussed 
in Section \ref{sec:ranging_protocol} is implemented in C on custom-made boards fitted with 
DWM1000 UWB transceivers \cite{dw1000}. Two boards are then fitted to Uvify IFO-S quadcopters 
approximately 45 cm apart. The experimental set-up is shown in Figure \ref{fig:exp_setup}. 
Three of these quadcopters are then used for the experimental results shown in this section, 
with multiple \tro{approximately-75-second-long} trajectories similar to the one shown in Figure \ref{fig:3_drones} 
in a roughly 5 m $\times$ 5 m area. \tro{The quadcopters in the experimental trajectories each cover a distance between 20 m and 35 m, with a maximum speed of 3.75 m/s. The maximum and mean angular velocities are 2.12 rad/s and 0.3 rad/s, respectively.} In order to analyze the error in the pose estimates of the robots, 
a 12-camera Vicon motion-capture system is used to record the ground-truth pose of each quadcopter.

To enable the 6 transceivers to take turn ranging, the common-list protocol discussed 
in Section \ref{sec:ranging_protocol} is implemented using the \emph{robot operating system} (ROS). 
This allows each robot to range with its neighbours at a rate of 90 Hz, and collect passive listening 
measurements at a rate of 150 Hz. These UWB measurements are corrected for antenna delays and power-induced 
biases using \cite{Shalaby2022a}, before fusing \jln{them} with the %IMU information and onboard 
\jln{onboard IMU and}
height measurements in \jln{the proposed} EKF. \ms{An ICM-20689 IMU is used with characteristics similar to the simulated ones given in Table \ref{tab:sim_params}, and the height measurements are obtained from a downward-facing camera. The height measurement error is assumed Gaussian with 5 cm of standard deviation}. To reject outliers in the range and passive-listening measurements, 
the \emph{normalized-innovation-squared} (NIS) test is used in the filter \cite[Section~5.4]{barshalom2002}. 

\tro{Note that before flight, all transceivers are allowed to range with one another to initialize the relative clock offset states using the second pseudomeasurement from Section \ref{sec:ranging_protocol}, alongside a pseudomeasurement $y^\gamma = \f{\tilde{\mathtt{R}}^3 - \tilde{\mathtt{R}}^2}{\tilde{\mathtt{T}}^3 - \tilde{\mathtt{T}}^2} - 1 \approx \gamma_{f_1 s_2}$ that is not used in the filter.} \tro{Meanwhile, the IMU biases are initialized using the motion capture system and are then assumed constant throughout the experiment, which is sufficient for the duration of the experiments presented here. Addressing IMU biases for longer experiments is presented in Section \ref{subsec:imu_biases}.}

% \newcolumntype{C}{>{\centering\arraybackslash}m{1.5cm}}
% % \newcolumntype{D}{>{\centering\arraybackslash}m{1.5cm}}
% \begin{table*}[h!]
%     \footnotesize
%     \renewcommand{\arraystretch}{1.2}
%     \caption{\tro{The RMSE of Robot 0's estimate of neighbouring robots' relative pose for multiple experimental trials.}}
%     \label{tab:exp_rmse}
%     \centering
%     \begin{tabular}{C|CCC|CC}
%     % \hline
%     & \multicolumn{3}{c}{\bfseries Position RMSE averaged over all Robots} & \multicolumn{2}{c}{\bfseries Percentage Change}\\
%     \hline 
%      & \bfseries Centralized aRMSE [m] & \bfseries No Passive aRMSE [m]  & \bfseries Proposed aRMSE [m] & \bfseries vs. Centralized [\%] & \bfseries vs. No Passive [\%]\\
%     \hline
%     Trial 1 & 0.390 & 0.437 & \bfseries 0.341 & -12.56 & -21.97 \\
%     Trial 2 & 0.614 & 0.954 & \bfseries 0.576 & -6.19 & -39.62 \\
%     Trial 3 & 0.462 & 0.593 & \bfseries 0.443 & -4.11 & -25.30 \\
%     Trial 4 & 0.580 & 0.859 & \bfseries 0.445 & -23.28 & -48.20 
%     % \hline
%     \end{tabular}
% \end{table*}

\newcolumntype{D}{>{\centering\arraybackslash}m{1cm}}
\newcolumntype{E}{>{\centering\arraybackslash}m{2.5cm}}
\newcolumntype{F}{>{\centering\arraybackslash}m{3.8cm}}
% \newcolumntype{D}{>{\centering\arraybackslash}m{1.5cm}}
\begin{table}[h!]
    \footnotesize
    \renewcommand{\arraystretch}{1.2}
    \caption{\tro{The RMSE of Robot 0's estimate of neighbouring robots' relative pose for multiple experimental trials.  The percentage change is $\f{\text{Proposed} - \text{Comparison}}{\text{Comparison}}$, where the Comparison is either Centralized or No Passive.}}
    \label{tab:exp_rmse}
    \centering
    \begin{tabular}{D|DDD|DD}
    % \hline
    & \multicolumn{3}{F}{\bfseries Position RMSE averaged over all Robots [m]} & \multicolumn{2}{E}{\bfseries Percentage change [\%]}\\
    \hline 
     & \bfseries Centr. & \bfseries No Passive & \bfseries Proposed & \bfseries Centr. & \bfseries No Passive\\
    \hline
    Trial 1 & 0.390 & 0.437 & \bfseries 0.341 & -12.56 & -21.97 \\
    Trial 2 & 0.614 & 0.954 & \bfseries 0.576 & -6.19 & -39.62 \\
    Trial 3 & 0.462 & 0.593 & \bfseries 0.443 & -4.11 & -25.30 \\
    Trial 4 & 0.580 & 0.859 & \bfseries 0.445 & -23.28 & -48.20 
    % \hline
    \end{tabular}
\end{table}

The pose-error plots for one of the trials are shown in Figure \ref{fig:exp_pose} for the centralized approach, and with and without fusing 
passive listening measurements. The RMSE comparison for 4 different trials with varying motion 
are shown in Table \ref{tab:exp_rmse}. Even though all scenarios result in error trajectories that fall 
within the error bounds, it is clear that with the additional passive listening measurements available 
to the robot at 150 Hz, the relative position estimates in particular become significantly less uncertain. 
Additionally, these error plots correspond to the first row in Table \ref{tab:exp_rmse}, showing that 
the improvement in the confidence of the estimator is additionally accompanied with a \tro{12.56\% and 21.97\% 
reduction in the RMSE as compared to the centralized and no passive listening position RMSE, respectively. This reduction in RMSE 
goes up to 23.28\% and 48.20\%, respectively, for one of the runs when passive listening measurements are utilized.}

% \begin{table*}
%     \footnotesize
%     \renewcommand{\arraystretch}{1.2}
%     \caption{The RMSE of Robot 0's estimate of neighbouring robots' relative pose for multiple experimental trials.}
%     \label{tab:exp_rmse}
%     \centering
%     \begin{tabular}{C|CCC|CCC}
%     % \hline
%     & \multicolumn{3}{c}{\bfseries Robot 1} & \multicolumn{3}{c}{\bfseries Robot 2} \\
%     \hline 
%     & \bfseries No Passive RMSE [m] & \bfseries Proposed RMSE [m] & \bfseries Change [\%] 
%     & \bfseries No Passive RMSE [m] & \bfseries Proposed RMSE [m] & \bfseries Change [\%] \\
%     \hline
    % Trial 1 & 0.518 & \bfseries 0.379 & -26.83 & 0.356 & \bfseries 0.303 & -14.89 \\
    % Trial 2 & 1.079 & \bfseries 0.641 & -40.59 & 0.828 & \bfseries 0.511 & -38.29 \\
    % Trial 3 & 0.603 & \bfseries 0.475 & -21.23 & 0.583 & \bfseries 0.410 & -29.67 \\
    % Trial 4 & 1.007 & \bfseries 0.445 & -55.81 & 0.711 & \bfseries 0.465 & -34.60
%     % \hline
%     \end{tabular}
%     \vspace{-10pt}
% \end{table*}

%% file: sections/practical.tex
\subsection{\tro{IMU Biases}} \label{subsec:imu_biases}

\tro{
    The IMU measurements typically suffer from time-varying biases, which must be estimated as part of the state for long-term navigation. It can be shown that, when modelling the evolution of biases as a random walk, the IMU biases can be incorporated into the process model while still maintaining the differential Sylvester equation form presented in Section \ref{subsec:derive_process_model}. To do so, each Robot $i$ estimates its own gyroscope bias $\mbs{\beta}_i^{\text{gyr},i}$ in its own body frame, and uses this estimate to correct the IMU measurements and inflating the covariance when constructing the RMI. Additionally, each robot estimates a relative accelerometer bias to every neighbour in the robot's own body frame, which does not affect the computed RMI. For example, Robot 0's estimate of Robot $i$'s relative accelerometer bias is defined as $$\mbs{\beta}_0^{\text{acc},0i} \triangleq \mbs{\beta}_0^{\text{acc},0} - \mbf{C}_{0i}\mbs{\beta}_i^{\text{acc},i},$$ where $\mbs{\beta}_i^{\text{acc},i}$ is Robot $i$'s accelerometer bias. The interested reader can refer to \cite{Shalaby2023d} for derivation of the pose and bias process models, corresponding linearization, preintegration, and simulation and experimental results.
}

\subsection{\tro{Incomplete and Dynamic Communication Graphs}} \label{subsec:dynamic_graphs}

\tro{
The proposed framework has been evaluated under the assumptions of a full communication graph, no packet drop, and no communication failures. Nonetheless, these are all real-world problems that must be addressed before implementing the proposed framework. This is beyond the scope of this paper; nonetheless, a brief discussion regarding these issues is provided in this section. 
}

\tro{
The ranging protocol and the proposed estimator do not require a full communication graph and the lack of communication failures. However, the common-list MAC protocol does. The common-list MAC protocol is a very simple approach made possible only due to passive listening, and is ideal for small teams of robots that will always be within communication range with one another, thus allowing a full communication graph. Whenever a ranging transaction between a pair of transceivers fails, it is reattempted multiple times until a timeout is triggered, after which the ranging pair and all other robots who have not heard a message for the duration of the timeout move onto the next entry in the list. The protocol can handle a robot's communication failure by having each robot eliminate an element in the list when it fails more than $\kappa$ times, where $\kappa$ is a user-defined threshold. 
}

\tro{
When extending to larger teams, it is not possible to assume that all robots are within communication range of one another, thus invalidating the full-communication-graph assumption. Additionally, robots might fall in and out of range with one another over time, thus necessitating an incomplete dynamic communication graph model. In such scenarios, the common-list protocol is no longer simple, as robots need to know what other robots out of communication range are doing. Therefore, such systems may benefit from other MAC protocols such as token passing \cite[Section 3.3]{miao2016}, which is still possible with the proposed ranging protocol and estimator as they are independent of the choice of the MAC protocol. The benefits of passive listening thus still stand, not just due to additional measurements, but because it also allows each robot to maintain a list of neighbours within communication range.
}

\tro{Another implication of incomplete graphs is that each robot only estimates relative poses for the subset of robots that lie within its communication range. This is useful as it reduces the dimensionality of the onboard estimator, since each robot only estimates the relative states of $m < n$ neighbouring robots. However, having dynamic graphs due to robots falling in and out of the communication range of the robot mean that the robots must initialize the states of neighbours when they appear and marginalize out the states of neighbours that have not been within the communication range for an extended period of time. The initialization can potentially be done by listening to a window of measurements from the new neighbour and formulating a least-squares problem.}

%% file: sections/conclusion.tex
In this paper, the problem of relative extended pose estimation has been addressed for a team of robots each equipped with UWB transceivers. A novel ranging protocol is proposed that allows neighbouring robots to passively listen-in on the measurements without any underlying assumptions on the hierarchy of the communication. This is then utilized to implement a simple MAC protocol and an efficient means for sharing preintegrated IMU information, which is then fused with the UWB measurements in a filter that estimates both the clock states of the transceivers and the relative poses of the robots. The relative poses and the preintegration are formulated directly on $SE_2(3)$. This is then all evaluated in simulation using different numbers of robots and Monte-Carlo trials, and in experiments using multiple trials of 3 quadcopters each equipped with 2 UWB transceivers. The method is shown to improve the localization performance significantly when compared to \tro{centralized scenarios} or to the case of no passive listening measurements. 

This work can be extended to address complications that arise in wireless communication, such as packet drop. When a packet drop occurs, neighbours miss an RMI which is required to propagate their estimates forward, and therefore this must be addressed in a real-world application, potentially by providing a means for robots to request a missed RMI from their neighbours. \ms{Future work will additionally consider more efficient \tro{MAC protocols} where only a subset of the transceivers range with one another in pairs while the remaining transceivers are always passively listening.} \tro{Alongside the discussion in Section \ref{sec:practical},} another potential extension of this paper includes collaboration between robots, as robots can share their state estimates with neighbours to reach a consensus on the clock and relative pose states.

%% file: sections/appx_fold.tex
\jln{When} there are $n+1$ robots and 2 transceivers per robot, the total 
number of transceivers is $n_t = 2(n+1)$. Therefore, the number of 
ranging pairs with transceivers on distinct robots is 
\begin{align*}
    n_\text{p} &= \f{2(n+1)(2(n+1)-1)}{2} - (n+1) = 2n(n+1).
\end{align*}
The number of direct measurements between all robots is then 2$n_\text{p}$ 
(one range and one offset measurement per pair), while the number of passive 
listening measurements recorded at all robots is $n_\text{p} (3 (n_\text{t} - 2)) = 6 n n_\text{p}$.
Therefore, the fold increase in measurements is 
\beq
    \f{2 n_\text{p} + 6 n n_\text{p}}{2 n_\text{p}} = 1 + 3n \nonumber
\eeq
when considering a centralized approach where passive listening measurements 
from all robots are available.

A similar analysis can be done from the perspective of one robot that does not have access to passive listening measurements recorded at neighbouring robots. Without passive listening it can be shown that the robot only gets $8n$ distinct measurements, while with listening-in on neighbouring robots' messages the robot gets $2n_\text{p} - 8n$ new measurements from the direct measurements between the neighbours and $12n^2$ new passive listening measurements. This can be shown to be a $(\f{1}{2}+2n)$-fold increase in the number of measurements from the individual robot's perspective.

%% file: sections/appx_linearize_range.tex
Consider \jln{as in \eqref{eq:dist} an expression} of the form
\begin{align}
    \label{eq:appx_meas}
    d &= \norm{\Big( \mbs{\Pi} \big(  \mbf{T}_{2} \mbftilde{r}_2 -  \mbf{T}_{1} \mbftilde{r}_1 \big) \Big)},
\end{align}
where $\mbf{T}_1, \mbf{T}_2 \in SE(3)$ and $\mbf{r}_1, \mbf{r}_2 \in \mathbb{R}^5$. Squaring both sides and perturbing the measurement and the pose states yields
\begin{align*}
    (\bar{d} + \delta{d})^2 \nonumber &= \Big( \mbs{\Pi} \big( \operatorname{Exp}(\mbsdel{\xi}_{2}) \mbfbar{T}_{2} \mbftilde{r}_2 - \operatorname{Exp}(\mbsdel{\xi}_{1}) \mbfbar{T}_{1} \mbftilde{r}_1 \big) \Big)^\trans \Big( \cdot \Big),
\end{align*}
which, \jln{using \eqref{eq:exp_approx}}, can be expanded to give
\begin{align*}
    \bar{d}^2 + 2\bar{d}\delta{d} %&\stackrel{\eqref{eq:exp_approx}}{\approx} 
    & \approx
    (\mbs{\Pi} \mbfbar{T}_{2} \mbftilde{r}_2)^\trans \mbs{\Pi} \mbfbar{T}_{2} \mbftilde{r}_2 + (\mbs{\Pi} \mbfbar{T}_{1} \mbftilde{r}_1)^\trans \mbs{\Pi} \mbfbar{T}_{1} \mbftilde{r}_1 \nonumber \\ 
    &\hspace{12pt}- (\mbs{\Pi} \mbfbar{T}_{2} \mbftilde{r}_2)^\trans \mbs{\Pi} \mbfbar{T}_{1} \mbftilde{r}_1 - (\mbs{\Pi} \mbfbar{T}_{1} \mbftilde{r}_1)^\trans \mbs{\Pi} \mbfbar{T}_{2} \mbftilde{r}_2 \nonumber \\ 
    &\hspace{12pt} - (\mbs{\Pi}\mbsdel{\xi}_2^\wedge \mbfbar{T}_{2} \mbftilde{r}_2)^\trans \mbs{\Pi} \mbfbar{T}_{1} \mbftilde{r}_1 
    - (\mbs{\Pi} \mbfbar{T}_{2} \mbftilde{r}_2)^\trans \mbs{\Pi} \mbsdel{\xi}_1^\wedge \mbfbar{T}_{1} \mbftilde{r}_1 
    \nonumber \\ 
    &\hspace{12pt}- (\mbs{\Pi}\mbsdel{\xi}_1^\wedge \mbfbar{T}_{1} \mbftilde{r}_1)^\trans \mbs{\Pi} \mbfbar{T}_{2} \mbftilde{r}_2 
    - (\mbs{\Pi} \mbfbar{T}_{1} \mbftilde{r}_1)^\trans \mbs{\Pi} \mbsdel{\xi}_2^\wedge \mbfbar{T}_{2} \mbftilde{r}_2,
\end{align*}
where higher order terms have been neglected. 
Cancelling out the nominal terms on both sides, \jln{using the fact that 
each term is scalar, and recalling \eqref{eq:odot},}
\begin{align*}
    &2\bar{d}\delta{d} = -2(\mbs{\Pi} \mbfbar{T}_{2} \mbftilde{r}_2)^\trans \mbs{\Pi} \mbsdel{\xi}_1^\wedge \mbfbar{T}_{1} \mbftilde{r}_1 
    - 2(\mbs{\Pi} \mbfbar{T}_{1} \mbftilde{r}_1)^\trans \mbs{\Pi} \mbsdel{\xi}_2^\wedge \mbfbar{T}_{2} \mbftilde{r}_2 \nonumber \\
    %&\stackrel{\eqref{eq:odot}}{=} 
    & \;\; =
    -2(\mbs{\Pi}^\trans \mbs{\Pi} \mbfbar{T}_{2} \mbftilde{r}_2)^\trans (\mbfbar{T}_{1} \mbftilde{r}_1)^\odot \mbsdel{\xi}_1 
    %\nonumber \\ 
    %%
    %&\hspace{12pt} 
    - 2(\mbs{\Pi}^\trans \mbs{\Pi} \mbfbar{T}_{1} \mbftilde{r}_1)^\trans (\mbfbar{T}_{2} \mbftilde{r}_2)^\odot \mbsdel{\xi}_2.
\end{align*}
Therefore, the linearized model for \eqref{eq:appx_meas} is 
\begin{align*}
    \delta{d} &= -\f{1}{\bar{d}} (\mbs{\Pi}^\trans \mbs{\Pi} \mbfbar{T}_{2} \mbftilde{r}_2)^\trans (\mbfbar{T}_{1} \mbftilde{r}_1)^\odot \mbsdel{\xi}_1 \nonumber 
    \\ 
    &\hspace{12pt} 
    - \f{1}{\bar{d}} (\mbs{\Pi}^\trans \mbs{\Pi} \mbfbar{T}_{1} \mbftilde{r}_1)^\trans (\mbfbar{T}_{2} \mbftilde{r}_2)^\odot \mbsdel{\xi}_2.
\end{align*}

%% file: sections/appx_U.tex
The matrices $\mbftilde{U}_{0,k}$ and $\mbftilde{U}_{i,k}$ in \eqref{eq:process_model_dt} are of the general form 
\beq
    \mbftilde{U} = \bma{cc}
        \mbf{u}^\wedge & \mbf{e}_4 \\
        \mbf{0}_{\jln{1 \times 4}} & 0
    \ema,
\eeq
where $\mbf{u} = \bma{cc} \mbs{\omega}^\trans & \mbs{\alpha}^\trans \ema^\trans$, $(\cdot)^\wedge$ is the wedge operator in $SE(3)$, and 
$\mbf{e}_4 = \bma{cc} \mbf{0}_{\jln{1 \times 3}} & 1\ema^\trans$. Consequently, 
\begin{align}
    \mbf{U} &= \operatorname{exp}(\mbftilde{U} \Delta t)  %\nonumber \\
    %%
    %&= 
    =\sum_{\ell = 0}^\infty \f{1}{\ell !} \left( \mbftilde{U} \Delta t \right)^\ell \nonumber \\
    &= \mbf{1} + \bma{cc}
        \mbf{u}^\wedge & \mbf{e}_4 \\
        \mbf{0} & 0
    \ema\Delta t + \f{1}{2!} \bma{cc}
        (\mbf{u}^\wedge)^2 & \mbf{u}^\wedge \mbf{e}_4 \\
        \mbf{0} & 0
    \ema (\Delta t)^2 \nonumber \\ &\hspace{12pt} + \f{1}{3!} \bma{cc}
        (\mbf{u}^\wedge)^3 & (\mbf{u}^\wedge)^2 \mbf{e}_4 \\
        \mbf{0} & 0
    \ema (\Delta t)^3 + \ldots \nonumber \\
    &= \bma{cc}
        \sum_{\ell = 0}^\infty \f{1}{\ell !} (\mbf{u}^\wedge \Delta t)^\ell & \sum_{\ell = 0}^\infty \f{1}{(\ell + 1) !} (\mbf{u}^\wedge \Delta t)^\ell \mbf{e}_4 \Delta t \\
        \mbf{0} & 1 
    \ema. \label{eq:appx_U_unexpanded}
\end{align}
Note that $\sum_{\ell = 0}^\infty \f{1}{\ell !} (\mbf{u}^\wedge \Delta t)^\ell = \operatorname{Exp}(\mbf{u} \Delta t)$, where $\operatorname{Exp}$ is the $SE(3)$ exponential operator, giving
\beq
    \label{eq:Exp_se3}
    \sum_{\ell = 0}^\infty \f{1}{\ell !} (\mbf{u}^\wedge \Delta t)^\ell = \bma{cc}
        \operatorname{Exp}(\mbs{\omega} \Delta t) & \Delta t \mbf{J}_l(\mbs{\omega} \Delta t) \mbs{\alpha} \\
        \mbf{0} & 1
    \ema.
\eeq
$\mbf{J}_l$ is the left Jacobian of $SO(3)$, which is of the form
\begin{align*}
    \mbf{J}_l (\mbs{\psi}) &= \sum_{\ell=0}^\infty \f{1}{(\ell+1)!} \left( \phi \mbs{\phi}^\times \right)^\ell \\
    &= \f{\operatorname{sin}\phi}{\phi} \mbf{1} + \left( 1 - \f{\operatorname{sin}\phi}{\phi} \right) \mbs{\phi}\mbs{\phi}^\trans + \f{1-\operatorname{cos}\phi}{\phi} \mbs{\phi}^\times,
\end{align*}
where $\phi = \vert \mbs{\psi} \vert$ and $\mbs{\phi} = \mbs{\psi} / \phi$. Meanwhile,
\begin{align}
    &\sum_{\ell = 0}^\infty \f{1}{(\ell + 1) !} (\mbf{u}^\wedge \Delta t)^\ell \nonumber \\
    &= \mbf{1} + \f{1}{2!} \bma{cc}
        \mbs{\omega}^\times & \mbs{\alpha} \\
        \mbf{0} & 0
    \ema\Delta t + \f{1}{3!} \bma{cc}
        (\mbs{\omega}^\times)^2 & \mbs{\omega}^\times \mbs{\alpha} \\
        \mbf{0} & 0
    \ema (\Delta t)^2 \nonumber \\ &\hspace{12pt} + \f{1}{4!} \bma{cc}
        (\mbs{\omega}^\times)^3 & (\mbs{\omega}^\times)^2 \mbs{\alpha} \\
        \mbf{0} & 0
    \ema (\Delta t)^3 + \ldots \nonumber \\
    &= \bma{cc}
        \sum_{\ell = 0}^\infty \f{1}{(\ell + 1) !} (\mbs{\omega}^\times \Delta t)^\ell & \sum_{\ell = 0}^\infty \f{1}{(\ell + 2) !} (\mbs{\omega}^\times \Delta t)^\ell \mbs{\alpha} \Delta t \\
        \mbf{0} & 1 
    \ema \nonumber \\
    &= \bma{cc}
        \mbf{J}_l(\mbs{\omega} \Delta t) & \frac{\Delta t}{2}\mbf{N}_l(\mbs{\omega} \Delta t) \mbs{\alpha} \\
        \mbf{0} & 1 
    \ema, \label{eq:appx_U_second_component}
\end{align}
where
\begin{align*}
    \mbf{N} (\mbs{\psi}) &= 2 \sum_{\ell=0}^\infty \f{1}{(\ell+2)!} \left( \phi \mbs{\phi}^\times \right)^\ell \\
    &= \mbs{\phi}\mbs{\phi}^\trans + 2\left( \f{1}{\phi} - \f{\operatorname{sin}\phi}{\phi^2} \right)\mbs{\phi}^\times + 2\f{\operatorname{cos}\phi-1}{\phi^2}\mbs{\phi}^\times\mbs{\phi}^\times.
\end{align*}
Substituting \eqref{eq:Exp_se3} and \eqref{eq:appx_U_second_component} back into \eqref{eq:appx_U_unexpanded} gives
\begin{equation*}
    \mbf{U} = \bma{ccc}
        \operatorname{Exp}(\mbs{\omega} \Delta t) & \Delta t \mbf{J}_l(\mbs{\omega} \Delta t) \mbs{\alpha} & \frac{\Delta t^2}{2}\mbf{N}_l(\mbs{\omega} \Delta t) \mbs{\alpha} \\
         & 1 & \Delta t \\
         & & 1
    \ema.
\end{equation*}